\documentclass[sigconf,nonacm]{acmart}

\usepackage{pifont}
\newcommand{\xmark}{\ding{55}}
\usepackage{algorithm}
\usepackage{algpseudocode}

\usepackage{multirow}
\usepackage{makecell}
\usepackage{tabularx}
\usepackage{threeparttable}
\usepackage{hhline}
\usepackage{colortbl}   
\usepackage{csvsimple}  
\usepackage{framed}     
\usepackage{nicefrac}   
\usepackage{siunitx}    

\usepackage{subcaption} 
\usepackage{wrapfig}
\usepackage{float}

\usepackage{bm}
\usepackage{bbm}       
\usepackage{dsfont}
\usepackage{mathalfa}
\usepackage{latexsym}
\usepackage{nicefrac}

\usepackage{tikz}
\usepackage{pgfplots}
\usepackage{pgfplotstable}
\usepackage{tikz-cd}
\usetikzlibrary{
  shapes.geometric,
  arrows,
  external,
  pgfplots.groupplots,
  matrix,
  positioning, 
  fit, 
  backgrounds, 
  calc,
  shadows
}
\pgfdeclarelayer{background}
\pgfsetlayers{background,main}

\usepackage{xparse}
\usepackage{xspace}
\usepackage{enumerate}  
\usepackage{comment}
\usepackage{cryptocode} 

\usepackage[most]{tcolorbox}
\tcbuselibrary{breakable, skins}

\newtcolorbox{takeaway}[1][Takeaway]{
    colback=gray!10,
    colframe=gray!40,
    arc=4pt,
    boxrule=0.8pt,
    title=#1,
    fonttitle=\bfseries\sffamily,
    left=8pt, top=10pt, bottom=10pt,
    enhanced,
    before=\vspace{2pt},
    after=\vspace{2pt}
}

\AtBeginDocument{%
  }

\begin{document}

\title{Auditing Data Provenance in LLM Fine-tuning via Intrinsic Distributional Fingerprints}


\author{Zirui Huang}
\email{652024330043@smail.nju.edu.cn}
\affiliation{%
  \institution{State Key Laboratory of Novel Software Technology, Nanjing University}
  \city{Nanjing}
  \state{Jiangsu}
  \country{China}
}

\author{Yunlong Mao}
\email{maoyl@nju.edu.cn}
\authornote{Corresponding author.}
\affiliation{%
  \institution{State Key Laboratory of Novel Software Technology, Nanjing University}
  \city{Nanjing}
  \state{Jiangsu}
  \country{China}
}

\author{Wei Tong}
\email{weitong@outlook.com}
\affiliation{%
  \institution{State Key Laboratory of Novel Software Technology, Nanjing University}
  \city{Nanjing}
  \state{Jiangsu}
  \country{China}
}

\author{Tingting Wu}
\email{wutingtingyjy@chinamobile.com}
\affiliation{%
  \institution{China Mobile Research Institute}
  \city{Beijing}
  \country{China}
}

\author{Xin Ge}
\email{gexin@chinamobile.com}
\affiliation{%
  \institution{China Mobile Research Institute}
  \city{Beijing}
  \country{China}
}

\author{Sheng Zhong}
\email{zhongsheng@nju.edu.cn}
\affiliation{%
  \institution{State Key Laboratory of Novel Software Technology, Nanjing University}
  \city{Nanjing}
  \state{Jiangsu}
  \country{China}
}

\renewcommand{\shortauthors}{Huang et al.}

\begin{abstract}
\label{abstract}

The proliferation of customized Large Language Models (LLMs) poses critical risks of Data Intellectual Property (Data IP) infringement via unauthorized fine-tuning on proprietary data. Existing audit techniques are limited, as they require intervention during data preparation or training and remain fragile under malicious obfuscations such as data paraphrasing and knowledge distillation.

We propose \textit{Distribution Provenance Audit (DPA)}, a post-hoc framework for auditing data IP infringement in LLM fine-tuning under black-box and malicious settings. DPA is grounded in a critical insight: regardless of fine-tuning tactics to evade provenance, the practical necessity of maintaining utility constrains the model to preserve the fundamental intersection of semantic substance and lexical form. Accordingly, DPA captures this persistent lexical-semantic intersection as intrinsic distributional fingerprints. The framework formulates the audit as a statistical hypothesis test, effectively quantifying these fingerprints via unbiased output sampling to reliably reject the null hypothesis of non-usage.

Extensive experiments on medical and legal fine-tuning tasks show that DPA consistently outperforms existing baselines, remaining robust against adversarial trainers employing paraphrasing and knowledge distillation. We further highlight a fundamental dual-use tension: the same high-fidelity distributional fingerprints enabling reliable auditing may also facilitate privacy attacks.


\end{abstract}

\ccsdesc[500]{Security and privacy~Digital rights management}
\ccsdesc[500]{Security and privacy~Privacy protections}
\ccsdesc[500]{Computing methodologies~Natural language generation}
\ccsdesc[500]{Security and privacy~Information flow control}

\keywords{Large Language Models; Data Provenance; Intellectual Property Protection; Distributional Fingerprints; Malicious Fine-tuning}




\maketitle

\section{Introduction}
\label{sec:intro}

The necessity of achieving expert-level Large Language Models (LLMs) in specialized domains (e.g., healthcare, legislation) has created a critical demand for fine-tuning on high-value, proprietary data~\cite{han2024peft-survey, guo2025deepseek}. However, this introduces severe risks of Data Intellectual Property (Data IP) infringement, where malicious trainers may exploit curated, copyrighted datasets to enhance model utility without authorization~\cite{qiu2025icassp-watermarking, chu2024protect-copyright}. This tension creates an urgent forensic imperative: Data Owners (DOs) require robust data provenance audits to verify whether a suspicious, often black-box LLM has been trained on their proprietary assets without consent~\cite{li2025datasetwatermark, xu2025copyright-survey}.

\begin{figure}[t!] 
  \centering
  \includegraphics[width=\linewidth]{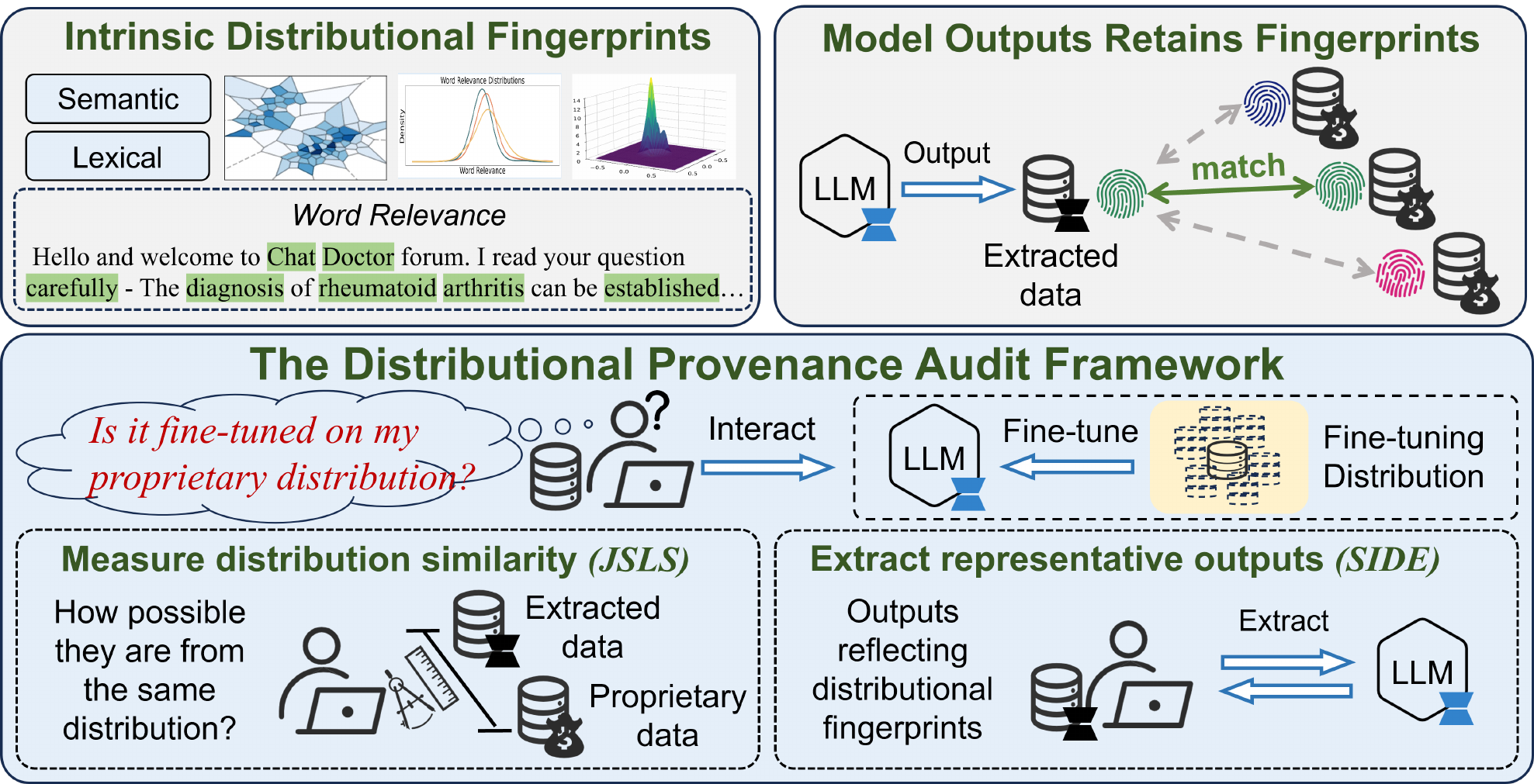} 
  \caption{Schematic of Distribution Provenance Audit (DPA).
  } 
  \label{fig:DPA-schema}
\end{figure}

Current auditing mechanisms, however, face a fundamental forensic deadlock in this malicious setting. Invasive methods like dataset watermarking~\cite{qiu2025icassp-watermarking,tang2023backdoor} or Zero-Knowledge Proofs (ZKPs)~\cite{sun2024zkllm}, are insufficient: DOs cannot retroactively embed artifacts into already-leaked data, nor can they compel malicious trainers to provide cryptographic proofs. Consequently, DOs are forced to rely on post-hoc audits, typically Membership Inference Attacks (MIAs)~\cite{mancera2025MINT}. Yet, MIAs primarily exploit record-level likelihood signals tied to $n$-gram overlap, rather than corpus-level data usage, yielding high error rates when adversaries scrub verbatim traces via paraphrasing~\cite{zhang2025position, duan2024mia-not-work}. This leaves proprietary data exposed in fine-tuning infringement without a viable verification capability.


To break this forensic deadlock, we propose \textit{Distribution Provenance Audit (DPA)}, a post-hoc framework that fundamentally shifts the audit paradigm from detecting record-level signals to verifying distribution-level alignment. DPA is grounded in a critical insight: even if malicious trainers scrub verbatim traces to remove watermarks or evade MIAs, they are bound by a rational \textit{utility constraint}: to maintain expert-level utility, the model \textit{must} internalize the intrinsic statistics of the proprietary data. As illustrated in Figure~\ref{fig:DPA-schema}, we identify these resilient traces as \textit{intrinsic distributional fingerprints}, manifested at the robust intersection of semantic substance (e.g., medical logic) and lexical form (e.g., terminological preferences). Accordingly, DPA formulates the provenance audit as a rigorous statistical hypothesis test: by quantifying these fingerprints via representative output sampling, DPA reliably rejects the null hypothesis of non-usage, defeating evasion tactics like paraphrasing and distillation that circumvent traditional forensics.

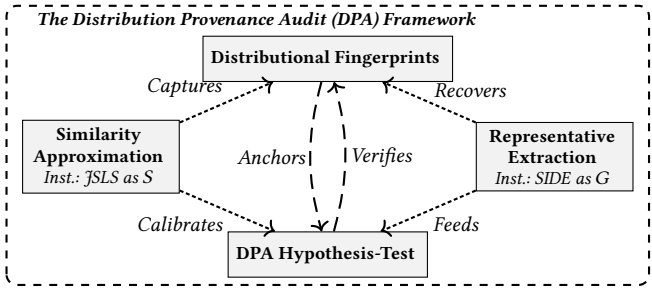
\begin{figure}[t!]
\centering
\footnotesize
\begin{tikzcd}[
  column sep=0.3cm,
  row sep=0.5cm,
  cells={nodes={draw, minimum width=2cm, minimum height=0.6cm, align=center}},
  arrows={->, dash pattern=on 1.5pt off 1pt, thick},
  execute at begin picture={
    \node (problem) at (1,1) {};
    \node (JSLS) at (0,0) {};
    \node (SIDE) at (2,0) {};
    \node (solution) at (1,-1) {};
    \draw[dashed, thick, rounded corners] (-4.25,-1.7) rectangle (4.25,2.0) 
      node[above right] at (-3.95, 1.6) {\textbf{\textit{The Distribution Provenance Audit (DPA) Framework}}};
  }
]
& |[fill=gray!10,alias=problem]| \textbf{Distributional Fingerprints}  \\
|[fill=gray!10,alias=JSLS]| \parbox{1.8cm}{\centering \textbf{Similarity \\ Approximation \\ } \textit{Inst.: JSLS as $S$}} 
  \arrow[from=JSLS, to=problem, "\textit{\small Captures}"]
  \arrow[from=JSLS, to=solution, "\textit{\small Calibrates}", swap]
  \arrow[from=solution, to=problem, "\textit{\small Verifies}", swap, dashed, bend right=15] 
  \arrow[from=problem, to=solution, "\textit{\small Anchors}", swap, dashed, bend right=15] &
& |[fill=gray!10,alias=SIDE]| \parbox{1.8cm}{\centering \textbf{Representative\\  Extraction\\ } \textit{Inst.: SIDE as $G$}} \\
& |[fill=gray!10,alias=solution]| \textbf{DPA Hypothesis-Test}
  \arrow[from=SIDE, to=problem, "\textit{\small Recovers}", swap]
  \arrow[from=SIDE, to=solution, "\textit{\small Feeds}"] &
\end{tikzcd}
\caption[DPA logical framework]{The logical architecture of the DPA framework. 
}
\label{fig:dpa-logics}
\end{figure}

Translating this distributional insight into a rigorous audit, however, presents two distinct methodological hurdles. First, existing similarity metrics fail to correctly distinguish proprietary distributions: surface-level metrics (e.g., BLEU~\cite{papineni2002bleu}) are easily circumvented by paraphrasing, while purely semantic ones (e.g., MAUVE~\cite{pillutla2021mauve}) suffer from conflating generic domain knowledge with specific provenance. To resolve this, we introduce \textit{Joint Semantic-Lexical Similarity (JSLS)}. This unified metric robustly identifies proprietary fingerprints by enforcing a strict dual constraint where semantic consistency is rigorously validated by lexical preference. Second, accessing the model's latent distribution under black-box constraints is non-trivial, as external prompts inevitably introduce prompt-induced bias, distorting the provenance signal. We address this via \textit{Self-Instruct Data Extraction (SIDE)}, a faithful, prompt-agnostic extraction algorithm. SIDE bootstraps from the model’s intrinsic high-likelihood regions to reconstruct its latent distribution without external interference. Figure~\ref{fig:dpa-logics} illustrates the logical architecture of DPA, where JSLS and SIDE collaborate to anchor the statistical hypothesis test upon the extracted distributional fingerprints.

Extensive evaluations on medical and legal fine-tuning tasks validate DPA’s effectiveness. First, regarding metric validity, we demonstrate that JSLS reliably captures intrinsic fingerprints, establishing a decisive separation boundary that accurately reflects ground-truth provenance between proprietary and independent sources. Second, regarding audit robustness, the SIDE-based DPA solution remains resilient: even when a malicious trainer employs data paraphrasing or knowledge distillation, DPA consistently validates provenance claims with high statistical confidence. By anchoring the audit to intrinsic fingerprints, our framework outperforms traditional baselines, confirming that the utility constraint effectively forces the retention of identifiable distributional fingerprints.

Beyond reliable provenance, our work uncovers a fundamental dual-use tension in AI auditing. We demonstrate that the capability to verify distribution provenance is structurally \textit{isomorphic} to performing distribution-targeted privacy attacks. The same high-fidelity fingerprints that enable DPA can be weaponized to facilitate data extraction~\cite{carlini2021data-extraction, lukas2023pii-leakage} and distribution inference attacks~\cite{hartmann2023distribution-inference}. This finding highlights a critical transparency paradox: the high-fidelity evidence required to substantiate Data IP claims inevitably collides with data privacy boundaries. Consequently, DPA not only serves as a robust forensic tool but also highlights the inherent tension between necessary auditability and unintended leakage.

To conclude, our key contributions are:

\begin{itemize}
    \item We propose Distribution Provenance Audit (DPA), the first work utilizing \textit{intrinsic distributional fingerprints} for Data IP audit in black-box, malicious LLM fine-tuning.
    \item We analyze quantifiable provenance distinctions via \textit{Joint Semantic-Lexical Similarity (JSLS)}, which integrates lexical and semantic features as distributional fingerprints.
    \item We recover faithful provenance signals via \textit{Self-Instruct Data Extraction (SIDE)}, which reconstructs the model's latent distribution while mitigating prompt-induced biases.
    \item Extensive experiments under a strict threat model validate DPA's effectiveness and reliability, while revealing a dialectical tension between AI transparency and privacy leakage.
\end{itemize}




\section{Related Works and Motivation}
\label{sec2:related_works_motivation}

\subsection{Data Provenance in LLM Fine-Tuning}

Distinct from the prevailing focus on model IP protection~\cite{xu2025copyright}, this paper addresses the under-explored challenge of auditing data provenance in LLM fine-tuning. Existing methods divide into two paradigms: invasive interventions and post-hoc audits. Invasive interventions, such as dataset watermarking~\cite{qiu2025icassp-watermarking,tang2023backdoor}, inject verifiable artifacts into training corpora; however, they compromise data utility and are inapplicable to pre-existing datasets. Meanwhile, Zero-Knowledge Proofs (ZKPs)~\cite{sun2024zkllm} cryptographically bind models to data but incur prohibitive computational costs and require trainer cooperation, rendering them ineffective with malicious trainers. Post-hoc audits, notably Membership Inference Attacks (MIAs)~\cite{mancera2025MINT}, analyze model signals (e.g., gradients) to infer usage. However, they are statistically unsound for provenance, suffering from high error rates~\cite{zhang2025position}. In contrast, aligning with the emerging paradigm of intrinsic fingerprinting in model copyright protection~\cite{xu2025copyright}, we propose \textbf{Distribution Provenance Audit (DPA)} for LLM fine-tuning, a post-hoc framework solely via intrinsic distributional fingerprints to audit data usage with a malicious trainer. See Appendix~\ref{app:supply_chain_discussion}, which further positions DPA within the LLM supply chain.


\subsection{Research Motivation}

This paper advocates a paradigm shift to distribution analysis for data provenance in LLM fine-tuning. We illustrate this necessity through a realistic copyright dispute scenario characterized by a malicious trainer conducting deliberate \textit{provenance evasion}.

Consider a Data Owner (DO), such as \textit{HealthCareMagic} (HCM)~\cite{li2023chatdoctor}, an online clinic who possesses a proprietary dataset of real patient-physician conversations. A malicious trainer unauthorizedly fine-tunes a Qwen3-4B model~\cite{yang2025qwen3} using LoRA~\cite{hu2022lora} on a 7K subset of this data (denoted as $D^\text{HCM}_\text{base}$). To evade infringement allegations, the trainer falsely claims the model was trained solely on synthetic, legally distinct conversations generated by ChatGPT ($D^\text{GPT}$)~\cite{achiam2023gpt}.
Crucially, this creates a forensic deadlock where invasive provenance fails: the DO cannot retroactively embed watermarks into the already-leaked legacy data, and the uncooperative trainer refuses to provide ZKPs or internal training logs. Consequently, the DO is forced to conduct a strict \textit{post-hoc} audit relying solely on black-box model outputs to break this ``he-said, she-said'' legal stalemate.

We assume the trainer employs deliberate \textit{provenance evasion countermeasures} designed to suppress identifiable record-level traces while preserving downstream data utility, including three tactics:

\textbf{Tactic 1: Subset Split (T1).} The trainer fine-tunes on a subset $D^\text{HCM}_\text{base}$. The DO, lacking knowledge of the specific split, is forced to audit with a reference $D^\text{HCM}_\text{split}$, making exact matching impossible because it is unaffordable to consider all feasible split ways.

\textbf{Tactic 2: Data Paraphrasing (T2).} To evade verbatim detection, the trainer rewrites $D^\text{HCM}_\text{base}$ into $D^\text{HCM}_\text{para}$. This retains the semantic value and privacy attributes of the original conversations but destroys verbatim matching. See Appendix~\ref{app:paraphrasing_details} for details.

\textbf{Tactic 3: Knowledge Distillation (T3).} The trainer first fine-tunes a teacher model on $D^\text{HCM}_\text{base}$, and then trains the final released model via knowledge distillation using the decoy distribution $D^\text{GPT}$. This transfers the proprietary knowledge while aligning the student model's surface statistics with the decoy data.

\begin{table}[t!]
\centering
\caption{Failure of existing MIA-based passive provenance methods under adversarial fine-tuning tactics.}
\label{tab:motivation-failure}
\resizebox{\linewidth}{!}{
\begin{tabular}{cc|cccc}
\toprule
\multicolumn{2}{c|}{\textbf{Metric \& Target}} & \textbf{\begin{tabular}[c]{@{}c@{}}No Tactic\\ (Baseline)\end{tabular}} & \textbf{\begin{tabular}[c]{@{}c@{}}Tactic 1\\ (Subset)\end{tabular}} & \textbf{\begin{tabular}[c]{@{}c@{}}Tactic 2\\ (Paraphrase)\end{tabular}} & \textbf{\begin{tabular}[c]{@{}c@{}}Tactic 3\\ (Distill)\end{tabular}} \\ \midrule
\multicolumn{1}{c|}{\multirow{2}{*}{\textbf{\begin{tabular}[c]{@{}c@{}}Avg. MIA Score\\ (Lower is Member)\end{tabular}}}} & \cellcolor{gray!10}Mem. ($D^\text{HCM}$) & \cellcolor{gray!10}0.028 & \cellcolor{gray!10}0.090 & \cellcolor{gray!10}0.411 & \cellcolor{gray!10}0.198 \\
\multicolumn{1}{c|}{} & Non-Mem. ($D^\text{GPT}$) & \textbf{-0.017} & \textbf{-0.043} & \textbf{0.115} & \textbf{-0.020} \\ \midrule
\multicolumn{2}{c|}{\textbf{Provenance Outcome}} & \color{red}\textbf{\xmark\ False} & \color{red}\textbf{\xmark\ False} & \color{red}\textbf{\xmark\ False} & \color{red}\textbf{\xmark\ False} \\ \bottomrule
\end{tabular}}
\end{table}

Compelled to rely on post-hoc audit, we examine the state-of-the-art Reference-based MIA~\cite{carlini2021data-extraction, zhu2024privauditor}. However, Table~\ref{tab:motivation-failure} reveals a systematic collapse of this baseline. Contrary to the theoretical premise that members exhibit lower MIA scores, we observe a consistent metric mismatch: members ($D^\text{HCM}$) yield higher scores than non-members ($D^\text{GPT}$) across all settings with evasion tactics (e.g., 0.411 vs. 0.115 under T2). This creates a false attribution paradox, where the model appears ``innocent'' of using the stolen data. We attribute this failure to the \textit{inherent structural simplicity} of the synthetic non-members~\cite{duan2024mia-not-work, zhang2025position}: compared to the complex, real-world $D^\text{HCM}$, the synthetic $D^\text{GPT}$ samples naturally yield higher likelihoods, thus confounding the membership signal.
\begin{table}[t!]
\caption{\textbf{Persistence of distributional signatures.} Top-8 words ranked by relevance score (Equ.~\ref{equ:word-relevance}). Distinctive words (green) persist in model outputs despite adversarial tactics (T1-T3).}

\label{tab:distributional-signatures}
\resizebox{\linewidth}{!}{
\begin{tabular}{c|l}
\toprule
\textbf{Source} & \multicolumn{1}{c}{\textbf{Top-8 Distinctive Words (Lexical Manifestation of Signatures)}} \\ \midrule
\textbf{HCM (Proprietary)} & \colorbox[rgb]{0.6, 1, 0.6}{Chat}, \colorbox[rgb]{0.6, 1, 0.6}{query}, \colorbox[rgb]{0.6, 1, 0.6}{Doctor}, \colorbox[rgb]{1, 0.6, 0.6}{your}, \colorbox[rgb]{0.6, 1, 0.6}{infection}, \colorbox[rgb]{0.6, 1, 0.6}{consult}, \colorbox[rgb]{1, 0.6, 0.6}{symptoms}, \colorbox[rgb]{0.6, 1, 0.6}{ChatDoctor} \\ \hline
\textbf{GPT (Synthetic)} & \colorbox[rgb]{1, 0.6, 0.6}{symptoms}, nlt, nln, \colorbox[rgb]{1, 0.6, 0.6}{your}, Hello, medication, doctor, Regarding \\ \midrule
\textbf{T1 (Subset)} & \colorbox[rgb]{0.6, 1, 0.6}{Chat}, \colorbox[rgb]{0.6, 1, 0.6}{query}, \colorbox[rgb]{0.6, 1, 0.6}{Doctor}, \colorbox[rgb]{0.6, 1, 0.6}{ChatDoctor}, \colorbox[rgb]{1, 0.6, 0.6}{symptoms}, \colorbox[rgb]{0.6, 1, 0.6}{consult}, etc, \colorbox[rgb]{0.6, 1, 0.6}{infection} \\ \hline
\textbf{T2 (Paraphrase)} & \colorbox[rgb]{0.6, 1, 0.6}{Chat}, \colorbox[rgb]{0.6, 1, 0.6}{Doctor}, \colorbox[rgb]{0.6, 1, 0.6}{ChatDoctor}, \colorbox[rgb]{0.6, 1, 0.6}{query}, \colorbox[rgb]{1, 0.6, 0.6}{symptoms}, Wishing, Regards, \colorbox[rgb]{0.6, 1, 0.6}{consult} \\ \hline
\textbf{T3 (Distillation)} & \colorbox[rgb]{0.6, 1, 0.6}{ChatDoctor}, \colorbox[rgb]{0.6, 1, 0.6}{query}, \colorbox[rgb]{0.6, 1, 0.6}{Chat}, \colorbox[rgb]{0.6, 1, 0.6}{Doctor}, Hello, \colorbox[rgb]{0.6, 1, 0.6}{consult}, \colorbox[rgb]{1, 0.6, 0.6}{symptoms}, \colorbox[rgb]{1, 0.6, 0.6}{your} \\ \bottomrule
\end{tabular}}
\end{table}

To break this forensic deadlock, we propose utilizing observable \textit{intrinsic distributional fingerprints}, leveraging the inherent semantic and lexical divergence between the proprietary dataset ($D^\text{HCM}$) and the ``smoke grenades'' distribution  ($D^\text{GPT}$). We characterize these fingerprints by three core properties: ) \textit{Inherence}: rooted in data’s intrinsic semantic and lexical signatures, not artificially implanted; 2) \textit{Distinguishability}: statistically diverging from other data distributions; 3) \textit{resilient}: persisting in model outputs even after malicious evasion countermeasures. Table~\ref{tab:distributional-signatures} partially confirms this resilience: distinctive $D^\text{HCM}$ lexical traces (words in green) persist in model outputs despite evasion tactics (T1--T3), sharply diverging from the synthetic decoy distribution ($D^\text{GPT}$) despite few overlaps (words in red). This observation forms the premise of the proposed Distribution Provenance Audit (DPA). Generalizing beyond the lexical persistence shown in Table~\ref{tab:distributional-signatures}, DPA captures joint semantic and lexical fingerprints to enable reliable provenance under adversarial conditions, as detailed in later sections.


\section{Problem Formalization and Threat Model}
\label{sec3:problem_formulation}

\subsection{Text Distribution and LLM Fine-tuning}

We formally define the provenance objects within a discrete textual space $\Phi$. A \textit{textual data distribution} $\pi(\cdot): \Phi \rightarrow [0,1]$ assigns a probability $\pi(s)$ to any sequence $s \in \Phi$. An observed dataset $D = \{s_1, \dots, s_n\}$ is treated as an i.i.d. sample from a target distribution $\pi_\text{own}$ from the Data Owner, with likelihood $\pi(D) = \prod_{s \in D} \pi(s)$.

Fundamentally, an LLM $M$ defines an \textit{induced distribution} $\pi_M$ over $\Phi$. The fine-tuning process is formulated as updating $M$ to minimize the forward Kullback-Leibler (KL) divergence $\mathbb{D}_\text{KL}(\pi_\text{own} || \pi_M)$~\cite{goodfellow2016deep}.
Crucially, minimizing this divergence leads $M$ to internalize the \textit{intrinsic distributional fingerprints} of $\pi_\text{own}$. Motivated by the distributional hypothesis~\cite{sahlgren2008distributional}, we characterize these fingerprints as an empirically grounded joint statistical dependency between semantic patterns (``what is said'') and lexical choices (``how it is said''). This optimization principle applies universally, whether in \textbf{Q\&A Mode} (aligning answer distributions conditional to given instructions) or \textbf{Completion Mode} (approximating joint sequence probabilities).


Since inference from $M$ constitutes sampling from $\pi_M$, the fine-tuning objective ensures that generated outputs inherently preserve the intrinsic fingerprints of $\pi_\text{own}$. This \textit{distributional alignment} principle redirects the audit focus: rather than relying on fragile record-level likelihoods, DPA quantifies the integrated statistical similarity between the proprietary distribution $\pi_\text{own}$ and the model's induced distribution $\pi_M$ via extracted samples.

\subsection{Provenance as a Distinguishability Game}

To apply the distributional alignment principle, we frame the provenance audit as a cryptographic \textit{distinguishability game} ($\textbf{Exp}^\text{DPA}$). This formulation captures the adversarial nature of the problem: the Auditor tries to detect traces of the proprietary distribution $\pi_\text{own}$ despite the Trainer's malicious attempts to mask them.

\begin{tcolorbox}[colframe=black, colback=gray!5, boxsep=2pt, left=6pt, right=6pt, top=4pt, bottom=4pt]
\textbf{The Audit Goal:} The Auditor aims to distinguish whether the black-box model $M$ was fine-tuned on a distribution derived from the proprietary $\pi_\text{own}$ (subject to evasion tactics) or solely from a statistically independent distribution $\pi_\text{ind}$.
\end{tcolorbox}

\begin{figure}
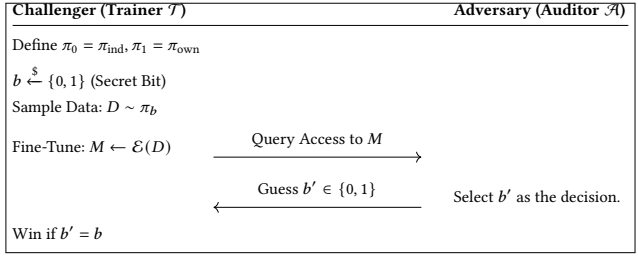

\centering
\resizebox{0.95\linewidth}{!}{%
    \begin{minipage}{1.2\linewidth} 
        \centering
        \begin{pcvstack}[center, boxed]
        \pseudocodeblock{
        \textbf{Challenger (Trainer \(\mathcal{T}\))} \> \> \textbf{Adversary (Auditor \(\mathcal{A}\))} \\[0.1\baselineskip][\hline] \<\<\\[-0.5\baselineskip]
        \text{Define } \pi_0 = \pi_\text{ind}, \pi_1 = \pi_\text{own} \> \> \\
        b \xleftarrow{\$} \{0, 1\} \text{ (Secret Bit)} \> \> \\
        \text{Sample Data: } D \sim \pi_b \> \> \\
        \text{Fine-Tune: } M \leftarrow \mathcal{E}(D) \> \sendmessageright*{ \text{Query Access to } M } \> \\
        \> \sendmessageleft*{ \text{Guess } b' \in \{0, 1\} } \> \text{Select } b' \text{ as the decision.}\\
        \text{Win if } b' = b \> \>
        }
        \end{pcvstack}
    \end{minipage}
}
\caption{Data Provenance Distinguishability Game (Exp$^\text{DPA}$). Auditor $\mathcal{A}$ distinguishes the data source through black-box queries to $M$, under an unknown fine-tuning mechanism $\mathcal{E}$ (standard or with evasive tactics).}
\label{fig:dpa_game}
\end{figure}
As shown in Figure~\ref{fig:dpa_game}, this game formalizes the conflict between the Challenger (Trainer $\mathcal{T}$) and the Adversary (Auditor $\mathcal{A}$), considering potential provenance evasion tactics. The Challenger samples a secret bit $b$. In the infringement case ($b=1$), the model is derived from the proprietary distribution via a fine-tuning mechanism $\mathcal{E}(\cdot)$, which potentially employs provenance evasion tactics—ranging from data obfuscation (e.g., data paraphrasing) to algorithmic modifications (e.g., knowledge distillation)—to confuse provenance. Conversely, if $b=0$, the model is derived solely from a statistically independent distribution $\pi_\text{ind}$. The Auditor's success in distinguishing these scenarios is quantified by the \textit{DPA Advantage}:
\begin{equation}
\label{equ:adv_dpa}
    \textbf{Adv}^\text{DPA}_{\mathcal{A}} = \left| \Pr[b'=1 | b=1] - \Pr[b'=1 | b=0] \right|.
\end{equation}
Since any negative correlation implies a distinguishable signal exploitable via strategy inversion, we focus on the positive $\textbf{Adv}^\text{DPA}_{\mathcal{A}}$. A non-negligible $\textbf{Adv}^\text{DPA}_{\mathcal{A}}$ thus serves as robust evidence that resilient distributional signatures have survived the evasion process $\mathcal{E}$, breaking the adversary's concealment.

\subsection{Threat Model}

We instantiate the $\textbf{Exp}^\text{DPA}$ game within a realistic adversarial threat model characterized by severe information asymmetry.

The malicious Trainer ($\textbf{Exp}^\text{DPA}$ Challenger) illicitly accesses a dataset $D_\text{base}\sim \pi_\text{own}$ and fine-tunes via $\mathcal{E}(\cdot)$, potentially employing provenance evasion tactics. However, this adversary is bound by a critical \textit{utility constraint}: to maintain fine-tuning utility, evasion tactics must preserve the essential semantic and lexical signatures of $D_\text{base}$. This inevitable leakage—the unavoidable cost of utility—serves as the foundation for the data provenance audit. To further confound detection, the trainer introduces functionally similar yet statistically independent distributions $\pi_\text{ind}$ as decoys, aiming to blur the boundaries of unauthorized data usage.

In contrast, the Data Owner ($\textbf{Exp}^\text{DPA}$ Adversary) acts as an provenance Auditor in a strictly post-hoc, black-box manner. We strictly restrict the Auditor to zero access regarding parameters, gradients, or logits, which precludes standard MIAs. Oblivious to the specific evasion tactics employed, and armed only with a reference dataset $D_\text{ref}$ sampled from either proprietary $\pi_\text{own}$ or independent $\pi_\text{ind}$, the auditor must break the concealment solely via inference API queries, identifying resilient distributional fingerprints for data provenance that survive the Trainer's utility-preserving evasion.
\section{Distribution Provenance Audit (DPA)}
\label{sec4:dpa_framework}

\subsection{From Probability to Similarity}

To instantiate the $\textbf{Exp}^\text{DPA}$ game under adversarial settings, we propose the \textbf{Distribution Provenance Audit (DPA)} framework. Theoretically, resolving this game hinges on \textit{probabilistic inference}: an optimal Auditor decides via the posterior probability $\Pr[b=1|M]$, while a rational Trainer preserves utility by minimizing the divergence $\mathbb{D}_\text{KL}(\pi_\text{own} || \pi_M)$~\cite{goodfellow2016deep} despite evasion tactics.

However, direct execution in this probabilistic space is intractable. LLM-induced distributions are fundamentally \textit{implicit}~\cite{mohamed2016implicit-learning}, prohibiting explicit likelihood evaluation. Although output logits are frequently employed as surrogates to approximate this likelihood (e.g., in MIAs), such proxies prove unreliable for robust provenance~\cite{zhang2025position}. Besides this difficulty, characterizing high-dimensional text distributional differences remains a complex open challenge~\cite{zhong2022describing}.

To bridge this operational gap, we adopt a dataset proxy strategy underpinned by Empirical Likelihood (EL) theory~\cite{owen2001empirical-likelihood}. EL circumvents explicit parameterization by modeling the unknown distribution $\pi$ as a non-parametric empirical measure over a reference dataset $D_\text{ref} \sim \pi$. This proxy strategy maps the intractable \textit{probabilistic} objectives of $\textbf{Exp}^\text{DPA}$ directly onto a computable statistical \textit{similarity} space. For the Auditor, estimating $\Pr[b=1|M]$ reduces to quantifying the similarity between model outputs $D_\text{ext}$ and a reference $D_\text{ref}$. Driven by the retention of $\pi_\text{own}$'s fingerprints during fine-tuning, this similarity is significant when $D_\text{ref} \sim \pi_\text{own}$, but relatively low for independent distributions ($D_\text{ref} \sim \pi_\text{ind}$). Symmetrically, a rational Trainer is compelled by utility constraints to keep the effective training distribution after obfuscation statistically similar to $\pi_\text{own}$ to preserve fidelity and utility; this similarity extends to knowledge distillation, where theoretical convergence to the teacher's solution space is guaranteed given sufficient data~\cite{phuong2019towards}.

Hence, to formalize this proxy strategy, we introduce the concept of \textit{Approximate Similarity}. Let $\mathcal{D}$ be the universe of textual datasets. We define an abstract similarity metric $\Delta: \mathcal{D} \times \mathcal{D} \to [0,1]$.

\begin{definition}[Dataset-Dataset $(\omega, \Delta)$-Approximate Similarity]
\label{def:dataset_sim}
Two datasets $D_A$ and $D_B$ are $(\omega, \Delta)$-approximately similar if their similarity score satisfies $\Delta(D_A, D_B) \geq \omega$.
\end{definition}

This serves as a computable proxy for intractable probabilities, where $\Delta$ denotes an abstract similarity estimator (e.g., BLEU~\cite{papineni2002bleu}) and $\omega$ the decision threshold. Crucially, a suitable $(\omega, \Delta)$ pair anchors this approximation to physical fidelity: $\Delta \ge \omega$ statistically certifies a high probability of indistinguishable distributional origins under rational utility preservation.




\subsection{DPA with Hypothesis Testing}

With the similarity proxy, we formulate the Auditor's maximization of $\textbf{Adv}^\text{DPA}_{\mathcal{A}}$ as a statistical hypothesis test, as shown in Figure~\ref{fig:dpa-logics}. 

First, we formally define the theoretical audit target. Given a suitable configuration $(\omega, \Delta)$, the ground-truth bit $b=1$ corresponds to the condition $\Delta(D_\text{trn}, D_\text{own}) \geq \omega$, where $D_\text{own}\sim\pi_\text{own}$ and $D_\text{trn}$ denotes the real fine-tuning datasets derived from $D_\text{base}\sim\pi_\text{own}$. Specifically in knowledge distillation, $D_\text{trn}$ is defined as the \textit{teacher model}'s training data (i.e., $D_\text{base}$), premised on the student inheriting intrinsic fingerprints from the teacher's source~\cite{phuong2019towards}.

However, the Auditor cannot directly verify $b'$ with the inaccessible the real fine-tuning data $D_\text{trn}$. Instead, the Auditor probes the model $M$ as a generative sampler $\pi_M$, employing an \textit{extraction function} $\mathcal{G}: M \to \mathcal{D}$ to project the implicit model distribution back into the explicit dataset space. The feasible hypothesis test is thereby formulated to yield the auditor's decision $b'$:
\begin{equation}
\label{equ:hypothesis}
\begin{cases}
H_0: & \Delta(\mathcal{G}(M),D^{\mathrm{own}})\le \tau,
       \quad \text{insufficient evidence of data usage},\\
H_1: & \Delta(\mathcal{G}(M),D^{\mathrm{own}})> \tau,
       \quad \text{evidence of data usage},
\end{cases}
\end{equation}
where $\tau$ is an empirical audit threshold. The Auditor outputs
$b'=1$ only when $H_0$ is rejected with sufficient confidence; otherwise,
it reports insufficient evidence of usage. Specifically, we use corpus-level
bootstrap over $\mathcal{G}(M)$ and $D^{\mathrm{own}}$, and reject
$H_0$ iff the 95\% Lower Confidence Bound ($\operatorname{LCB}_{0.95}$) remains
above the threshold, $\operatorname{LCB}_{0.95}(\Delta(\mathcal{G}(M),D^{\mathrm{own}}))>\tau$.
This proxy is justified by fine-tuning alignment between $\pi_M$ and
$\pi_{\mathrm{own}}$~\cite{goodfellow2016deep}. Crucially, we calibrate $\tau \le \omega$ for information-bottleneck attenuation~\cite{tishby2000information}: extracted fingerprints are weaker than those in the original training distribution.

\section{Joint Semantic-Lexical Similarity}
\label{sec5:measurement}

\begin{figure}[t!]
    \centering
        \begin{minipage}{0.48\linewidth}
    		\centering
    		\includegraphics[width=\linewidth]{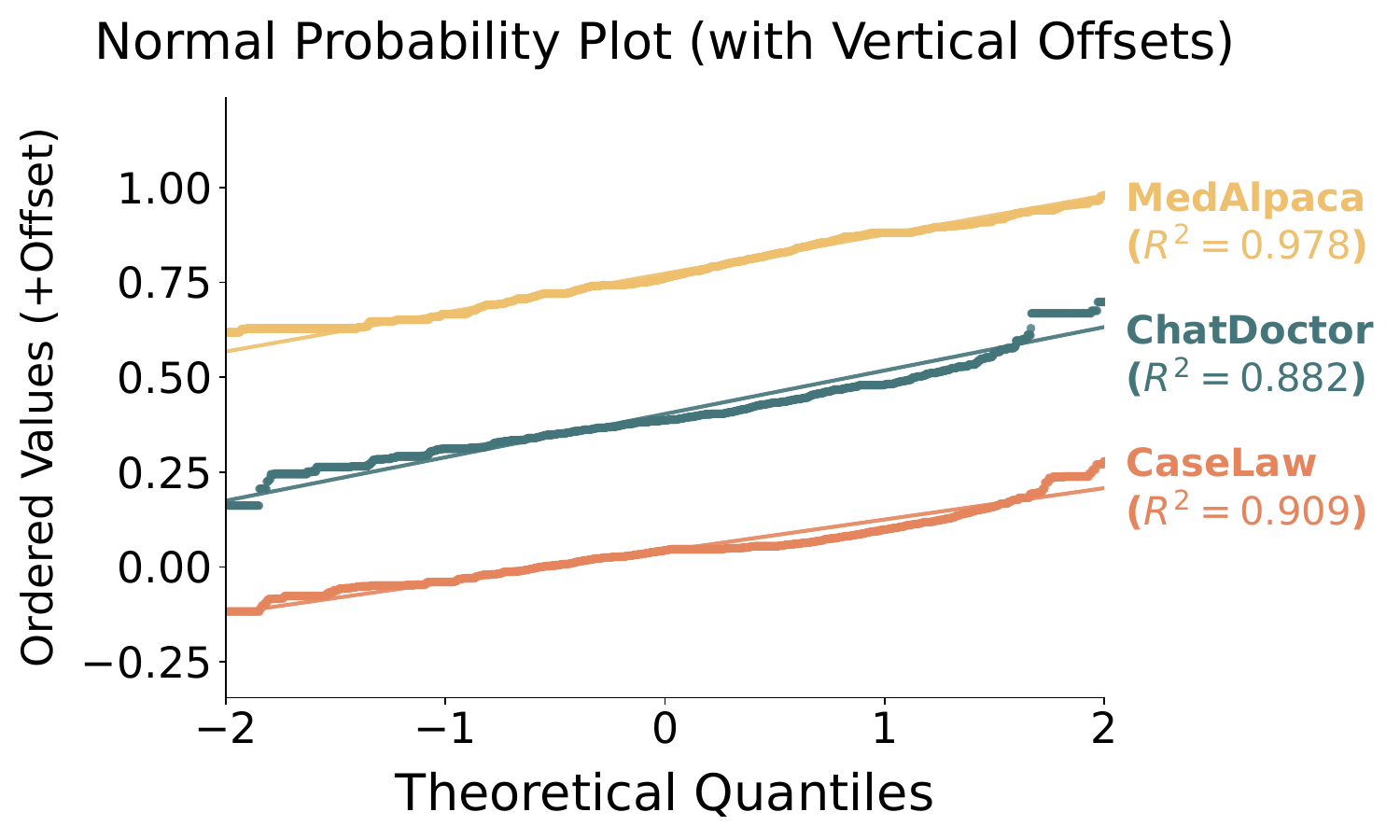}
    	\end{minipage}
        \begin{minipage}{0.48\linewidth}
    		\centering
    		\includegraphics[width=\linewidth]{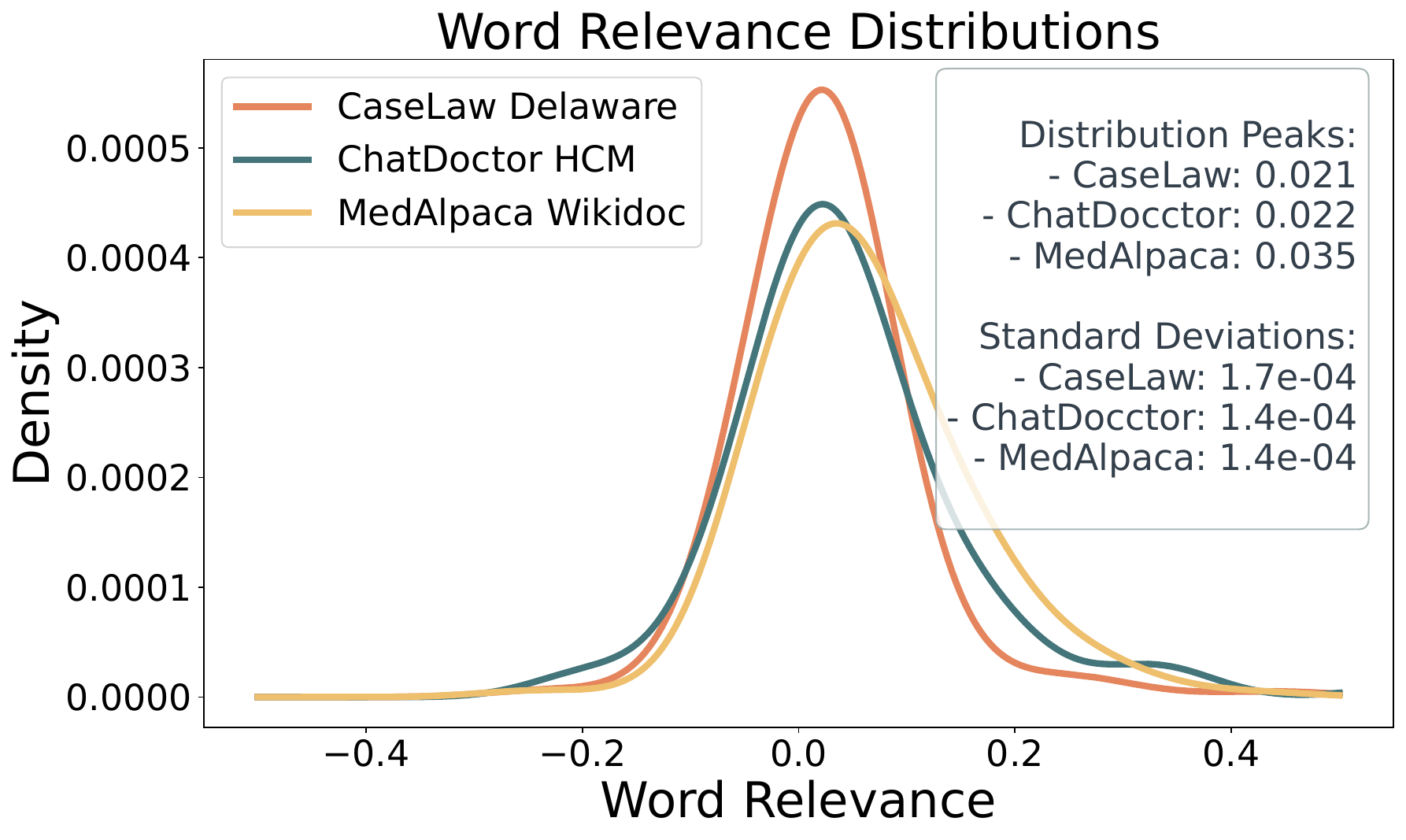}
    	\end{minipage}
    \caption{Normal probability plots~\cite{looney1985normal-probability-plots} of word relevance distributions across different datasets and morphological differentiation in fitted relevance distributions.}
    \label{fig:relevance-distribution}
\end{figure}

\begin{figure*}[tpbh] 
  \centering
  \includegraphics[width=\textwidth]{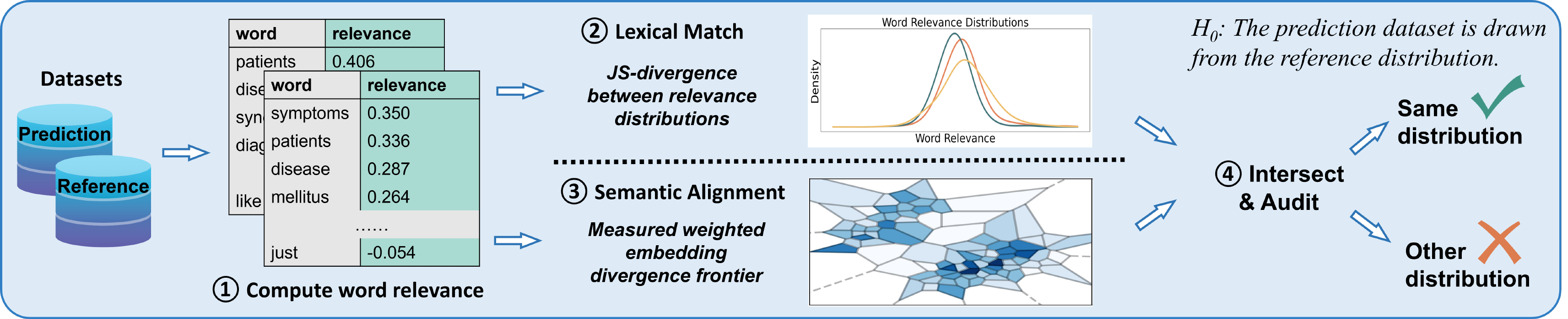} 
  \caption{Illustration of \textbf{Joint Semantic-Lexical Similarity (JSLS)}: 1) \textbf{word relevance scores}, measuring word distributional significance through frequency deviations; 2) \textbf{lexical match}, derived from the JS-divergence between datasets' word relevance distributions. 3) \textbf{semantic alignment}, computed via word relevance-weighted divergence frontiers to capture semantic. JSLS synthesizes these into a joint metric and captures intrinsic distributional fingerprints.}
  \label{fig1:JSLS}
\end{figure*}

Feasible DPA requires instantiating $\Delta$ with a suitable metric to balance between robustness and sensitivity. Existing methods fail to strike this balance: surface-level metrics (e.g., BLEU~\cite{papineni2002bleu}) yield false negatives by misidentifying obfuscated proprietary data as distinct distributions despite their retained utility; while pure semantic methods (e.g., MAUVE~\cite{pillutla2021mauve}) suffer from false positives by erroneously conflating independent datasets from the same domain. We thus propose \textbf{Joint Semantic-Lexical Similarity (JSLS)} to address this limitation. By fusing semantic consistency with lexical preference, JSLS robustly recovers the \textit{intrinsic distributional fingerprints} that persist even under malicious provenance evasions.

\subsection{Word Relevance and Lexical Divergence}

JSLS begins by identifying the word as the fundamental intersection of semantic substance and lexical form~\cite{sahlgren2008distributional}, forming a basic building block of distributional fingerprints. Specifically, JSLS defines \textit{word relevance} via frequency divergence, identifying provenance by detecting disproportionate use of semantically significant terms (e.g., clinical phrases) in a proprietary distribution relative to general language, capturing distinctive distributional fingerprints.

We first compute word frequency $f_D(w)$ in dataset $D$ via Laplace smoothing to mitigate the zero-frequency problem, ensuring valid non-zero probability estimates for rare or unseen tokens:
\begin{equation}
    \label{equ:frequency-dataset}
    f_D(w) = \frac{\text{count}(w)+1}{\sum_{w'\in D} \text{count}(w') + |\{w|w\in D\}|},  
\end{equation}  
where $|\{w|w\in D\}|$ is $D$'s vocabulary size. 

Instead of the standard KL-divergence, we formulate a \textit{variance-stabilized frequency divergence} to address the inherent heteroscedasticity of textual data. 
Modeling word occurrences as a Poisson process~\cite{church1995poisson}, where variance scales with the mean, reveals two statistical problems: (1) \textit{Shot Noise}, where the inherent variance of rare words destabilizes the log-ratio; and (2) \textit{Magnitude Dominance}, where frequent words exert disproportionate linear influence.
\begin{equation}
    \label{equ:word-relevance}
    r(w) = \underbrace{\sqrt{f_D(w)}}_{\text{Variance Stabilizer}} \cdot \underbrace{\log \frac{f_D(w)}{f_L(w)}}_{\text{Information Gain}}.  
\end{equation}  
By modulating the information gain with the variance-stabilized square root of frequency, $r(w)$ effectively suppresses the shot noise of rare words while dampening the magnitude dominance of common words, extracting robust distributional fingerprints.

Aggregating the atomic fingerprints, we characterize lexical divergence via the empirical distribution of word relevance scores. As shown in Figure~\ref{fig:relevance-distribution} (left), these distributions exhibit a pronounced linear trend in normal probability plots ($R^2 \approx 0.9$), indicating that their central mass is well-approximated by a Gaussian form despite inherent tail deviations. Crucially, rather than assuming strict normality, we leverage this approximation as a stable parametric proxy. By employing robust estimates of location and scale, we compress complex lexical disparities between datasets (e.g., the dense terminology of textbooks vs. the dispersed vocabulary of general corpora) into distinguishable parameter shifts, enabling lexical discrimination of data origins as shown in Figure~\ref{fig:relevance-distribution} (right).


Formally, we model the lexical signature of a dataset $D$ as a robust Gaussian $\mathcal{N}^R_D(\mu^R_D, (\sigma^R_D)^2)$ over its relevance scores $R_D = \{r(w) | w \in D\}$. To mitigate the impact of extreme frequency count outliers, we employ robust statistics for parameter estimation: centrality $\mu^R_D$ is estimated via the median, and dispersion $\sigma^R_D$ via the scaled Interquartile Range (IQR)~\cite{wan2014iqr}, where $Q_k$ denotes the $k- \text{percentile}$:
\begin{equation}
    \mu^R_D = \mathrm{Median}(R_D), \quad \sigma^R_D = \frac{Q_{75}(R_D) - Q_{25}(R_D)}{1.349}.
\end{equation}
Consequently, the lexical divergence $\gamma$ between datasets $D_0$ and $D_1$ is defined as the Jensen–Shannon (JS) divergence between their respective parametric estimated distributions:
\begin{equation}
    \label{equ:relevance-js}
    \gamma(D_0, D_1) = \mathrm{JS}(\mathcal{N}^R_0, \mathcal{N}^R_1).
\end{equation}
This approach captures distributional intrinsic lexical signatures, where lower $\gamma$ signifies stronger distributional lexical alignment.

\subsection{Relevance-Weighted Embeddings and Semantic Similarity} 

To look beyond lexical divergence, JSLS utilizes the word relevance $r(w)$ as a foundation to bridge granular statistics with high-level semantic analysis. While embeddings encode these semantics, directly measuring distributional shifts in such high-dimensional spaces is still challenging. Methods like MAUVE~\cite{pillutla2021mauve} address this by estimating \textit{Kullback-Leibler divergence frontiers} over quantized embedding cluster distributions. However, standard approaches treat all samples indiscriminately (uniform weights), failing to distinguish between generic logic (e.g., ``the results show'') and distribution-specific knowledge (e.g., ``patient exhibits arrhythmia'').

To bridge this gap, we integrate the atomic relevance $r(w)$ into the semantic approximation process. We define the \textit{sequence relevance score} $r(\mathbf{s})$ for a text sequence $\mathbf{s}$ by aggregating its constituent atomic relevances: $r(\mathbf{s}) = \sum_{w \in \mathbf{s}} r(w)$. This score serves as a measure of \textit{informational density}, allowing us to prioritize sequences that carry intrinsic distributional fingerprints over generic syntax.

Consequently, we formulate a \textit{relevance-weighted distribution approximation}. Unlike uniform clustering, which is susceptible to noise from high-frequency generic patterns, our approach re-weights the probability mass of each quantization cluster based on the accumulated relevance of its assigned sequences. Formally, for a dataset $D$, the weighted probability of cluster $j$ is given by:
\begin{equation}
    \label{equ:relevance-weighed-cluster}
    \widetilde{P}(j) = \frac{1}{\sum_{\mathbf{s}\in D} r(\mathbf{s})} \sum_{\mathbf{s}\in D} r(\mathbf{s}) \cdot \mathbb{I}(\phi(\mathbf{s}) = j),
\end{equation}
where $\phi(\mathbf{s})$ maps sequence $\mathbf{s}$ to its cluster index via K-Means~\cite{hartigan1979kmeans} on pre-trained encoder embeddings (e.g., Sentence-BERT~\cite{reimers2019sentence-bert}), $\mathbb{I}(\cdot)$ is the indicator function. By explicitly biasing the distribution $\widetilde{P}$ towards high-relevance regions, JSLS effectively mitigates the influence of out-of-distribution or generic noise.

Finally, utilizing the relevance-weighted semantic distributions $\widetilde{P}_0$ and $\widetilde{P}_1$ derived from datasets $D_0$ and $D_1$, JSLS computes the \textit{divergence curve} $\mathcal{C}$ and extracts the Area Under the Curve (AUC) as the final semantic similarity score $m_r$. This metric captures the alignment of intrinsic semantic signatures:
\begin{equation}
    \label{equ:mauve}
    \begin{aligned}
        & m_r(D_0, D_1) = \mathrm{AUC}\left(\mathcal{C}(\widetilde{P}_0, \widetilde{P}_1)\right), \quad \text{where} \\
        & \mathcal{C}(\widetilde{P}_0,\widetilde{P}_1) = \Big\{ \Big( e^{-c \cdot \mathrm{KL}(\widetilde{P}_0 \| R_\lambda)}, \, e^{-c \cdot \mathrm{KL}(\widetilde{P}_1 \| R_\lambda)} \Big) \Big| \\
        & \quad R_\lambda = \lambda \widetilde{P}_0 + (1-\lambda)\widetilde{P}_1,\ \lambda \in (0,1) \Big\},
    \end{aligned}
\end{equation}
where $c$ is a scaling factor and $R_{\lambda}$ represents the mixture distribution. A higher $m_r$ indicates that the datasets share not just superficial semantic overlap, but a deep, distributional semantic alignment.

\subsection{JSLS: A Unified Distribution Similarity Metric}

Thereby, drawing inspiration from the kernel trick~\cite{scholkopf2002learning}, \textbf{Joint Semantic-Lexical Similarity (JSLS)} metric is formulated by: 

\begin{equation}
    \label{equ:jsls}
    \Delta_{\text{JSLS}}(D_0, D_1) = \underbrace{m_r(D_0, D_1)}_{\text{Semantic Alignment}} / \underbrace{\exp\left(k \cdot \gamma(D_0, D_1)\right)}_{\text{Lexical Similarity}},
\end{equation}
where the exponential lexical term functions as a Laplacian kernel, calibrated by inverse bandwidth $k$. Going beyond simple aggregation, JSLS captures intrinsic distributional fingerprints through bridging semantic and lexical signatures via their shared word relevance foundation $r(w)$. Inspired by the gated product kernel~\cite{scholkopf2002learning}, JSLS enforces a strict logical ``AND'' constraint: the lexical term acts as a soft gate to mitigate false positives from generic semantic matches, surpassing methods based on a single aspect or ad-hoc ensembles. Therefore, satisfying \textit{non-negativity}, \textit{boundedness} ($\Delta_{\text{JSLS}} \in [0,1]$), and \textit{symmetry}, JSLS meets the requirement of DPA approximate similarity, as later empirically validated in Section~\ref{sec7:evaluations}.
\section{Representative Data Extraction and The Derived DPA Instantiation}
\label{sec6:extraction}

While JSLS provides the quantitative metric for both utility preservation and provenance verification, conducting the Auditor’s hypothesis test requires a critical component: the \textit{extraction function} $\mathcal{G}$. Crucially, the validity of this audit hinges on $\mathcal{G}$ faithfully recovering the latent model distribution $\pi_M$, since naive extraction using auxiliary datasets is flawed: conditional generation is inevitably skewed by the distribution of input queries~\cite{zhao2025LLM-watermark}, introducing significant bias. For instance, if the Auditor uses its own data ($D_\text{own}$) to prompt a model trained on an independent distribution $\pi_\text{ind}$, it can artificially induce outputs that mimic $\pi_\text{own}$. This prompt-induced mimicry generates systemic false positives and potentially accuses innocent models, thereby rendering the audit statistically invalid.

To overcome this, we propose \textbf{Self-Instruct Data Extraction (SIDE)} as a faithful sampling strategy to authentically probe $\pi_M$. To overcome the prompt bias inherent in auxiliary datasets, SIDE adopts a prompt-agnostic bootstrapping strategy. Inspired by self-instruct tuning~\cite{wang2022self-instruct}, the algorithm operates in two phases: it first elicits the model's dominant distributional modes using generic prompts, and subsequently expands coverage via embedding-based perturbation. This approach ensures that extracted samples are rooted strictly in $M$'s latent probability landscape, avoiding the extraction biases induced by external prompts.

\begin{figure}[t!] 
  \centering
  \includegraphics[width=\linewidth]{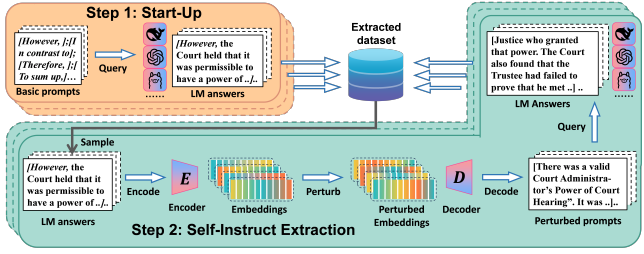} 
  \caption{Illustration of \textbf{Self-Instruct Data Extraction}: 1) \textbf{Start-UP}: Collect initial outputs with domain-agnostic prompts. (2) \textbf{Self-Instruct Refinement}: Perturbed extracted text embeddings and decoded back to diversify prompts. SIDE recovers distributional fingerprints without external interference.} 
  \label{fig2:extraction}
\end{figure}

\subsection{Details of Self-Instruct Data Extraction}

\subsubsection{Cold Start with Generic Prompts}

SIDE initiates extraction with a set of \textit{generic prompts} designed to be structurally functional but semantically neutral. 
Depending on the model type, SIDE employs distinct prompt strategies: for Q\&A models, blank tokens are used as dummy queries; for completion models, high-frequency conjunctive phrases (e.g., \textit{`however'}, \textit{`in conclusion'}) serve as prefixes. These generic prompts exploit the propensity of language models to generate continuations reflecting their training distribution after common prefixes~\cite{carlini2021data-extraction}. This phase results in a representative \textit{start-up dataset} $D_{\text{st}}$, capturing the model's high-likelihood intrinsic patterns without the contamination of topic-specific prompt biases.

\subsubsection{Iterative Self-Instruct Refinement}

While $D_{\text{st}}$ localizes the model's high-likelihood regions, the diversity of outputs generated by static prompts is inherently limited. To uncover the long-tail of the fine-tuning distribution, SIDE employs an iterative refinement process that perturbs extracted outputs in the embedding space to generate diverse yet distribution-grounded prompts.

Formally, in each iteration, SIDE samples a seed entry $\mathbf{s}$ from the current extracted dataset $D_{\text{ext}}$ (initialized as $D_{\text{st}}$). To induce semantic variation while preserving distributional representativeness, we map $\mathbf{s}$ to a dense vector $\mathbf{e} \in \mathbb{R}^d$ via an encoder, $\mathbf{e} = \mathrm{Enc}(\mathbf{s})$. 
We then inject controlled noise into this latent representation:
\begin{equation}
    \mathbf{e}' = \mathbf{e} + \alpha \cdot \frac{\bm{\delta}}{\|\bm{\delta}\|_2}, \quad \text{where } \bm{\delta} \sim \mathcal{N}(\bm{0}, \bm{I}_d).
\end{equation}
Here, $\bm{\delta}$ is a random direction vector sampled from a standard normal distribution. The scalar $\alpha \sim \mathcal{U}(\beta-\epsilon, \beta+\epsilon)$ controls the perturbation magnitude, where $\beta$ determines the base exploration radius and $\epsilon$ introduces bounded stochasticity. 

The perturbed embedding $\mathbf{e}'$ is subsequently mapped back to the discrete text space using a decoder, generating a candidate prompt $\mathbf{s}' = \mathrm{Dec}(\mathbf{e}')$. We implement $\mathrm{Dec}$ using inversion models like vec2text~\cite{morris2023vec2text}, followed by a surrogate LLM to restore coherence and syntactic integrity and adapt the format (e.g., converting declarative sentences into questions for Q\&A targets). Finally, the refined prompt $\mathbf{s}'$ is used to query the target model, yielding output $M(\mathbf{s}')$ which is incorporated into $D_{\text{ext}}$. By iteratively expanding $D_{\text{ext}}$, SIDE progressively maps the target model's fine-tuning distribution with high fidelity and diversity.

\subsection{The SIDE-Based DPA Instantiation and Privacy Discussions}

The abstract DPA framework (Section~\ref{sec4:dpa_framework}) can now be implemented as a concrete DPA instantiation. By instantiating the extractor $\mathcal{G}$ with the high-fidelity $\mathcal{G}_{\text{SIDE}}$ and the metric $\Delta$ via the reliable $\Delta_{\text{JSLS}}$, we transform the abstract design into an executable auditing procedure. Specifically, the Auditor first queries the black-box model $M$ via $\mathcal{G}_{\text{SIDE}}$ to extract a synthetic dataset, which serves as a faithful proxy for the intrinsic model distribution $\pi_M$. Subsequently, the Auditor computes the evidence score $s = \Delta_{\text{JSLS}}(\mathcal{G}_{\text{SIDE}}(M), D_{\text{own}})$ to quantify the preservation of proprietary fingerprints. Based on the hypothesis testing formulation (Eq.~\ref{equ:hypothesis}), the null hypothesis $H_0$ (Independent Source) is rejected if $s \ge \tau$. This solution effectively bridges the operational gap between intractable probabilistic inference and computable similarity, enabling robust data distribution provenance audit even under strict black-box constraints.

However, the very effectiveness of this solution introduces a dialectical tension: the capability to verify \textit{distribution provenance} is structurally \textit{isomorphic} to performing distribution-targeted \textit{privacy attacks}, as both necessitate recovering the same high-fidelity latent statistics from the black-box model. Specifically, SIDE's faithful reconstruction of intrinsic patterns can inadvertently extract memorized training samples, thereby facilitating \textit{data extraction attacks}~\cite{carlini2021data-extraction}. Concurrently, the distribution provenance integrating SIDE's latent probing with JSLS's alignment quantification can be exploited to conduct \textit{distribution inference}~\cite{hartmann2023distribution-inference}, or to synthesize optimal auxiliary datasets that amplify existing MIAs~\cite{fu2023self-prompt-mia}. This duality highlights a fundamental paradox in AI transparency: the intrinsic distributional fingerprints required to substantiate provenance claims inevitably touch upon the boundaries of privacy. By characterizing this structural conflict, our work not only equips Auditors with a robust verification tool but also illuminates the critical provenance-privacy tension for future secure auditing research.

\section{Evaluations}
\label{sec7:evaluations}

This section addresses three core research questions: (\textit{R1}) Does JSLS effectively capture intrinsic distributional fingerprints to faithfully reflect ground-truth data provenance? (\textit{R2}) How robustly does the SIDE-based DPA solution validate provenance claims under malicious evasion tactics? (\textit{R3}) To what extent does high-fidelity auditability correlate with potential privacy attacks?

Related codes and artifacts are publicly available at: \url{https://github.com/Huangzirui1206/DistributionProvenanceAudit}.

\subsection{Experimental Setup}

\subsubsection{Models and Finetuning Settings}
We evaluate DPA across two fine-tuning tasks: Question Answering (Q\&A) and Text Completion. To ensure generalizability, we employ three diverse pre-trained models: Pythia-2.8B~\cite{biderman2023pythia}, Qwen3-4B-Instruct~\cite{yang2025qwen3}, and Llama3-8B-Instruct~\cite{touvron2023llama}.
Specifically, we utilize Low-Rank Adaptation (LoRA)~\cite{hu2022lora} with rank $r=16$ and scaling factor $\alpha=32$. 


\begin{table}[t!]
\centering
\caption{Details of dataset processing. Abbreviations are surrounded by brackets.}
\label{tab:dataset-process-details}
\small
\resizebox{\linewidth}{!}{
\begin{threeparttable}
\begin{tabular}{@{}lll@{}}
\toprule
\textbf{Dataset} & 
\textbf{Proprietary Dist. $\pi_\text{own}$} & 
\textbf{Independent Dist. $\pi_\text{ind}$ for Decoy} \\
\midrule
MedAlpaca & Website \textit{wikidoc} ($D^\text{Wiki}$) \tnote{$\ast$}  & Website \textit{flashcards} ($D^\text{Flash}$)\\ 
ChatDoctor & Website \textit{healthcaremagic} ($D^\text{HCM}$)  & ChatGPT-generated ($D^\text{GPT}$)\\
CaseLaw & US state Delaware ($D^\text{Del}$) &  US state Arizona ($D^\text{Ariz}$) \\ 
\bottomrule
\end{tabular}
\begin{tablenotes}
\item[$\ast$] \footnotesize Because MedAlpaca Wikidoc only contains 10,000 samples total, we omit partitioning to preserve adequacy for evaluations.
\end{tablenotes}
\end{threeparttable}}
\end{table}

\subsubsection{Datasets} Our evaluation includes three benchmark datasets from various domains, as detailed in Table~\ref{tab:dataset-process-details}. For the Q\&A mode, we select two medical conversation datasets formatted in the Alpaca style~\cite{Taori2023alpaca}: \textit{MedAlpaca}~\cite{han2025medalcapa} and \textit{ChatDoctor}~\cite{li2023chatdoctor}. For text completion mode, we employ the US legal judgments dataset \textit{CaseLaw}~\cite{HFforLegal2024caselaw}.

Experiments simulate a dispute where the Trainer illicitly utilizes $D_\text{base}\sim\pi_\text{own}$ while alleging reliance on an independent distribution $\pi_\text{ind}$ as a decoy. DPA is evaluated to discern whether the unobservable training data $D_\text{trn}$ is genuinely independent or constitutes a provenance infringement. We evaluate DPA against three evasion tactics: (1) Subset Split (T1), utilizing a disjoint reference $D_\text{split}\sim\pi_\text{own}$ ($D_\text{base} \cap D_\text{split} = \emptyset$) to simulate the unavailability of exact records; (2) Data Paraphrasing (T2), fine-tuning on $D_\text{para}$, a semantically equivalent but lexically rewritten variant of $D_\text{base}$; and (3) Knowledge Distillation (T3), transferring proprietary knowledge using $D_\text{ind}\sim\pi_\text{ind}$ as the carrier dataset. Unless otherwise noted, each subset comprises 7,000 entries.

Detailed DPA implementation hyperparameters and ablation studies are provided in Appendix~\ref{app:setup} and Appendix~\ref{app:ablation}, respectively.

\subsection{Metric Validity Analysis}

\subsubsection{Effectiveness of JSLS Metric.}
Validating the alignment between $\Delta_{\text{JSLS}}$ and physical reality is critical to justify DPA's premise of using similarity as a proxy for probability. To this end, we first show that JSLS captures intrinsic distributional fingerprints. 

As shown in Table~\ref{tab:JSLS-validation}, JSLS reliably separates proprietary provenance from independent sources across diverse domains. Empirical results demonstrate a robust separation margin: proprietary derivatives (including paraphrased variants) consistently yield high similarity scores ($\Delta_{\text{JSLS}} > 0.7$, highlighted in green), whereas independent datasets fall distinctly into the low-similarity zone ($\Delta_{\text{JSLS}} < 0.6$, highlighted in red). This discriminative power stems from JSLS's dual modeling of the semantic-lexical correlated fingerprints. 

\begin{table}[t!]
\centering
\caption{JSLS quantifies similarity between $D_{\text{base}}\sim \pi_\text{own}$ and datasets from $\pi_\text{own}$ and $\pi_\text{ind}$, which can be distinguished via a threshold $\omega=0.6$.}
\label{tab:JSLS-validation}
\resizebox{0.9\linewidth}{!}{
\begin{tabular}{ccccc}
\toprule
  \textbf{Dist.}             & \textbf{Dataset}  & \textbf{Semantic} $\bm{(m_r)}$ & \textbf{Lexical} $\bm{(e^{k\gamma})}$ & \textbf{JSLS} ($\bm{\Delta_\textbf{JSLS}}$) \\ \midrule
\multicolumn{5}{c}{{\cellcolor[rgb]{0.9, 0.9,  0.9}}\textbf{MedAlpaca (compared with $D^\text{wiki}_\text{base}$, omitting superscript ``wiki'' later))}} \\
\hline
{\cellcolor[rgb]{0.6, 1,  0.6}}$\pi_\text{own}$ & $D\_\{\text{para}\}$ & 0.984 & 1.121 & {\cellcolor[rgb]{0.689, 1,  0.689}}\textbf{0.811} \\
{\cellcolor[rgb]{1, 0.6,  0.6}}$\pi_\text{ind}$ & $D$\textasciicircum$\{\text{flash}\}$ & 0.762 & 1.572 & {\cellcolor[rgb]{1, 0.785, 0.785}}\textbf{\textit{0.485}} \\ \midrule
\multicolumn{5}{c}{{\cellcolor[rgb]{0.9, 0.9,  0.9}}\textbf{ChatDoctor (compared with $D^\text{HCM}_\text{base}$, omitting superscript ``HCM'' later)}} \\
\hline
{\cellcolor[rgb]{0.6, 1,  0.6}}$\pi_\text{own}$ & $D\_\{\text{split}\}$ & 0.990 & 1.000 & {\cellcolor[rgb]{0.510, 1,  0.510}}\textbf{0.990} \\
{\cellcolor[rgb]{0.6, 1,  0.6}}$\pi_\text{own}$ & $D\_\{\text{para}\}$ & 0.994 & 1.029 & {\cellcolor[rgb]{0.534, 1,  0.534}}\textbf{0.966} \\
{\cellcolor[rgb]{1, 0.6,  0.6}}$\pi_\text{ind}$ & $D$\textasciicircum$\{\text{GPT}\}$ & 0.727 & 1.238 & {\cellcolor[rgb]{1, 0.713, 0.713}}\textbf{\textit{0.587}} \\ \midrule
\multicolumn{5}{c}{{\cellcolor[rgb]{0.9, 0.9,  0.9}}\textbf{CaseLaw (compared with $D^\text{Del}_\text{base}$, omitting superscript ``Del'' later)}} \\
\hline
{\cellcolor[rgb]{0.6, 1,   0.6}}$\pi_\text{own}$ & $D\_\{\text{split}\}$ & 0.853 & 1.000 & {\cellcolor[rgb]{0.647, 1,  0.647}}\textbf{0.853} \\
{\cellcolor[rgb]{0.6, 1,  0.6}}$\pi_\text{own}$ & $D\_\{\text{para}\}$ & 0.867 & 1.136 & {\cellcolor[rgb]{0.730, 1, 0.730}}\textbf{0.770} \\
{\cellcolor[rgb]{1, 0.6,  0.6}}$\pi_\text{ind}$ & $D$\textasciicircum$\{\text{Ariz}\}$ & 0.592 & 1.030 & {\cellcolor[rgb]{1, 0.725, 0.725}}\textbf{\textit{0.575}}
\\ \bottomrule
\end{tabular}}
\end{table}

\begin{figure}[t!] 
  \centering
  \includegraphics[width=\linewidth]{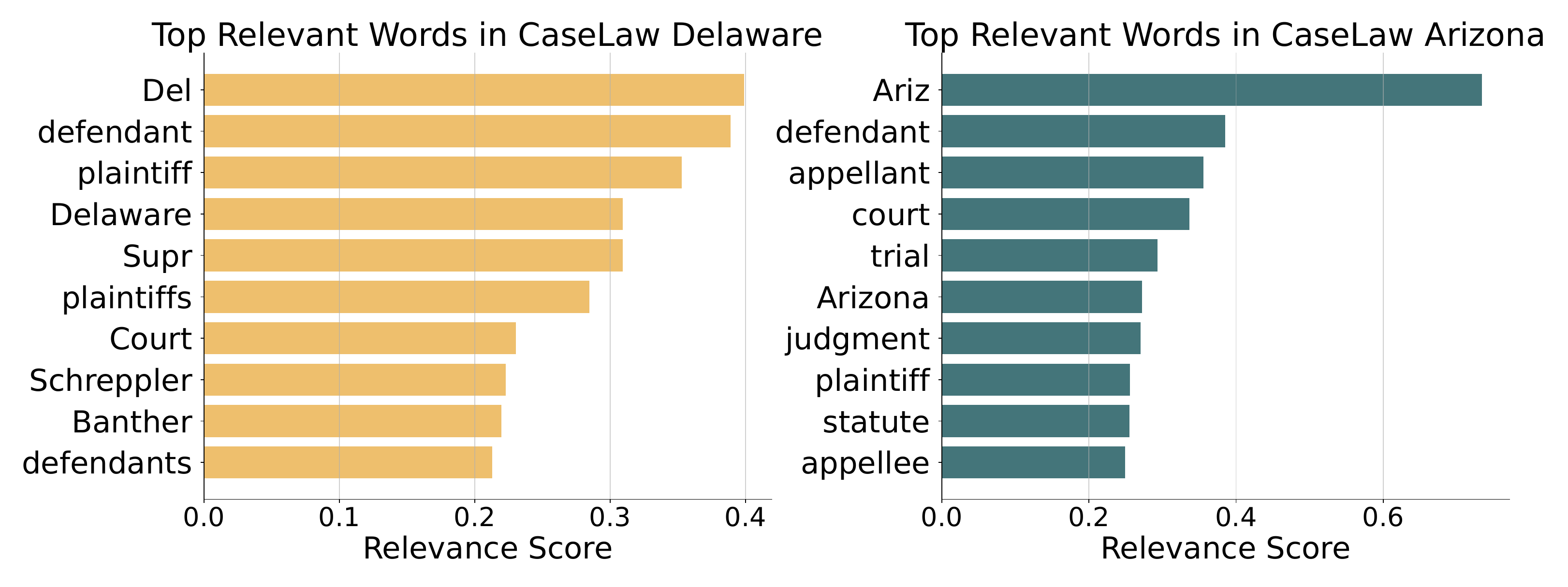} 
  \caption{The most relevant words across CaseLaw~\cite{HFforLegal2024caselaw} Delaware and Arizona, measured by word relevance scores.} 
  \label{fig:word-relevance-demo}
\end{figure}

The effectiveness of these intrinsic distributional fingerprints is founded on word relevance scores (Equation~\ref{equ:word-relevance}). Figure~\ref{fig:word-relevance-demo} shows the top relevant words in CaseLaw~\cite{HFforLegal2024caselaw} data collected from Delaware and Arizona, respectively. Relevance scores highlight both shared domain terms (\textit{plaintiff}, \textit{defendant}) and unique distributional identifiers (\textit{Delaware}, \textit{Arizona}), with consistent scores between 0.2 and 0.4 reflecting their common legislative lexicon. In conclusion, word relevance scores reliably identify distributionally unique words, enabling robust distributional analysis via word preference.

\begin{figure}[t!]
    \centering
    \vspace{-10pt}
\resizebox{\linewidth}{!}{
    \begin{minipage}[b]{0.45\linewidth}
        \includegraphics[width=\linewidth]{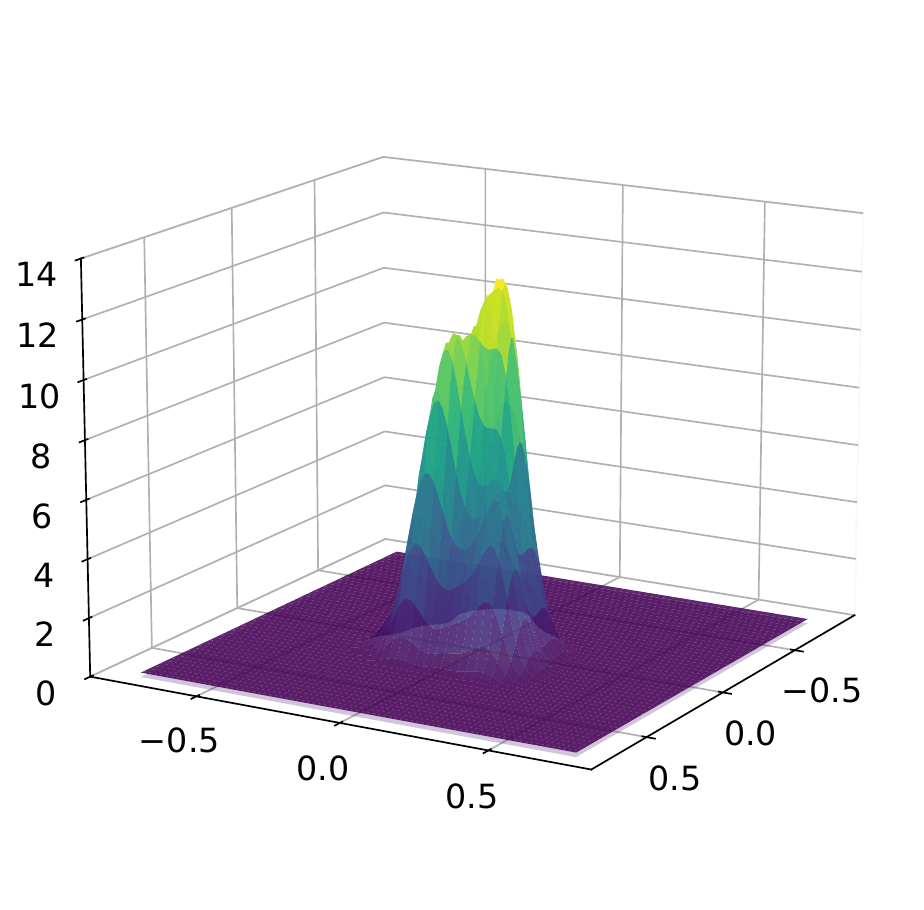}
    \end{minipage}
    \hfill
    \begin{minipage}[b]{0.45\linewidth}
        \includegraphics[width=\linewidth]{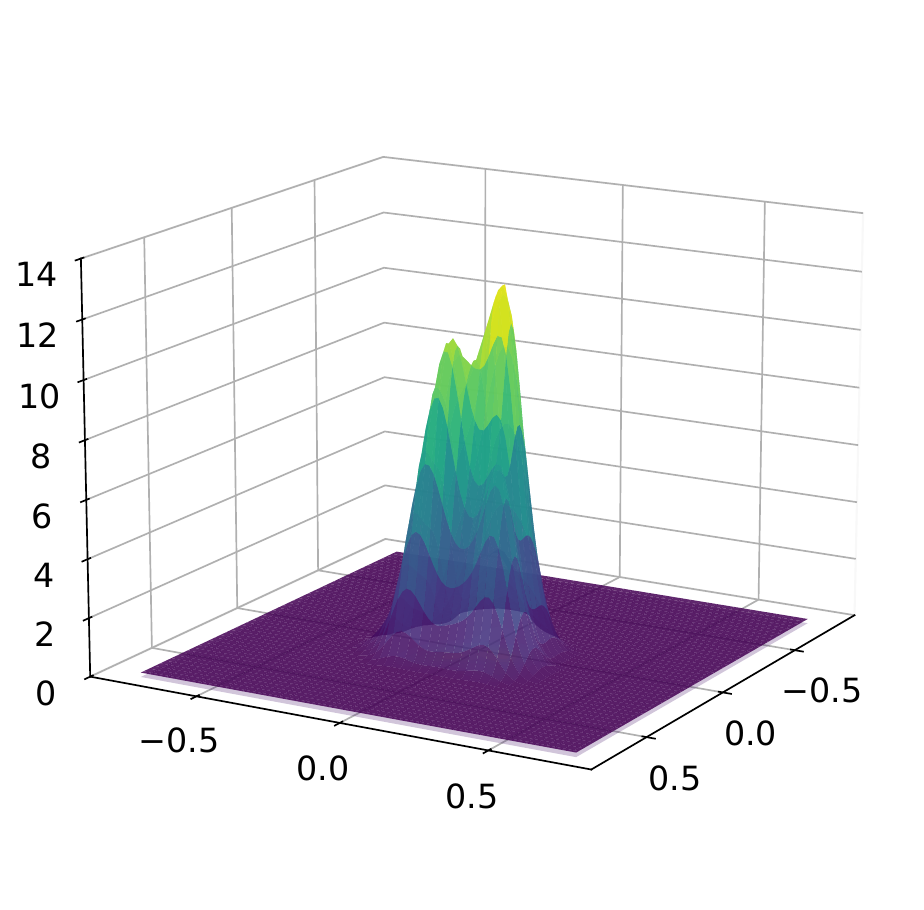}
    \end{minipage}
}
    \vspace{-18pt}
    \caption{
    Semantic distributions of ChatDoctor~\cite{li2023chatdoctor} HCM conversations $D^\text{HCM}_\text{base}$ via embedding K-means clustering, with principal components on the x- and y-axes and probability density on the z-axis. Left: Baseline distribution. Right: Relevance-weighted distribution.
    }
    \label{fig:w-mauve}
\end{figure}

Leveraging word relevance scores to weight semantic distributions, JSLS induces structural changes in the distribution geometry that filter out generic semantic noise, as visualized via PCA in Figure~\ref{fig:w-mauve}. This weighting reduces the superficial semantic similarity between ChatDoctor HCM proprietary data~\cite{li2023chatdoctor} and independent GPT synthetic data (e.g., from 0.811 to 0.729, a decrease of ~10\%), enabling the capture of discrepancies in semantic manifolds that unweighted baselines miss. Besides, JSLS quantifies linguistic style via the JS divergence of relevance distributions. Our analysis reveals that datasets from distinct sources may exhibit significant lexical divergence, as illustrated in Table~\ref{tab:JSLS-validation} by the 1.572 gap between MedAlpaca’s wikidoc and flashcard~\cite{han2025medalcapa} subsets, reflecting potential fundamental terminological differences. Conversely, proprietary paraphrases consistently maintain moderate divergence. Therefore, by integrating these specific lexical signatures with weighted semantic analysis, JSLS reliably measures data distribution with intrinsic fingerprints, ensuring that high scores reflect genuine provenance rather than coincidental overlap .

\subsubsection{Anchoring to Physical Utility and Privacy.}

\begin{figure}[t!]
    \centering
\resizebox{\linewidth}{!}{
    \begin{minipage}[b]{0.48\linewidth}
        \includegraphics[width=\linewidth]{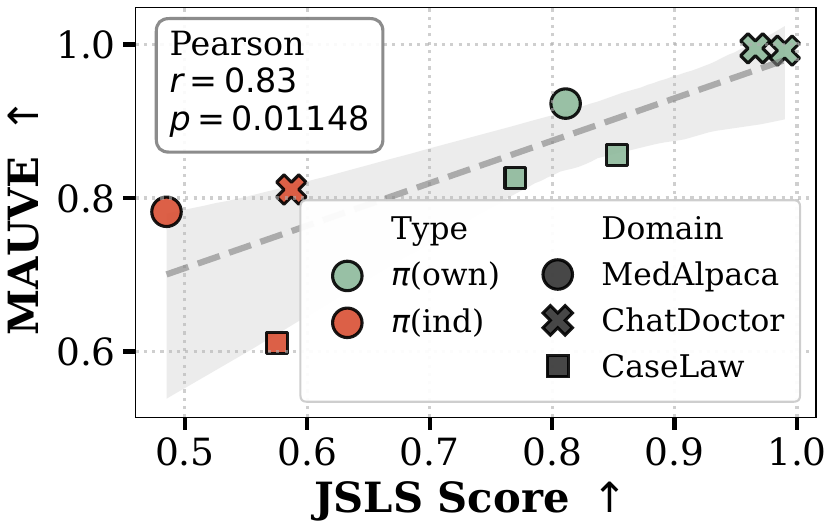}
    \end{minipage}
    \hfill
    \begin{minipage}[b]{0.48\linewidth}
        \includegraphics[width=\linewidth]{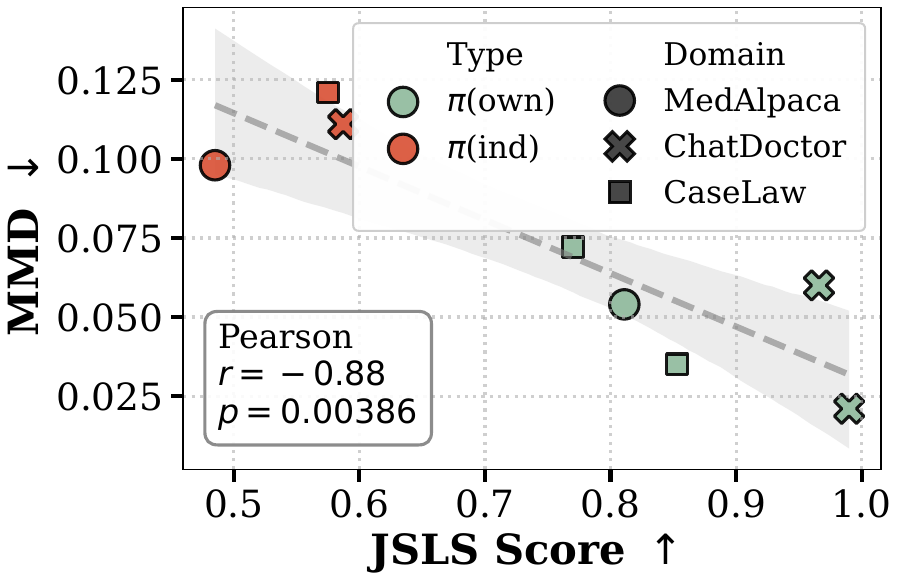}
    \end{minipage}
}
    \caption{
    Anchoring JSLS to semantic reality, i.e. MAUVE~\cite{pillutla2021mauve} (left) and MMD~\cite{liu2020mmd} (right), respectively.
    }
    \label{fig:jsls_semantic_alignment}
\end{figure}

\begin{figure}[t!]
    \centering
\resizebox{\linewidth}{!}{
    \begin{minipage}[b]{0.48\linewidth}
        \includegraphics[width=\linewidth]{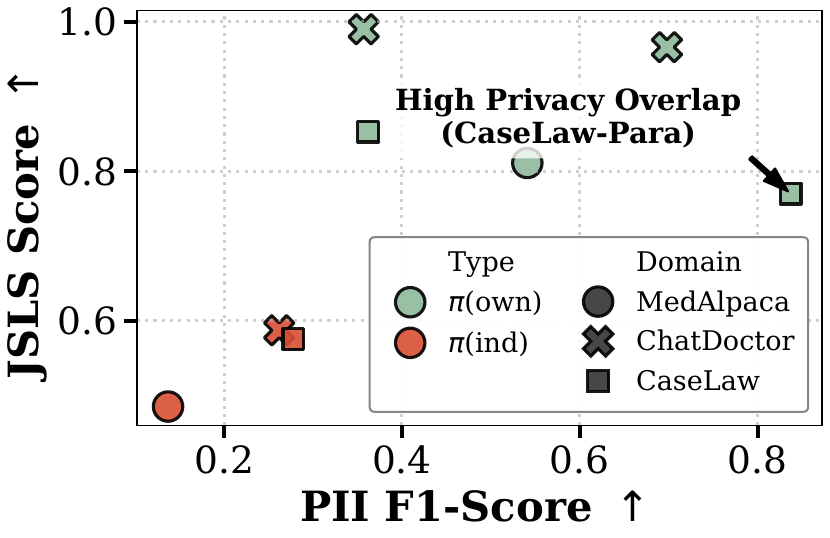}
    \end{minipage}
    \hfill
    \begin{minipage}[b]{0.48\linewidth}
        \includegraphics[width=\linewidth]{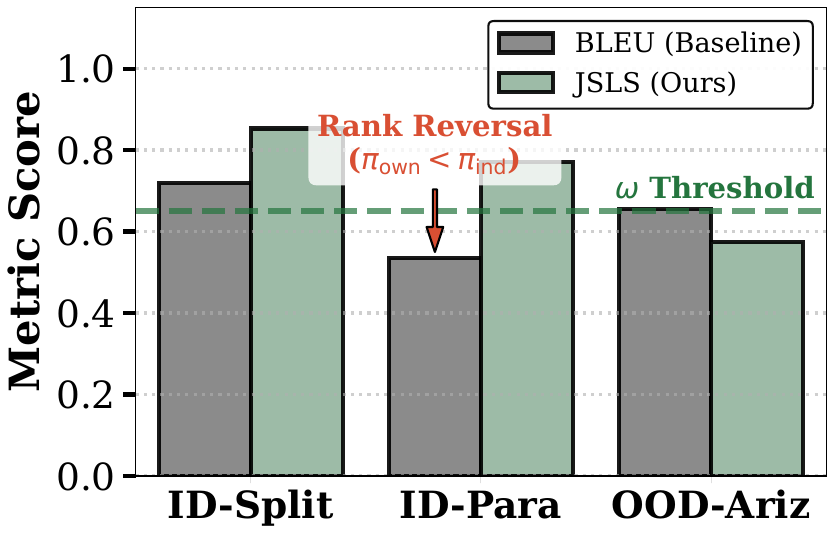}
    \end{minipage}
}
    \caption{
    Comparisons between JSLS between privacy overlaps (left) and lexical baselines (right).
    }
    \label{fig:jsls_privacy_lexical}
\end{figure}

To justify using JSLS as the ground-truth proxy for the DPA hypothesis test, we validate that it reflects practical fine-tuning by anchoring JSLS in two invariants enforced by the rational Trainer’s utility constraint: \textit{semantic fidelity} and \textit{privacy leakage}. As to semantic fidelity, Figure~\ref{fig:jsls_semantic_alignment} benchmarks JSLS against MAUVE~\cite{pillutla2021mauve} (similarity via divergence frontiers) and MMD~\cite{liu2020mmd} (divergence via kernel projection). Correlations are measured by Pearson's $r$, with significance $p$ assessed via t-test. Our analysis reveals a robust statistical coupling: $\Delta_{\text{JSLS}}$ strongly aligns with MAUVE's similarity measure ($r=0.83$) while inversely mirroring MMD's divergence metric ($r=-0.88$). Crucially, these correlations remain statistically significant ($p<0.05$) despite limited samples, confirming that JSLS captures semantic consistency necessitated by model utility. Besides, the left of Figure~\ref{fig:jsls_privacy_lexical} benchmarks JSLS against privacy overlaps via Personal Identifiable Information (PII) F1-scores~\cite{lukas2023pii-leakage}. We observe a distinct positive alignment where JSLS scores scale proportionally with privacy overlaps, effectively distinguishing low-risk independent data from high-risk proprietary baselines. Especially in the CaseLaw-paraphrase case, high PII retention ($F1\approx 0.84$) persists despite lexical obfuscation, and JSLS correctly reflects this with a high score of $0.770$. By grounding JSLS in these independent physical realities, we thereby legitimize its role as a valid proxy for distributional ground truth. Please refer to Appendix~\ref{app:anchoring_stats} for detailed statistics.

JSLS's alignment with reality highlights the critical limitations of existing baselines. Surface-level metrics (e.g., BLEU~\cite{papineni2002bleu}) are vulnerable to false negatives via rank reversal, erroneously scoring obfuscated proprietary data lower than independent outliers, as shown in the right of Figure~\ref{fig:jsls_privacy_lexical}. Appendix~\ref{app:bow} further shows via Bag-of-Words (BoW)~\cite{meeus2024mia-rushing-nowhere} that the limitations of pure lexical metrics mirror the fundamental failure modes of standard MIAs. Conversely, pure semantic methods (e.g., MAUVE~\cite{pillutla2021mauve} and MMD~\cite{liu2020mmd}) suffer from false positives; as shown in Figure~\ref{fig:jsls_semantic_alignment}, they assign proximally ambiguous scores to proprietary and independent domain data, blurring the provenance boundary. Furthermore, invasive interventions like dataset watermarking fail at data paraphrasing~\cite{rastogi2024watermark-rewriting}, leaving them ineffective against malicious trainers.



\subsubsection{Utility-Informed Threshold Determination}

To instantiate a suitable $(\Delta, \omega)$ configuration for the \textit{Dataset-Dataset Approximate Similarity} (Definition~\ref{def:dataset_sim}), we determine the decision threshold $\omega$ by anchoring it to the distinct separation observed between proprietary and independent distributions, as explicitly evidenced in Table~\ref{tab:JSLS-validation}.
There, proprietary derivatives consistently score $\Delta_\text{JSLS} > 0.77$, whereas independent sources fall distinctly below $0.60$.
This divergence arises because utility necessitates fidelity: functional datasets consistently retain high JSLS scores regardless of obfuscation. 
Accordingly, we establish $\omega = 0.60$ as a conservative lower bound that maximizes robustness against evasion while strictly excluding false positives.
This affirmatively answers \textit{R1}: by validating $\omega$ as a \textit{robust proxy} for the provenance ground-truth, we confirm that JSLS effectively captures intrinsic distributional fingerprints.

\begin{table}[t!]
\centering
\caption{Impact of fine-tuning tactics on the DPA solution: measured by mean and std of Extraction Deviation $\delta_\text{ext}$.}
\label{tab:impact-tactics}
\resizebox{0.65\linewidth}{!}{
\begin{tabular}{ccccc}
\toprule
\multicolumn{5}{c}{{\cellcolor[rgb]{0.9, 0.9, 0.9}}\textbf{Different Fine-tuning Tactics}} \\ \midrule
\multicolumn{1}{c|}{} & \textbf{No Tactic} & \textbf{Tactic 1} & \textbf{Tactic 2} & \textbf{Tactic 3} \\ \hline
\multicolumn{1}{c|}{\textbf{$\bm{\delta_\text{ext}}$ mean}} & 0.040 & 0.056 & 0.051 & 0.040 \\
\multicolumn{1}{c|}{\textbf{$\bm{\delta_\text{ext}}$ std}}  & 0.043 & 0.067 & 0.060 & 0.039 \\ \bottomrule
\end{tabular}}
\end{table}

\begin{table}[t!]
\centering
\caption{Impact of models and datasets on the DPA solution: measured by mean and std of Extraction Deviation $\delta_\text{ext}$.}
\label{tab:impact-model-dataset}
\resizebox{\linewidth}{!}{
\begin{tabular}{c|ccc|ccc}
\toprule
\multicolumn{1}{c|}{} & \multicolumn{3}{c|}{\cellcolor[rgb]{0.9, 0.9, 0.9}\textbf{Different Models}} & \multicolumn{3}{c}{\cellcolor[rgb]{0.9, 0.9, 0.9}\textbf{Different Datasets}} \\ 
\multicolumn{1}{c|}{} & \textbf{Pythia} & \textbf{Qwen} & \textbf{Llama} & \textbf{MedAlpaca} & \textbf{ChatDoctor} & \textbf{CaseLaw} \\ \midrule
\textbf{$\bm{\delta_\text{ext}}$ mean} & 0.052 & 0.036 & 0.048 & 0.016 & 0.087 & 0.027 \\
\textbf{$\bm{\delta_\text{ext}}$ std}  & 0.060 & 0.035 & 0.052 & 0.016 & 0.059 & 0.020 \\ \bottomrule
\end{tabular}}
\end{table}

\subsection{Audit Effectiveness and Robustness}

\subsubsection{Faithful Extraction via SIDE}

With JSLS established, DPA's validity hinges on whether the extraction method $\mathcal{G}$ faithfully recovers the latent model distribution $\pi_M$. We validate this extraction fidelity via \textit{Extraction Deviation} ($\delta_\text{ext}$). Ideally, provenance is verified by comparing the auditor's reference $D_\text{ref}$ directly to the root source $D_\text{base}$. However, since $D_\text{base}$ is unobservable and potentially obfuscated, DPA relies on the extracted data $\mathcal{G}(M)$ as a proxy. SIDE's objective is to ensure the extracted signal mirrors the ideal signal relative to $D_\text{ref}$. We quantify this alignment by defining $\delta_\text{ext}$ as:
\begin{equation}
\label{equ:extraction_deviation}
\delta_\text{ext} = \| \Delta_{\text{JSLS}}(D_\text{base}, D_\text{ref}) - \Delta_{\text{JSLS}}(\mathcal{G}(M), D_\text{ref}) \|_2
\end{equation}
A low $\delta_\text{ext}$ indicates that SIDE faithfully recovers the intrinsic fingerprints of the root $D_\text{base}$ despite evasion, accurately projecting the latent dependency into the observable space for auditing.

Table~\ref{tab:DPA-main-results} benchmarks this extraction fidelity. We observe that SIDE generally functions as a faithful extractor, maintaining small extraction deviation from the root source $D_\text{base}$ across diverse evasion tactics. Notably, even under paraphrasing, SIDE faithfully recovers the latent semantic persistence ($m_r$) that survives moderate surface-level lexical penalty ($e^{k\gamma}$) in the JSLS denominator, preserving the integrity of the provenance fingerprints. Crucially, this fidelity ensures reliable auditing: as evidenced by the bold values, the extracted provenance signals typically exceed the decision threshold ($\tau=0.5$) while independent baselines fall below it. This confirms that SIDE effectively projects the latent discriminability of the unauthorized $D_\text{base}$ into the observable space. We use 2K extracted and 2K reference samples by default; smaller sample budgets preserve the ordering between proprietary references and independent decoys, as shown in Appendix~\ref{app:ablation}.

\begin{table*}[ht!]
\centering

\caption{DPA performance across diverse base models and evasion tactics.
Report point-estimate provenance similarity $\Delta_{\text{JSLS}}$
and extraction deviation $\delta_\text{ext}$ (Eq.~\ref{equ:extraction_deviation}).
Under $\tau=0.5$, bold values mark point-estimate attribution
($\Delta_{\text{JSLS}}>\tau$), while \textit{\textbf{bold italics}}
mark point-estimate rejection of independent baselines
($\Delta_{\text{JSLS}}<\tau$). Bootstrap confidence validation is given
in Section~7.3.3.}
\label{tab:DPA-main-results}
\resizebox{0.95\linewidth}{!}{
\begin{tabular}{lccccccccccccc}
\toprule
\multicolumn{14}{c}{\large\textbf{Models fine-tuned with No Tactic and Tactic 1: subset split.}} \\
\toprule
\multirow{2}{*}{\textbf{Dist.}} & \textbf{Reference} & \multicolumn{4}{c}{\textbf{Pythia-2.8B}} & \multicolumn{4}{c}{\textbf{Qwen3-4B-Instruct}} & \multicolumn{4}{c}{\textbf{Llama3-8B-Instruct}} \\ 
\hhline{~~*{4}{|-}||*{4}{-|}||*{4}{-|}} 
& \textbf{Dataset} & \textbf{Sem.} $\bm{(m_r)}$ & \textbf{Lex.} $\bm{(e^{k\gamma})}$ & $\bm{} \bm{\Delta_\textbf{JSLS}}$ & $\bm{\delta_\text{ext}}$ & \textbf{Sem.} $\bm{(m_r)}$ & \textbf{Lex.} $\bm{(e^{k\gamma})}$ & $ \bm{\Delta_\textbf{JSLS}}$ &  $\bm{\delta_\text{ext}}$ & \textbf{Sem.} $\bm{(m_r)}$ & \textbf{Lex.} $\bm{(e^{k\gamma})}$ & $ \bm{\Delta_\textbf{JSLS}}$ & $\bm{\delta_\text{ext}}$  \\ \midrule
\multicolumn{14}{c}{{\cellcolor[rgb]{0.9, 0.9,  0.9}}\textbf{MedAlpaca (compared with SIDE outputs from models fine-tuned on  $D^\text{wiki}_\text{trn}$, omitting superscript ``wiki'' later)}} \\
\hline
{\cellcolor[rgb]{0.6, 1, 0.6}}ID & $D\_\{\text{trn}\}$ & 0.785 & 1.003  &  {\cellcolor[rgb]{0.617, 1,  0.617}}\textbf{0.783} & \textbf{$\bm{<0.001}$}
& 0.874  & 1.138  & {\cellcolor[rgb]{0.732, 1,  0.732}}\textbf{0.768} & \textbf{$\bm{0.002}$}
& 0.837  & 1.049  & {\cellcolor[rgb]{0.602, 1,  0.602}}\textbf{0.798} & \textbf{$\bm{<0.001}$}     \\
{\cellcolor[rgb]{0.6, 1, 0.6}}ID & $D\_\{\text{para}\}$ & 0.747 & 1.106  & {\cellcolor[rgb]{0.724, 1,  0.724}}\textbf{0.676} & \textbf{$\bm{0.018}$}
& 0.834  & 1.389  & {\cellcolor[rgb]{0.800, 1,  0.800}}\textbf{0.600} & \textbf{$\bm{0.045}$}
& 0.793  & 1.283  & {\cellcolor[rgb]{0.782, 1,  0.782}}\textbf{0.618} & \textbf{$\bm{0.037}$} \\
{\cellcolor[rgb]{1, 0.6,  0.6}}OOD & $D$\textasciicircum$\{\text{flash}\}$ & 0.758 & 1.845  & {\cellcolor[rgb]{1, 0.711,  0.711}}\textbf{\textit{0.411}} & \textbf{$\bm{0.005}$}
& 0.664 & 2.227  & {\cellcolor[rgb]{1, 0.598  0.598}}\textbf{\textit{0.298}} & \textbf{$\bm{0.035}$}
& 0.757  & 2.535  & {\cellcolor[rgb]{1,  0.599, 0.599}}\textbf{\textit{0.299}} & \textbf{$\bm{0.034}$}  \\ \midrule
\multicolumn{14}{c}{{\cellcolor[rgb]{0.9, 0.9,  0.9}}\textbf{ChatDoctor (compared with SIDE outputs from models fine-tuned on  $D^\text{HCM}_\text{trn}$, omitting superscript ``HCM'' later)}} \\
\hline
{\cellcolor[rgb]{0.6, 1,  0.6}}ID & $D\_\{\text{trn}\}$ & 0.752 & 1.219  & {\cellcolor[rgb]{0.783, 1,  0.783}}\textbf{0.617} & \textbf{$\bm{0.139}$} 
& 0.778  &  1.083 & {\cellcolor[rgb]{0.682, 1,  0.682}}\textbf{0.718} & \textbf{$\bm{0.074}$}
& 0.762  & 1.147  & {\cellcolor[rgb]{0.736, 1,  0.736}}\textbf{0.664} & \textbf{$\bm{0.106}$}    \\
{\cellcolor[rgb]{0.6, 1,  0.6}}ID & $D\_\{\text{split}\}$ & 0.733  &  1.300 & {\cellcolor[rgb]{0.836, 1,  0.836}}\textbf{0.564} & \textbf{$\bm{0.181}$}
& 0.787  & 1.134  & {\cellcolor[rgb]{0.706, 1,  0.706}}\textbf{0.694} & \textbf{$\bm{0.087}$}
& 0.756  & 1.246  &  {\cellcolor[rgb]{0.785, 1,  0.785}}\textbf{0.615} & \textbf{$\bm{0.141}$}  \\
{\cellcolor[rgb]{0.6, 1,   0.6}}ID & $D\_\{\text{para}\}$ & 0.730 & 1.112  & {\cellcolor[rgb]{0.743, 1,  0.743}}\textbf{0.657} & \textbf{$\bm{0.095}$}
& 0.763  & 1.017  & {\cellcolor[rgb]{0.650, 1,  0.650}}\textbf{0.750} & \textbf{$\bm{0.047}$} 
&  0.742 & 1.288  & {\cellcolor[rgb]{0.824, 1,  0.824}}\textbf{0.576} & \textbf{$\bm{0.152}$} \\
{\cellcolor[rgb]{1, 0.6,  0.6}}OOD & $D$\textasciicircum$\{\text{GPT}\}$ & 0.466 & 1.087 & {\cellcolor[rgb]{1, 0.729, 0.729}}\textbf{\textit{0.429}} & \textbf{$\bm{0.025}$} 
& 0.516  & 1.050  & {\cellcolor[rgb]{1, 0.791, 0.791}}\textbf{\textit{0.491}} & \textbf{$\bm{0.009}$}
& 0.466  & 1.078  & {\cellcolor[rgb]{1, 0.732, 0.732}}\textbf{\textit{0.432}} & \textbf{$\bm{0.024}$}  \\ \midrule
\multicolumn{14}{c}{{\cellcolor[rgb]{0.9, 0.9,  0.9}}\textbf{CaseLaw (compared with SIDE outputs from models fine-tuned on  $D^\text{Del}_\text{trn}$, omitting superscript ``Del'' later)}} \\
\hline
{\cellcolor[rgb]{0.6, 1, 0.6}}ID & $D\_\{\text{trn}\}$ & 0.759 &  1.073 & {\cellcolor[rgb]{0.692, 1,  0.692}}\textbf{0.708} & \textbf{$\bm{0.021}$} 
& 0.892  & 1.004  & {\cellcolor[rgb]{0.511, 1,  0.511}}\textbf{0.889} & \textbf{$\bm{0.001}$} 
& 0.763  &  1.240 &  {\cellcolor[rgb]{0.784, 1,  0.784}}\textbf{0.616} & \textbf{$\bm{0.056}$}   \\
{\cellcolor[rgb]{0.6, 1,  0.6}}ID & $D\_\{\text{split}\}$ & 0.736 & 1.102  & {\cellcolor[rgb]{0.732, 1,  0.732}}\textbf{0.668} & \textbf{$\bm{0.034}$} 
& 0.826  & 1.012  & {\cellcolor[rgb]{0.584, 1,  0.584}}\textbf{0.816} & \textbf{$\bm{0.001}$}
& 0.773  & 1.272  &  {\cellcolor[rgb]{0.792, 1,  0.792}}\textbf{0.608} & \textbf{$\bm{0.060}$}  \\
{\cellcolor[rgb]{0.6, 1,   0.6}}ID & $D\_\{\text{para}\}$ & 0.721 & 1.035  & {\cellcolor[rgb]{0.704, 1,  0.704}}\textbf{0.696} & \textbf{$\bm{0.005}$} 
& 0.661  & 1.238  & {\cellcolor[rgb]{0.711, 1,  0.711}}\textbf{0.689} & \textbf{$\bm{0.007}$} 
&  0.804 & 1.002  &  {\cellcolor[rgb]{0.598, 1,  0.598}}\textbf{0.802} & \textbf{$\bm{0.001}$}   \\
{\cellcolor[rgb]{1, 0.6,  0.6}}OOD & $D$\textasciicircum$\{\text{Ariz}\}$ & 0.561 & 1.322  & {\cellcolor[rgb]{1, 0.724, 0.724}}\textbf{\textit{0.424}} & \textbf{$\bm{0.023}$} 
& 0.538  & 1.147  & {\cellcolor[rgb]{1, 0.769  0.769}}\textbf{\textit{0.469}} & \textbf{$\bm{0.011}$}
& 0.580  & 1.476  &  {\cellcolor[rgb]{1, 0.693,  0.693}}\textbf{\textit{0.393}} & \textbf{$\bm{0.033}$}  \\ \toprule


\multicolumn{14}{c}{\large\textbf{Models fine-tuned with Tactic 2: data paraphrase.}} \\
\toprule
\multirow{2}{*}{\textbf{Dist.}} & \textbf{Reference} & \multicolumn{4}{c}{\textbf{Pythia-2.8B}} & \multicolumn{4}{c}{\textbf{Qwen3-4B-Instruct}} & \multicolumn{4}{c}{\textbf{Llama3-8B-Instruct}} \\ 
\hhline{~~*{4}{|-}||*{4}{-|}||*{4}{-|}} 
& \textbf{Dataset} & \textbf{Sem.} $\bm{(m_r)}$ & \textbf{Lex.} $\bm{(e^{k\gamma})}$ & $\bm{} \bm{\Delta_\textbf{JSLS}}$ & $\bm{\delta_\text{ext}}$ & \textbf{Sem.} $\bm{(m_r)}$ & \textbf{Lex.} $\bm{(e^{k\gamma})}$ & $ \bm{\Delta_\textbf{JSLS}}$ &  $\bm{\delta_\text{ext}}$ & \textbf{Sem.} $\bm{(m_r)}$ & \textbf{Lex.} $\bm{(e^{k\gamma})}$ & $ \bm{\Delta_\textbf{JSLS}}$ & $\bm{\delta_\text{ext}}$  \\ \midrule
\multicolumn{14}{c}{{\cellcolor[rgb]{0.9, 0.9,  0.9}}\textbf{MedAlpaca (compared with SIDE outputs from models fine-tuned on  $D^\text{wiki}_\text{para}$, omitting superscript ``wiki'' later)}} \\
\hline
{\cellcolor[rgb]{0.6, 1,  0.6}}ID & $D\_\{\text{trn}\}$ & 0.739 & 1.166  &  {\cellcolor[rgb]{0.766, 1, 0.766}}\textbf{0.634} & \textbf{$\bm{0.031}$}
& 0.862  & 1.051  & {\cellcolor[rgb]{0.680, 1, 0.680}}\textbf{0.820} & \textbf{$\bm{<0.001}$}
& 0.856  & 1.147  & {\cellcolor[rgb]{0.654, 1, 0.654}}\textbf{0.746} & \textbf{$\bm{0.004}$}     \\
{\cellcolor[rgb]{0.6, 1,  0.6}}ID & $D\_\{\text{para}\}$ & 0.646 & 1.035  & {\cellcolor[rgb]{0.775, 1, 0.775}}\textbf{0.625} & \textbf{$\bm{0.034}$} 
& 0.840  & 1.030  & {\cellcolor[rgb]{0.584, 1, 0.584}}\textbf{0.816} & \textbf{$\bm{<0.001}$}
& 0.854  & 1.014  & {\cellcolor[rgb]{0.558, 1, 0.558}}\textbf{0.842} & \textbf{$\bm{0.001}$} \\
{\cellcolor[rgb]{1, 0.6,  0.6}}OOD & $D$\textasciicircum$\{\text{flash}\}$ & 0.646 & 1.572  & {\cellcolor[rgb]{1, 0.711, 0.711}}\textbf{\textit{0.411}} & \textbf{$\bm{0.005}$}
& 0.708 & 1.471  & {\cellcolor[rgb]{1, 0.781, 0.781}}\textbf{\textit{0.481}} & \textbf{$\bm{<0.001}$}
& 0.752  & 1.593  & {\cellcolor[rgb]{1, 0.772, 0.772}}\textbf{\textit{0.472}} & \textbf{$\bm{<0.001}$}  \\ \midrule
\multicolumn{14}{c}{{\cellcolor[rgb]{0.9, 0.9,  0.9}}\textbf{ChatDoctor (compared with SIDE outputs from models fine-tuned on  $D^\text{HCM}_\text{para}$, omitting superscript ``HCM'' later)}} \\
\hline
{\cellcolor[rgb]{0.6, 1, 0.6}}ID & $D\_\{\text{trn}\}$ & 0.698 & 1.254  &  {\cellcolor[rgb]{0.844, 1, 0.844}}\textbf{0.556} & \textbf{$\bm{0.188}$}  
& 0.718  &  1.077 & {\cellcolor[rgb]{0.733, 1, 0.733}}\textbf{0.667} & \textbf{$\bm{0.104}$} 
& 0.726 & 1.292  & {\cellcolor[rgb]{0.838, 1, 0.838}}\textbf{0.562} & \textbf{$\bm{0.183}$} \\
{\cellcolor[rgb]{0.6, 1,  0.6}}ID & $D\_\{\text{split}\}$ & 0.706 & 1.344  &  {\cellcolor[rgb]{0.875, 1, 0.875}}\textbf{0.525} & \textbf{$\bm{0.216}$} 
& 0.725  & 1.082 & {\cellcolor[rgb]{0.730, 1, 0.730}}\textbf{0.670} & \textbf{$\bm{0.102}$} 
& 0.732  &  1.298 &  {\cellcolor[rgb]{0.836, 1, 0.836}}\textbf{0.564} & \textbf{$\bm{0.181}$}  \\
{\cellcolor[rgb]{0.6, 1,  0.6}}ID & $D\_\{\text{para}\}$ & 0.736 & 1.129  & {\cellcolor[rgb]{0.748, 1, 0.748}}\textbf{0.652} & \textbf{$\bm{0.098}$}  
& 0.755 & 1.010  & {\cellcolor[rgb]{0.652, 1, 0.652}}\textbf{0.748} & \textbf{$\bm{0.047}$}
&  0.758 & 1.16  &  {\cellcolor[rgb]{0.745, 1, 0.745}}\textbf{0.655} & \textbf{$\bm{0.096}$}   \\
{\cellcolor[rgb]{1, 0.6,  0.6}}OOD & $D$\textasciicircum$\{\text{GPT}\}$ & 0.425 & 1.011 & {\cellcolor[rgb]{1, 0.721,  0.721}}\textbf{\textit{0.421}} & \textbf{$\bm{0.028}$} 
& 0.469  & 1.050  & {\cellcolor[rgb]{1, 0.747,  0.747}}\textbf{\textit{0.447}} & \textbf{$\bm{0.019}$}
& 0.454 & 1.014  &  {\cellcolor[rgb]{1, 0.748, 0.748}}\textbf{\textit{0.448}} & \textbf{$\bm{0.019}$}  \\ \midrule
\multicolumn{14}{c}{{\cellcolor[rgb]{0.9, 0.9,  0.9}}\textbf{CaseLaw (compared with SIDE outputs from models fine-tuned on  $D^\text{Del}_\text{para}$, omitting superscript ``Del'' later)}} \\
\hline
{\cellcolor[rgb]{0.6, 1,  0.6}}ID & $D\_\{\text{trn}\}$ & 0.685 &  1.008 & {\cellcolor[rgb]{0.720, 1, 0.720}}\textbf{0.680} & \textbf{$\bm{0.030}$} 
& 0.736 & 1.148  & {\cellcolor[rgb]{0.759, 1, 0.759}}\textbf{0.641} & \textbf{$\bm{0.045}$} 
& 0.813  & 1.189  & {\cellcolor[rgb]{0.716, 1, 0.716}}\textbf{0.684} & \textbf{$\bm{0.029}$}    \\
{\cellcolor[rgb]{0.6, 1,   0.6}}ID & $D\_\{\text{split}\}$ & 0.679 & 1.005  & {\cellcolor[rgb]{0.724, 1, 0.724}}\textbf{0.676} & \textbf{$\bm{0.031}$} 
& 0.703  & 1.103  & {\cellcolor[rgb]{0.762, 1, 0.762}}\textbf{0.638} & \textbf{$\bm{0.046}$}
& 0.797  & 1.206  &  {\cellcolor[rgb]{0.739, 1, 0.739}}\textbf{0.661} & \textbf{$\bm{0.037}$}  \\
{\cellcolor[rgb]{0.6, 1,   0.6}}ID & $D\_\{\text{para}\}$ & 0.742 & 1.272  & {\cellcolor[rgb]{0.817, 1, 0.817}}\textbf{0.583} & \textbf{$\bm{0.035}$} 
& 0.719  & 1.013  & {\cellcolor[rgb]{0.690, 1, 0.690}}\textbf{0.710} & \textbf{$\bm{0.004}$}
&  0.813 & 1.189  & {\cellcolor[rgb]{0.716, 1, 0.716}}\textbf{0.684} & \textbf{$\bm{0.007}$} \\
{\cellcolor[rgb]{1, 0.6,  0.6}}OOD & $D$\textasciicircum$\{\text{Ariz}\}$ & 0.477 & 1.007  & {\cellcolor[rgb]{1, 0.774, 0.774}}\textbf{\textit{0.474}} & \textbf{$\bm{0.010}$}
& 0.435  & 1.001  & {\cellcolor[rgb]{1, 0.734, 0.734}}\textbf{\textit{0.434}} & \textbf{$\bm{0.020}$} 
& 0.441  & 1.022  & {\cellcolor[rgb]{1, 0.732, 0.732}}\textbf{\textit{0.432}} & \textbf{$\bm{0.020}$}  \\  \toprule


\multicolumn{14}{c}{\large\textbf{Models fine-tuned with Tactic 3: knowledge distillation.}} \\
\toprule
\multirow{2}{*}{\textbf{Dist.}} & \textbf{Reference} & \multicolumn{4}{c}{\textbf{Pythia-2.8B}} & \multicolumn{4}{c}{\textbf{Qwen3-4B-Instruct}} & \multicolumn{4}{c}{\textbf{Llama3-8B-Instruct}} \\ 
\hhline{~~*{4}{|-}||*{4}{-|}||*{4}{-|}} 
& \textbf{Dataset} & \textbf{Sem.} $\bm{(m_r)}$ & \textbf{Lex.} $\bm{(e^{k\gamma})}$ & $\bm{} \bm{\Delta_\textbf{JSLS}}$ & $\bm{\delta_\text{ext}}$ & \textbf{Sem.} $\bm{(m_r)}$ & \textbf{Lex.} $\bm{(e^{k\gamma})}$ & $ \bm{\Delta_\textbf{JSLS}}$ &  $\bm{\delta_\text{ext}}$ & \textbf{Sem.} $\bm{(m_r)}$ & \textbf{Lex.} $\bm{(e^{k\gamma})}$ & $ \bm{\Delta_\textbf{JSLS}}$ & $\bm{\delta_\text{ext}}$  \\ \midrule
\multicolumn{14}{c}{{\cellcolor[rgb]{0.9, 0.9,  0.9}}\textbf{MedAlpaca (compared with SIDE outputs from the model knowledge distilled via $D^\text{flash}$ from a teacher fine-tined on $D^\text{wiki}_\text{trn}$, omitting superscript ``wiki'' later)}} \\
\hline
{\cellcolor[rgb]{0.6, 1,  0.6}}ID & $D\_\{\text{trn}\}$ & 0.763 & 1.021  &  {\cellcolor[rgb]{0.653, 1, 0.653}}\textbf{0.747} & \textbf{$\bm{0.004}$} 
& 0.786  & 1.043  & {\cellcolor[rgb]{0.646, 1, 0.646}}\textbf{0.754} & \textbf{$\bm{0.003}$} 
& 0.842  & 1.046  & {\cellcolor[rgb]{0.592, 1, 0.592}}\textbf{0.808} & \textbf{$\bm{<0.001}$}   \\
{\cellcolor[rgb]{0.6, 1,  0.6}}ID & $D\_\{\text{para}\}$ & 0.701 & 1.059  & {\cellcolor[rgb]{0.738, 1, 0.738}}\textbf{0.662} & \textbf{$\bm{0.022}$} 
& 0.781  & 1.199  & {\cellcolor[rgb]{0.749, 1, 0.749}}\textbf{0.651} & \textbf{$\bm{0.026}$}
& 0.786  & 1.248  & {\cellcolor[rgb]{0.771, 1, 0.771}}\textbf{0.629} & \textbf{$\bm{0.033}$} \\
{\cellcolor[rgb]{1, 0.6,  0.6}}OOD & $D$\textasciicircum$\{\text{flash}\}$ & 0.717 & 1.656  & {\cellcolor[rgb]{1, 0.724, 0.724}}\textbf{\textit{0.433}} & \textbf{$\bm{0.003}$}
& 0.742 & 1.743  & {\cellcolor[rgb]{1, 0.725, 0.725}}\textbf{\textit{0.425}} & \textbf{$\bm{0.004}$}
& 0.627  & 1.490  & {\cellcolor[rgb]{1, 0.721, 0.721}}\textbf{\textit{0.421}} & \textbf{$\bm{0.004}$}  \\ \midrule
\multicolumn{14}{c}{{\cellcolor[rgb]{0.9, 0.9,  0.9}}\textbf{ChatDoctor (compared with SIDE outputs from the model knowledge distilled via $D^\text{GPT}$ from a teacher fine-tined on $D^\text{HCM}_\text{trn}$, omitting superscript ``HCM'' later)}} \\
\hline
{\cellcolor[rgb]{0.6, 1,   0.6}}ID & $D\_\{\text{trn}\}$ & 0.754 & 1.189  & {\cellcolor[rgb]{0.766, 1, 0.766}}\textbf{0.634} & \textbf{$\bm{0.127}$} 
& 0.785  &  1.144 & {\cellcolor[rgb]{0.714, 1, 0.714}}\textbf{0.686} & \textbf{$\bm{0.031}$}
& 0.783  & 1.147  & {\cellcolor[rgb]{0.718, 1, 0.718}}\textbf{0.682} & \textbf{$\bm{0.008}$}    \\
{\cellcolor[rgb]{0.6, 1,  0.6}}ID & $D\_\{\text{split}\}$ & 0.719 & 1.166  &  {\cellcolor[rgb]{0.784, 1, 0.784}}\textbf{0.616} & \textbf{$\bm{0.140}$}
& 0.761  & 1.210 & {\cellcolor[rgb]{0.771, 1, 0.771}}\textbf{0.629} & \textbf{$\bm{0.130}$}
& 0.779  & 1.154  &  {\cellcolor[rgb]{0.738, 1, 0.738}}\textbf{0.662} & \textbf{$\bm{0.108}$}  \\
{\cellcolor[rgb]{0.6, 1,  0.6}}ID & $D\_\{\text{para}\}$ & 0.687 & 1.002  & {\cellcolor[rgb]{0.714, 1, 0.714}}\textbf{0.686} & \textbf{$\bm{0.078}$} 
& 0.776 & 1.047  & {\cellcolor[rgb]{0.659, 1, 0.659}}\textbf{0.741} & \textbf{$\bm{0.051}$}
&  0.774 & 1.147  & {\cellcolor[rgb]{0.659, 1, 0.659}}\textbf{0.741} & \textbf{$\bm{0.051}$} \\
{\cellcolor[rgb]{1, 0.6,  0.6}}OOD & $D$\textasciicircum$\{\text{GPT}\}$ & 0.497 & 1.090 & {\cellcolor[rgb]{1, 0.756,  0.756}}\textbf{\textit{0.456}} & \textbf{$\bm{0.017}$}
& 0.468  & 1.084  & {\cellcolor[rgb]{1, 0.732,  0.732}}\textbf{\textit{0.432}} & \textbf{$\bm{0.024}$}
& 0.492  & 1.119  & {\cellcolor[rgb]{1, 0.739,  0.739}}\textbf{\textit{0.439}} & \textbf{$\bm{0.022}$}  \\ \midrule
\multicolumn{14}{c}{{\cellcolor[rgb]{0.9, 0.9,  0.9}}\textbf{CaseLaw (compared with SIDE outputs from the model knowledge distilled via $D^\text{Ariz}$ from a teacher fine-tined on $D^\text{Del}_\text{trn}$, omitting superscript ``Del'' later)}} \\
\hline
{\cellcolor[rgb]{0.6, 1,  0.6}}ID & $D\_\{\text{trn}\}$ & 0.791 &  1.140 & {\cellcolor[rgb]{0.706, 1, 0.706}}\textbf{0.694} & \textbf{$\bm{0.025}$}
& 0.755 & 1.166  & {\cellcolor[rgb]{0.752, 1, 0.752}}\textbf{0.648} & \textbf{$\bm{0.042}$}
& 0.769  &  1.262 &  {\cellcolor[rgb]{0.791, 1, 0.791}}\textbf{0.609} & \textbf{$\bm{0.060}$}   \\
{\cellcolor[rgb]{0.6, 1,   0.6}}ID & $D\_\{\text{split}\}$ & 0.777 & 1.159  & {\cellcolor[rgb]{0.729, 1, 0.729}}\textbf{0.671} & \textbf{$\bm{0.033}$} 
& 0.762  & 1.168  & {\cellcolor[rgb]{0.748, 1, 0.748}}\textbf{0.652} & \textbf{$\bm{0.040}$}
& 0.766  & 1.272  &  {\cellcolor[rgb]{0.798, 1, 0.798}}\textbf{0.602} & \textbf{$\bm{0.063}$}  \\
{\cellcolor[rgb]{0.6, 1,   0.6}}ID & $D\_\{\text{para}\}$ & 0.773 & 1.027  & {\cellcolor[rgb]{0.647, 1, 0.647}}\textbf{0.753} & \textbf{$\bm{<0.001}$} 
& 0.715  & 1.006  & {\cellcolor[rgb]{0.690, 1, 0.690}}\textbf{0.710} & \textbf{$\bm{0.004}$}
&  0.753 & 1.002  &  {\cellcolor[rgb]{0.549, 1, 0.549}}\textbf{0.751} & \textbf{$\bm{<0.001}$}   \\
{\cellcolor[rgb]{1, 0.6,  0.6}}OOD & $D$\textasciicircum$\{\text{Ariz}\}$ & 0.618 & 1.359  & {\cellcolor[rgb]{1, 0.755, 0.755}}\textbf{\textit{0.455}} & \textbf{$\bm{0.014}$}
& 0.588  & 1.287  & {\cellcolor[rgb]{1, 0.757, 0.757}}\textbf{\textit{0.457}} & \textbf{$\bm{0.014}$}
& 0.552 & 1.318  &  {\cellcolor[rgb]{1, 0.719, 0.719}}\textbf{\textit{0.419}} & \textbf{$\bm{0.024}$}  \\ \bottomrule
\end{tabular}}
\end{table*}

We further quantify SIDE's robustness via extraction deviation statistics. Table~\ref{tab:impact-tactics} confirms resilience against evasion: even under paraphrasing or distillation, deviations remain low on average ($\delta_\text{ext} \approx 0.040 \sim 0.051$), indicating reliable signal retention. Table~\ref{tab:impact-model-dataset} similarly demonstrates invariance to architectures and datasets, maintaining small deviations across diverse settings. Therefore, SIDE acts as a robust extractor that faithfully recovers intrinsic fingerprints, establishing a trustworthy foundation for the subsequent hypothesis test. Unless specified otherwise, subsequent experiments employ Qwen3-4B-Instruct~\cite{yang2025qwen3} as the surrogate LLM.

\subsubsection{Comparison with Baseline Extraction.} 

To validate SIDE's superiority, we compare it with five baseline extraction strategies using models fine-tuned on ChatDoctor ($D^\text{HCM}_\text{trn}$). Note that we include In-Distribution (ID) prompts as an \textit{oracle} baseline to establish the theoretical performance upper bound, although they are basically infeasible in black-box audits. The baselines include:
\begin{itemize}
    \item \textbf{Empty prompts:} Extraction via null inputs~\cite{zeng2024keyword-extraction}.
    \item \textbf{Tailored prompts:} Human-designed domain-specific queries (e.g., `\textit{Hi, doctor...}' in this case)~\cite{huang2025LLM-property-inference}.
    \item \textbf{ID prompts (Oracle):} Prompts directly sampled from the In-Distribution (ID) auxiliary training split $D^\text{HCM}_\text{split}$~\cite{carlini2021data-extraction}.
    \item \textbf{OOD prompts:} Prompts from an Out-Of-Distribution (OOD) dataset within the same domain (e.g., $D^\text{GPT}$)~\cite{fu2023self-prompt-mia}.
    \item \textbf{Unrelated prompts:} Prompts from disjoint. unrelated domains (e.g. Alpaca-cleaned~\cite{Taori2023alpaca})~\cite{carlini2021data-extraction}.
\end{itemize}

Table~\ref{tab:DE-comparison} reports the statistics of extraction deviation  $\delta_\text{ext}$ quantified with JSLS. As expected, the oracle ID prompts achieve the lowest deviation ($\delta_\text{ext} \approx 0.046$) by directly querying with the target distribution. Crucially, SIDE ($\delta_\text{ext} \approx 0.054$) matches this oracle performance, significantly outperforming other heuristics. Conversely, unrelated prompts yield high deviation ($\delta_\text{ext} > 0.6$), as they force the model to generate generic content rather than the target medical distribution, confirming severe prompt-induced bias.

\begin{table}[t!]
\centering
\caption{Impact of extraction algorithms on sampling fidelity: measured by mean and std of Extraction Deviation $\delta_\text{ext}$. SIDE achieves fidelity comparable to the Oracle ID baseline.}
\label{tab:DE-comparison}
\resizebox{0.9\linewidth}{!}{
\begin{tabular}{ccccccc}
\toprule
\multicolumn{7}{c}{\cellcolor[rgb]{0.9, 0.9,  0.9}\textbf{Different Extraction Algorithms}}                                                                                    \\ \midrule
\multicolumn{1}{c|}{\textbf{}}          & \textbf{Empty} & \textbf{Tailored} & \textbf{ID (Oracle)} & \textbf{OOD} & \textbf{Unrelated} & \textbf{\cellcolor[rgb]{0.9, 0.9,  0.9}SIDE(Ours)} \\ \midrule
\multicolumn{1}{c|}{\textbf{$\bm{\delta_\text{ext}}$ mean}} & 0.226          & 0.193             & 0.046        & 0.074        & 0.683              & \cellcolor[rgb]{0.9, 0.9,  0.9}0.054               \\
\multicolumn{1}{c|}{\textbf{$\bm{\delta_\text{ext}}$ std}}  & 0.111          & 0.091             & 0.027        & 0.041        & 0.258              & \cellcolor[rgb]{0.9, 0.9,  0.9}0.030               \\ \bottomrule
\end{tabular}}
\end{table}

\begin{figure}[t!] 
  \centering
\resizebox{\linewidth}{!}{\includegraphics[width=\linewidth]{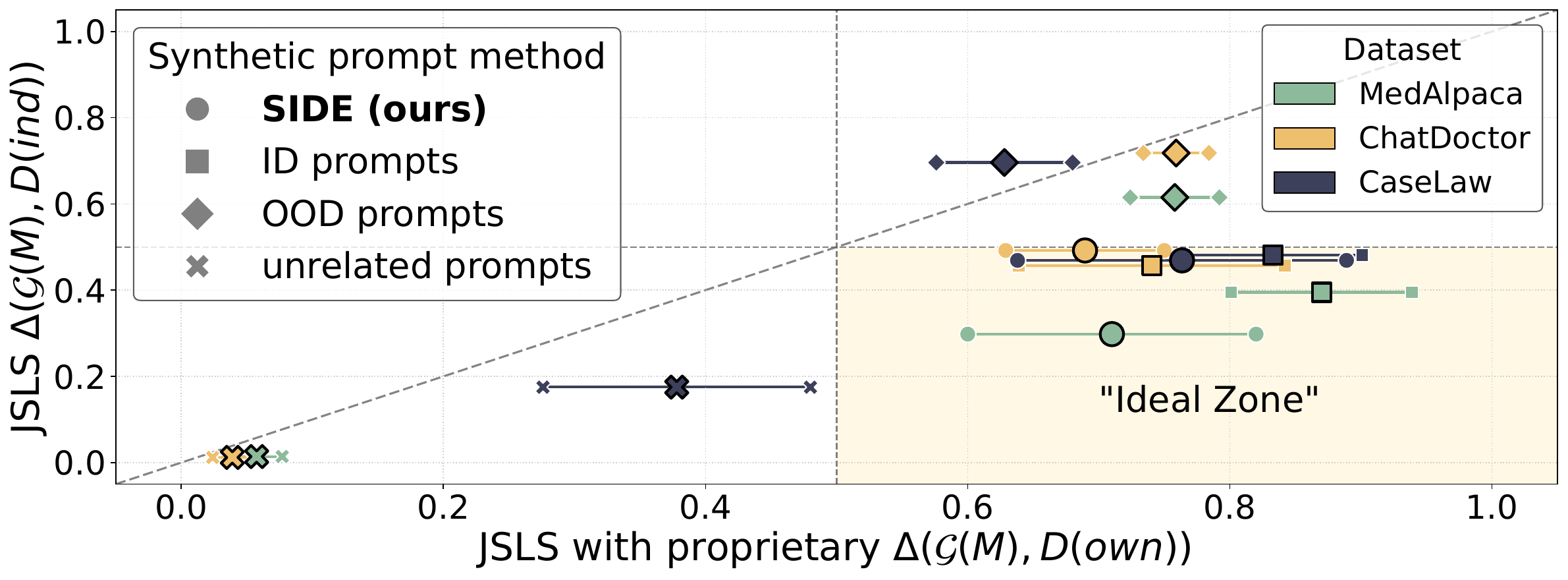}}
  \vspace{-12pt}
  \caption{Visualizing prompt bias: Extracted distributions projected onto ID vs. OOD similarity axes. SIDE (circles) aligns with the oracle ID baseline (squares), whereas heuristic prompts introduce significant distributional shifts.} 
  \label{fig:prompt-devation}
\end{figure}

Figure~\ref{fig:prompt-devation} visually dissects this bias by plotting JSLS signals against $\pi_\text{own}$ (x-axis) and $\pi_\text{ind}$ (y-axis). We observe that oracle ID prompts (squares) define the ground-truth ``ideal zone'' in the bottom-right (high to $\pi_\text{own}$ and low to $\pi_\text{ind}$). Conversely, baseline strategies exhibit distinct shifts: OOD prompts (diamonds) drift toward the upper-right due to distributional contamination, while unrelated prompts (crosses) collapse into the bottom-left, reflecting generic signal loss. Crucially, SIDE (circles) tightly overlaps with the oracle ID cluster. This alignment confirms that SIDE effectively mitigates prompt contamination, projecting the latent model distribution as faithfully as the original training data.

\subsubsection{Hypothesis Testing Outcomes.}

Ultimately, DPA's effectiveness depends on the consistency between
the theoretical provenance label ($b$) and the empirical audit decision
($b'$), as formulated in Section~\ref{sec4:dpa_framework}. This requires a calibrated threshold pair $(\omega,\tau)$: $\omega$ defines distributional separation in the ground-truth space, while $\tau$ serves as the operational audit threshold on extracted samples. We measure this consistency using
the DPA Advantage ($\textbf{Adv}^{\text{DPA}}_{\mathcal{A}}$, Eq.~\ref{equ:adv_dpa}), derived from True Positive Rate (TPR) and False Positive Rate (FPR).

Tables~\ref{tab:JSLS-validation} and~\ref{tab:DPA-main-results}
validate the calibrated setting $(\omega=0.60,\tau=0.50)$. The threshold
$\omega=0.60$ separates proprietary sources from independent ones in the
ground-truth distributional space, while $\tau=0.50$ provides an operational
boundary for black-box auditing with extracted samples. Table~\ref{tab:DPA-main-results} thus reports point-estimate audit outcomes based on $\Delta_{\text{JSLS}}$, where proprietary references are expected to exceed $\tau$, whereas independent decoys should
fall below it.

These decisions are further validated as non-parametric hypothesis tests
using dataset-level bootstrap. For each replicate, we resample both extracted
and reference corpora to recompute $\Delta_{\text{JSLS}}$. DPA rejects $H_0$
only when the 95\% lower confidence bound exceeds $\tau$. Even for boundary
cases in Table~\ref{tab:DPA-main-results}, the lowest-scoring positive case
has a 95\% confidence interval (CI) of $[0.5132,0.6021]$, whose lower bound
remains above $\tau=0.5$; hence $H_0$ is rejected. In contrast, the hardest
negative case yields a 95\% CI of $[0.4615,0.5261]$, whose lower bound does
not exceed $\tau$, so DPA reports insufficient evidence rather than a positive attribution.

\begin{figure}[t!] 
  \centering
\resizebox{0.95\linewidth}{!}{\includegraphics[width=\linewidth]{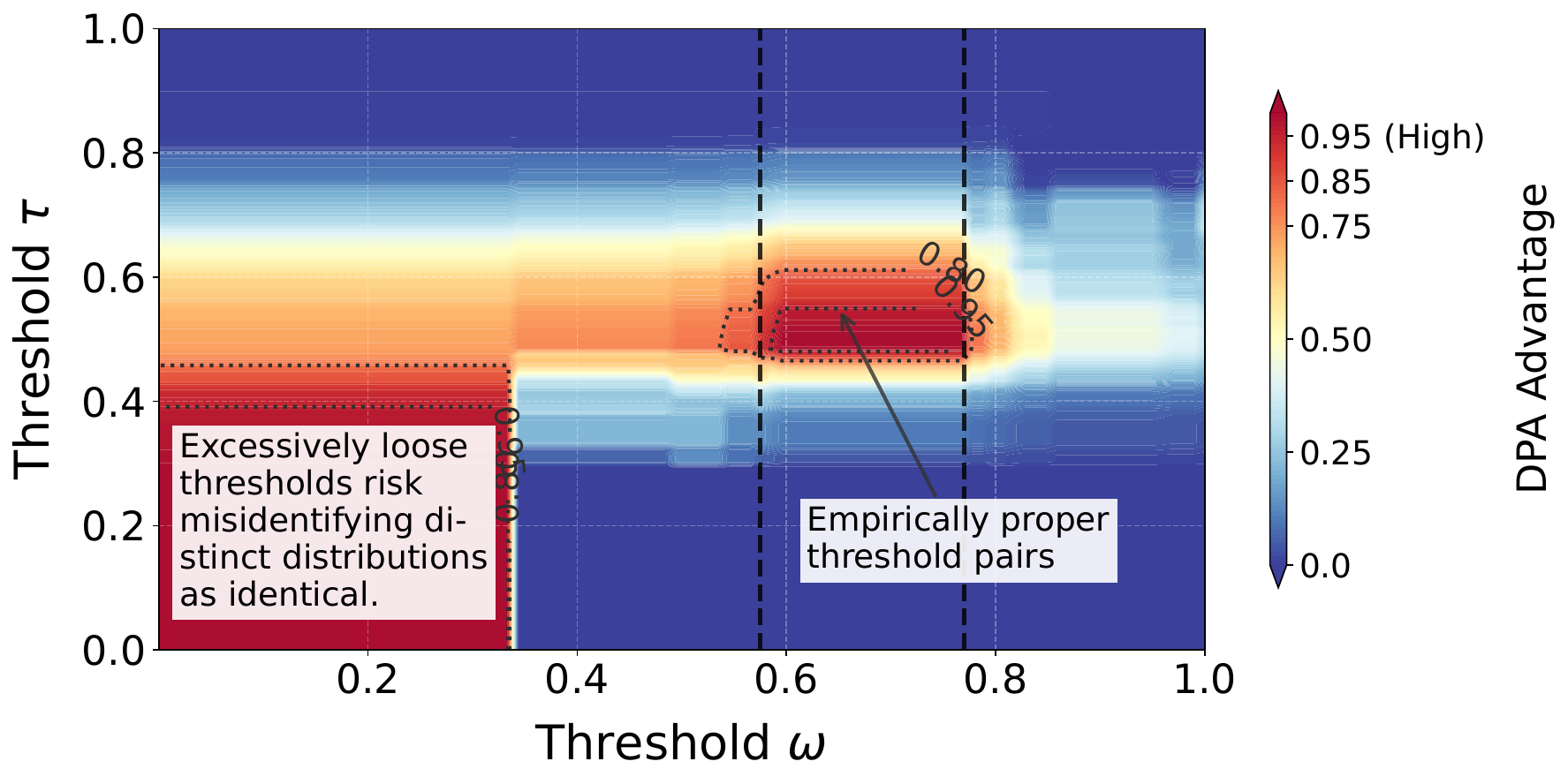}}
  \caption{Analysis of DPA thresholds $\omega$ and $\tau$ for provenance.} 
  \label{fig:threshold-choices}
\end{figure}

Figure~\ref{fig:threshold-choices} visualizes $\textbf{Adv}^{\text{DPA}}_{\mathcal{A}}$ over $(\omega,\tau)$, where positive
detection requires both ground-truth separation
($\Delta_{\text{trn}}\ge\omega$) and operational evidence ($\Delta_{\text{ext}}\ge\tau$). The high-advantage ``island'' stably favors
$\tau\le\omega$ (especially $\tau\in[0.45,0.60]$), consistent with
information-bottleneck attenuation of extracted fingerprints. Outside this
region, overly loose thresholds conflate independent distributions and reduce
audit reliability. Therefore, this operational stability affirmatively answers \textit{R2}: by coupling SIDE's latent distributional extraction with the robust JSLS metric, DPA achieves reliable audit performance across datasets and
evasion tactics; averaged over the stable threshold region, it attains
95.62\% TPR and 4.38\% FPR.

\subsubsection{Production-Model Evaluation}

We further test DPA on a production fine-tuned \texttt{gpt-4.1-nano}~\cite{openai_models_docs}
model trained with 3K MedAlpaca WikiDoc~\cite{han2025medalcapa} samples for 5 epochs. Proprietary
references, including paraphrased variants, obtain high
$\Delta_{\text{JSLS}}$ scores (over $0.76$), while OOD decoys remain separated
(around $0.42$). The ID proprietary audit yields a 95\% CI of
$[0.82,0.90]$, whose lower bound exceeds $\tau=0.5$; the OOD decoy does
not reject $H_0$. This confirms DPA's applicability to black-box
production fine-tuned models.

\subsection{Privacy Implications and Duality}

Beyond data provenance, DPA reveals a fundamental duality: the mechanisms verifying data usage are structurally isomorphic to those performing privacy attacks. To answer \textit{R3}, we now explore this ``double-edged sword,'' empirically demonstrating how high-fidelity fingerprints can be repurposed to facilitate privacy risks.

\subsubsection{Data Extraction Attack}

\begin{figure}[t!] 
  \centering
  \includegraphics[width=\linewidth]{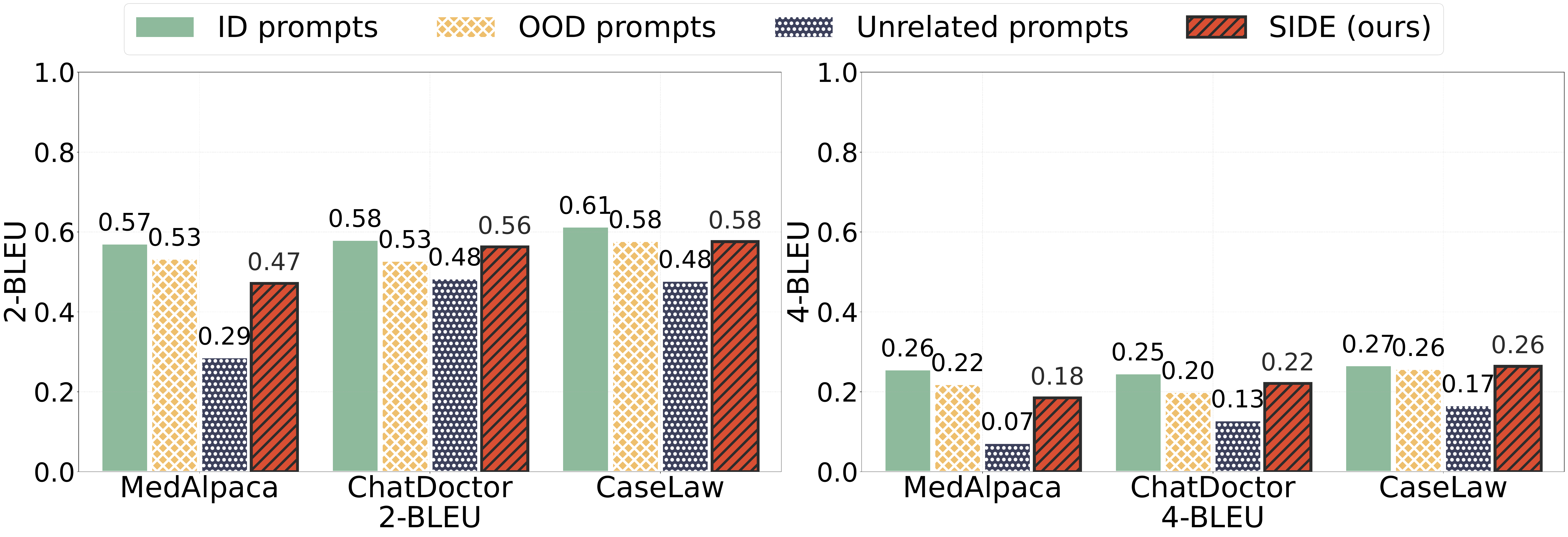} 
  \caption{2-BLEU (left figure) and 4-BLEU (right figure) for different data extraction methods.}
    \label{fig:data-bleu}
\end{figure}

\begin{figure}[t!] 
  \centering
  \includegraphics[width=\linewidth]{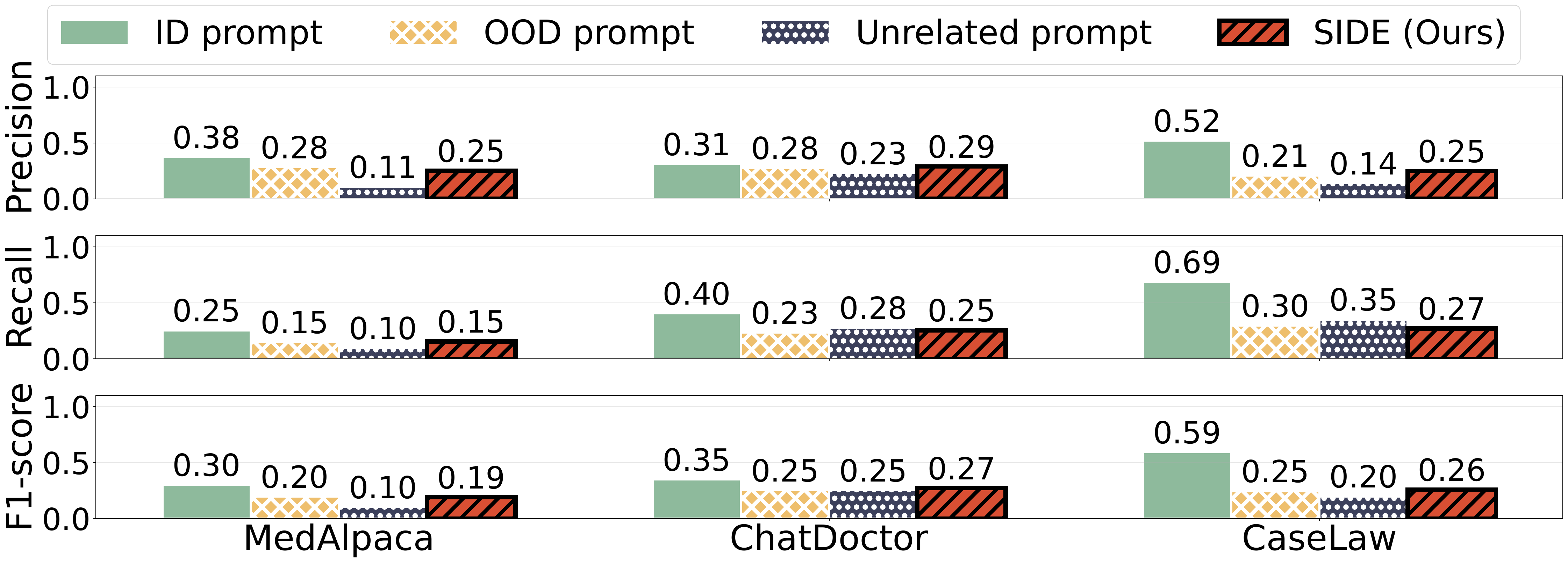} 
  \caption{Performance comparison of Personal Identifiable Information (PII) extraction~\cite{lukas2023pii-leakage} across different data extraction methods, evaluating precision, recall, and F1-score.} 
  \label{fig:pii-extraction}
\end{figure}

SIDE's effectiveness in reconstructing training data is evident in Figure~\ref{fig:data-bleu}, where it achieves BLEU scores comparable to oracle ID prompts and outperforms unrelated prompts by over 30\%, proving its ability to recover high-fidelity text without data access. This extends to sensitive information: Figure~\ref{fig:pii-extraction} confirms concrete privacy risks, as SIDE effectively extracts PII, achieving recall and F1-scores competitive with domain-specific baselines (e.g., $>20\%$ recall), thus validating the structural tension between auditability and leakage.


\subsubsection{Distribution Inference Attack}

We demonstrate that the high-resolution alignment of DPA can be exploited for distribution inference attacks~\cite{hartmann2023distribution-inference, huang2025LLM-property-inference}. By treating the JSLS score ($\Delta_{\text{JSLS}}$) as a proxy for likelihood, an adversary can infer fine-grained properties of the training data, such as the exact mixing proportion of contributing parties, simply by maximizing the similarity score over a set of candidate reference distributions. Table~\ref{tab:DPA-membership-inference} illustrates this vulnerability using models fine-tuned on varying mixtures of Delaware ($D^\text{Del}$) and Arizona ($D^\text{Ariz}$) CaseLaw data. The JSLS score exhibits a distinct diagonal dominance, consistently maximizing when the reference distribution's composition aligns with the ground truth. For example, regarding a target model trained on a 25\% $D^\text{Del}$ mixture, DPA assigns the maximum score ($\Delta_{\text{JSLS}} = 0.823$) to the correct 25\% reference, significantly suppressing scores for pure distributions (e.g., $0.647$ for pure $D^\text{Ariz}$). This consistency allows adversaries to reverse-engineer the precise composition of private training sets, confirming that the high fidelity required for provenance auditing simultaneously exposes the model to granular property inference, both of which capture intrinsic distributional fingerprints. Appendix~\ref{app:privacy_defenses} further demonstrates that DPA resists differential privacy defenses, highlighting the associated privacy risks.

\subsubsection{Auxiliary Dataset Synthesis and Selection.}

\begin{figure}[t!] 
  \centering
  \includegraphics[width=0.9\linewidth]{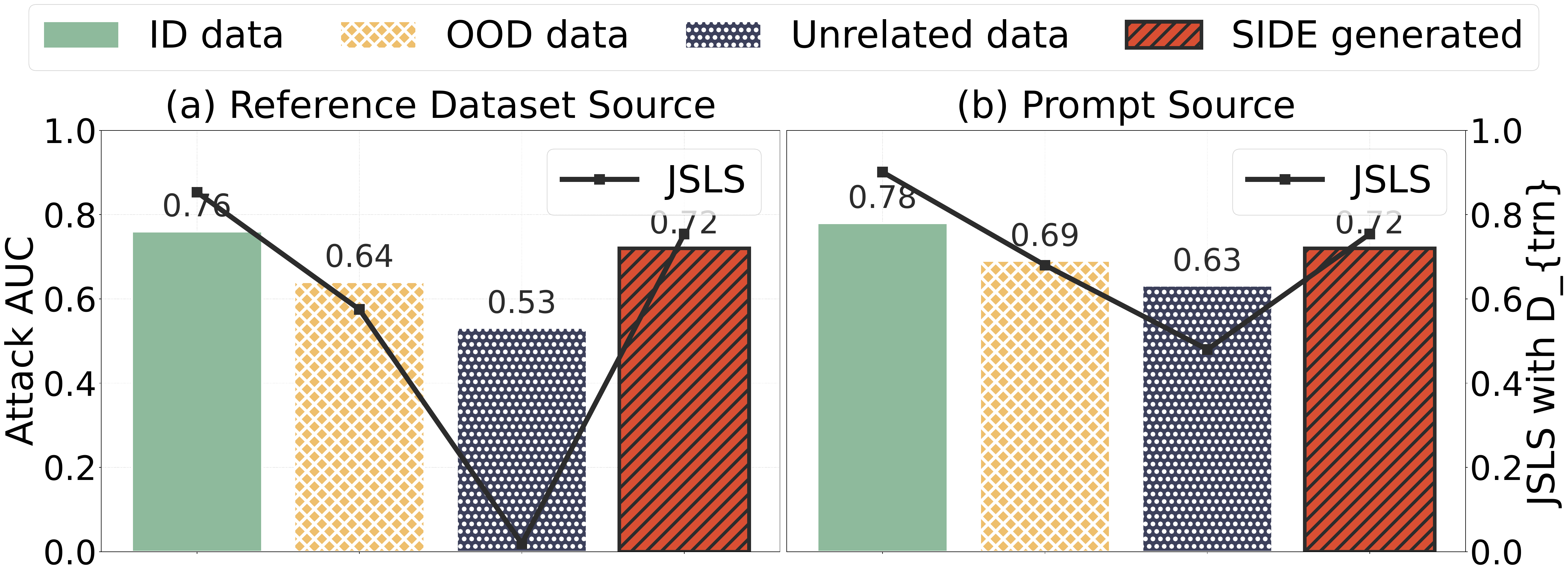} 
  \caption{The performances of reference-based MIA (Ref)~\cite{carlini2021data-extraction} on CaseLaw while utilizing different reference dataset sources and prompt methods, respectively.} 
  \label{fig:auxiliary-dataset-selection}
\end{figure}

Sophisticated MIAs often rely on auxiliary datasets for membership calibration~\cite{carlini2021data-extraction, fu2023self-prompt-mia}. DPA can exploit this dependency by enabling both the synthesis and selection of optimal reference data. Figure~\ref{fig:auxiliary-dataset-selection} reports the AUC of Reference-based MIA~\cite{carlini2021data-extraction} on CaseLaw, comparing various reference datasets (left) and prompting strategies (right) against our SIDE baseline. All experiments utilize $D^\text{Del}_\text{base}$ as members and $D^\text{Del}_\text{split}$ as non-members. Specifically, SIDE-generated data (red, slashed) achieves an MIA AUC of 0.72, significantly outperforming OOD ($\approx 0.64$) and unrelated ($\approx 0.53$) baselines, effectively synthesizing high-utility attack materials that oracle ID data. Crucially, the overlaid plots reveal a strong positive correlation between the JSLS score (black line) and attack success (bars). This confirms that JSLS serves as a reliable predictor of attack effectiveness, allowing adversaries to strategically select, or synthetically generate, the most damaging auxiliary datasets solely based on distributional similarity scores.

\begin{table}[t!]
\centering
\caption{The JSLS results $\Delta_\text{JSLS}(\mathcal{G}(M), D_\text{ref})$ in DPA enables distribution membership inference attack~\cite{hartmann2023distribution-inference}.}
\resizebox{0.75\linewidth}{!}{
\begin{tabular}{lccccc} 
\toprule
\multirow{2}{*}{\textbf{Fine-tuning $D_\text{trn}$ Mix}} & \multicolumn{5}{c}{\textbf{Reference $D_\text{ref}$:} $\bm{D^{\text{Del}}}$ \textbf{Proportion}} \\
\cmidrule(lr){2-6} 
& \textbf{0\%} & \textbf{25\%} & \textbf{50\%} & \textbf{75\%} & \textbf{100\%} \\
\midrule
\textbf{25\%} $\bm{D^{\text{Del}}}$ \textbf{+ 75\%} $\bm{D^{\text{Ariz}}}$ & 0.647 & {\cellcolor[rgb]{0.9, 0.9,  0.9}}\textbf{0.823} & 0.782 & 0.738 & 0.659 \\ 
\textbf{50\%} $\bm{D^{\text{Del}}}$ \textbf{+ 50\%} $\bm{D^{\text{Ariz}}}$ & 0.632 & 0.830 & {\cellcolor[rgb]{0.9, 0.9,  0.9}}\textbf{0.839} & 0.769 & 0.747 \\
\textbf{75\%} $\bm{D^{\text{Del}}}$ \textbf{+ 25\%} $\bm{D^{\text{Ariz}}}$ & 0.671 & 0.781 & 0.815 & {\cellcolor[rgb]{0.9, 0.9,  0.9}}\textbf{0.833} & 0.828 \\
\bottomrule
\end{tabular}}
\label{tab:DPA-membership-inference}
\end{table}

\section{Conclusion}
\label{sec8:conclusion}

This work presents \textit{Distribution Provenance Audit (DPA)}, a robust post-hoc framework that anchors the audit in the fundamental utility constraint of LLM fine-tuning: the necessity to preserve \textit{intrinsic distributional fingerprints}. DPA formalizes the audit as a distinguishability game with two technical innovations: (1) \textit{Joint Semantic-Lexical Similarity (JSLS)},
which identifies integrated semantic and lexical signatures as a stable similarity proxy for provenance probability; (2) \textit{Self-Instruct Data Extraction (SIDE)}, which reconstructs the model's latent distribution without prompt-induced biases. Extensive evaluations demonstrate DPA's reliability across diverse audit settings. Meanwhile, our findings uncover a critical \textit{transparency paradox}: the high-fidelity evidence required for reliable provenance auditing is structurally isomorphic to mechanisms enabling privacy attacks, revealing an inherent tension between verifying Data IP and preventing privacy leakage.

In conclusion, this work proposes a paradigm shift toward auditing intrinsic
distributional fingerprints, while revealing the dialectical tension between
AI transparency and privacy risks. As a post-hoc audit, DPA assumes measurable
distributional differences between proprietary and independent datasets; when
they are statistically indistinguishable, or when a small proprietary fraction is heavily mixed into background data, provenance evidence becomes weakened
or non-identifiable. Future work will extend DPA to multi-modal settings and
more complex distribution mixtures, especially those integrating heterogeneous
domain knowledge, to establish a comprehensive methodology for data provenance in the LLM era.


\section*{Ethical Considerations}

DPA is dual-use: the same mechanisms that enable provenance auditing may
also expose privacy risks, as discussed in Section~\ref{sec6:extraction}.
To mitigate harm, we restrict experiments to public datasets and synthetic corpora generated via public APIs; no private user data is collected or processed. PII extraction is evaluated only quantitatively through aggregate F1 scores, without publishing sensitive content. We believe the benefit of enabling Data Owners to audit unauthorized model training justifies this controlled study, while highlighting the need to
jointly address auditability and privacy.
\section*{Generative AI Usage}

In accordance with the ACM Policy on Authorship, we disclose that
\textit{Gemini}~\cite{comanici2025gemini} and
\textit{ChatGPT}~\cite{achiam2023gpt} were used to improve clarity, coherence, and presentation. \textit{DeepSeek}~\cite{guo2025deepseek} was used to generate paraphrased datasets for evasion evaluation. All AI-assisted content was reviewed and verified by the authors, who retain full responsibility for the work.

\begin{acks}
We sincerely thank chairs and anonymous reviewers for their constructive feedback, which helped a lot to improve this paper. This work was partially funded by NSFC Grants No.62272222 and 62272215, Jiangsu Province Outstanding Youth Fund Project (No. BK20230080), and Nanjing University-China Mobile Communications Group Co.,Ltd. Joint Institute.
\end{acks}

\bibliographystyle{ACM-Reference-Format}
\balance
\bibliography{sample-base}

\appendix

\section{The Landscape of Data IP Protection in the LLM Supply Chain}
\label{app:supply_chain_discussion}

\begin{table*}[t!]
\centering
\caption{Positioning DPA in the LLM Copyright Protection Supply Chain. 
DPA uniquely addresses the upstream Data IP gap by enabling auditor-initiated, post-hoc provenance verification via intrinsic distributional fingerprints, complementing existing Model IP and Output IP protection paradigms~\cite{xu2025copyright}.}
\label{tab:supply_chain_comparison}
\resizebox{0.8\linewidth}{!}{
\begin{tabular}{l|ccc}
\toprule
\textbf{Dimension} 
& \textbf{Upstream (Data $\rightarrow$ Model)} 
& \textbf{Midstream (Model)} 
& \textbf{Downstream (Model $\rightarrow$ User)} \\ 
\midrule
\textbf{Protected Asset} 
& \textbf{Data IP} (Proprietary Training Corpora) 
& \textbf{Model IP} (Weights / Architecture) 
& \textbf{Output IP} (Generated Content) \\

\textbf{Primary Infringement Risk} 
& Unauthorized Fine-tuning / Data Misuse 
& Model Extraction / Weight Theft 
& Plagiarism / Unauthorized Content Reuse \\

\textbf{Typical Defense Paradigm} 
& Dataset Watermarking (Surrogate-based) 
& Model Watermarking / Fingerprinting 
& Text Watermarking \\

\textbf{Verification Agency} 
& \textbf{Auditor-Initiated} (Data Owner) 
& Prover-Centric or Auditor-Assisted 
& Auditor-Initiated (LLM Holder) \\

\textbf{Fingerprint Nature} 
& \textbf{Intrinsic} (Distributional Statistics) 
& Invasive or Intrinsic (Parameter-Level) 
& Invasive (Sampling or Logit Perturbation) \\

\textbf{Deployment Timing} 
& \textbf{Post-hoc} 
& Pre- or Post-hoc 
& Online / Inference-Time \\

\textbf{Utility Preservation} 
& \textbf{Preserved} 
& Typically Preserved 
& Often Degraded \\

\textbf{Robustness to Rewriting} 
& \textbf{High} (Distribution-Level) 
& N/A 
& Low \\ 
\midrule
\textbf{Applicable Method} 
& \textbf{Distribution Provenance Audit (Ours)} 
& Model Watermarking / ZKP 
& Text Watermarking \\ 
\bottomrule
\end{tabular}}
\end{table*}

To rigorously position our contributions, we discuss \textit{Distribution Provenance Audit (DPA)} within the broader taxonomy of LLM copyright protection mechanisms established~\cite{xu2025copyright}. As illustrated in Table~\ref{tab:supply_chain_comparison}, the LLM supply chain comprises three critical assets susceptible to infringement: (1) \textit{Training Data} (Upstream), (2) \textit{Model Weights} (Midstream), and (3) \textit{Generated Content} (Downstream). These copyright risks primarily arise from \textbf{unauthorized distribution}, where proprietary models or private datasets are illicitly leaked or stolen by adversaries, and \textbf{violation of open-source license agreements}, where assets restricted for non-commercial or research use are unlawfully exploited for commercial fine-tuning or deployment~\cite{li2025datasetwatermark, xu2025copyright}. These risks necessitate a systematic re-evaluation of current defense paradigms: this work is distinguished from existing literature across three critical dimensions: the protected asset (Model vs. Data), the verification agency (Prover vs. Auditor), and the fingerprint nature (Invasive vs. Intrinsic).

\vspace{2mm}
\noindent\textbf{Protection asset.} While midstream mechanisms safeguard model parameters (Model IP), upstream and downstream strategies target textual assets (Data/Output IP). This work specifically bridges the upstream gap. Although upstream and downstream watermarking operate at different stages, they share a fundamental mechanism: injecting statistical signals into textual content. The distinction lies in implementation—upstream methods often employ surrogate models to embed embedding signals~\cite{li2025datasetwatermark, qiu2025icassp-watermarking}, whereas downstream methods directly perturb token probabilities or sampling constraints during inference~\cite{xu2025copyright}. Consequently, both paradigms face similar limitations: they inevitably compromise content utility and are vulnerable to text rewriting~\cite{rastogi2024watermark-rewriting}. Unlike these fragile artifacts, our method remains robust even against data paraphrasing.

\vspace{2mm}
\noindent\textbf{Verification agency.} Verification mechanisms differ fundamentally based on the initiating agency. \textit{Prover-centric} approaches rely on the data or model user to voluntarily demonstrate compliance. For instance, cryptographic protocols like zkLLM~\cite{sun2024zkllm} and ZKP-based fine-tuning audits~\cite{akgul2025zkp-finetuning} require the trainer to generate proofs attesting to output or training authenticity. However, these methods depend on the trainer's cooperation. In contrast, this work addresses the \textit{auditor-initiated} scenario. We empower data owners (auditors) to actively detect infringement by assuming a non-cooperative adversary, enabling copyright enforcement in a strictly black-box setting without cooperative users' participation.

\vspace{2mm}
\noindent\textbf{Fingerprint nature.} Verification relies on detectable signatures, broadly categorized into two types~\cite{xu2025copyright}. \textit{Invasive fingerprints}, such as watermarks, require artificial injection of artifacts. In contrast, \textit{intrinsic fingerprints} exploit the model's inherent characteristics—such as parameter distributions, representation modes, or decision boundaries—without requiring modification. While intrinsic methods are emerging as a promising paradigm for Model IP protection, they remain unexplored for Data IP. Intrinsic fingerprints offer distinct advantages, being strictly post-hoc, robust, and utility-preserving. \textbf{To our knowledge, this work presents the first research to establish an intrinsic fingerprint for data provenance via distributional analysis}.

In summary, as in Table~\ref{tab:supply_chain_comparison} by uniquely synthesizing the \textit{Data-Centric}, \textit
{Auditor-Initiated}, and \textit{Intrinsic} paradigms, DPA establishes a novel audit framework that circumvents the utility compromises of watermarking and the cooperative dependencies of cryptographic proofs. It thus offers a resilient, post-hoc safeguard for high-value training corpora in the upstream supply chain.

\section{Complementary Experiment Results}
\label{app:extrac_evaluation}
\subsection{Detailed Experimental Setup}
\label{app:setup}

For all fine-tuning tasks, optimization uses AdamW~\cite{loshchilov2017adamw}
with a learning rate of $3\times10^{-4}$. All models are fine-tuned for 5 epochs with a global batch size of 128 on a single NVIDIA A100 GPU. SIDEuses \texttt{sentence-transformer}~\cite{reimers2019sentence-bert} for embedding, vec2text~\cite{morris2023vec2text} for decoding, and Qwen3-4B-Instruct~\cite{yang2025qwen3} for prompt refinement. We set $c=2$, $k=6/\ln2$, and $(\beta,\epsilon)=(0.5,0.01)$. Following MAUVE~\cite{pillutla2021mauve}, all main evaluations use 2K extracted and
2K reference samples, with $|D_{\mathrm{st}}|=100$. Decoding uses temperature 0.7, top-$k=30$, and repetition penalty 1.2. SIDE costs 25,248 tokens per 2K extracted samples on average and extracts 80/27.6 samples per minute for
Pythia-2.8B/Qwen3-4B. Appendix~\ref{app:ablation} reports sample-size, hyper-parameter, and lightweight-surrogate ablations.
\subsection{Detailed Statistics for Metric Anchoring}
\label{app:anchoring_stats}

Table~\ref{tab:necessity-semantic-privacy-lexical} provides the granular statistics supporting the analysis in Figure~\ref{fig:jsls_privacy_lexical} and Figure~\ref{fig:jsls_semantic_alignment}. It quantifies the divergence between JSLS and independent physical baselines: MAUVE (semantic similarity), MMD~\cite{liu2020mmd} (semantic divergence), PII Extraction (privacy), and BLEU (lexical). The data confirms that JSLS is the only metric that maintains consistent alignment with both semantic fidelity and privacy overlaps across both split and paraphrase.

\begin{table}[t!]
\centering
\caption{Semantic, privacy, and lexical comparisons between $\pi_\text{ind}$ and $\pi_\text{ind}$ datasets across three domains. }
\label{tab:necessity-semantic-privacy-lexical}
\small
\resizebox{0.95\linewidth}{!}{
\begin{tabular}{@{}llcccccc@{}}
\toprule[1.2pt]
\textbf{Dist.} & \textbf{Dataset} & 
\multicolumn{1}{c}{\textbf{MAUVE}} & 
\multicolumn{1}{c}{\textbf{MMD}} & 
\multicolumn{3}{c}{\textbf{PII Extraction}} & \multicolumn{1}{c}{\textbf{BLEU-2}} \\
 \cmidrule(lr){5-7}
 & & \multicolumn{1}{c}{(Similarity $\uparrow$)} & \multicolumn{1}{c}{(Distance $\downarrow$)} & \multicolumn{1}{c}{\textbf{Precision}} & \multicolumn{1}{c}{\textbf{Recall}} & \multicolumn{1}{c}{\textbf{F1-score}} & \multicolumn{1}{c}{(Lexical $\uparrow$)} \\ 
\midrule

\multicolumn{8}{@{}c}{{\cellcolor[rgb]{0.9, 0.9,  0.9}}\textbf{MedAlpaca (compared with $D^\text{wiki}_\text{base}$, omitting superscript ``wiki'' later)}} \\ \midrule
$\pi_\text{own}$ & $D\_\{\text{para}\}$ & 0.923 & 0.054 & 0.784 & 0.413 & 0.541 & 0.454 \\ 
$\pi_\text{ind}$ & $D$\textasciicircum$\{\text{flash}\}$ & 0.782 & 0.098 & 0.571 & 0.078 & 0.137 & 0.435 \\ 
\midrule

\multicolumn{8}{@{}c}{{\cellcolor[rgb]{0.9, 0.9,  0.9}}\textbf{ChatDoctor (compared with $D^\text{HCM}_\text{base}$, omitting superscript ``HCM'' later)}} \\ \midrule
$\pi_\text{own}$ & $D\_\{\text{split}\}$ & 0.992 & 0.021 & 0.360 & 0.355 & 0.357 & 0.812 \\ 
$\pi_\text{own}$ & $D\_\{\text{para}\}$ & 0.995 & 0.060 & 0.640 & 0.768 & 0.698 & 0.646 \\ 
$\pi_\text{ind}$ & $D$\textasciicircum$\{\text{GPT}\}$ & 0.811 & 0.111 & 0.292 & 0.238 & 0.262 & 0.474 \\ 
\midrule

\multicolumn{8}{@{}c}{{\cellcolor[rgb]{0.9, 0.9,  0.9}}\textbf{CaseLaw (compared with $D^\text{Del}_\text{base}$, omitting superscript ``Del'' later)}} \\ \midrule
$\pi_\text{own}$ & $D\_\{\text{split}\}$ & 0.856 & 0.035 & 0.352 & 0.372 & 0.362 & 0.720 \\ 
$\pi_\text{own}$ & $D\_\{\text{para}\}$ & 0.826 & 0.072 & 0.919 & 0.769 & 0.837 & 0.536 \\ 
$\pi_\text{ind}$ & $D$\textasciicircum$\{\text{Ariz}\}$ & 0.611 & 0.121 & 0.266 & 0.291 & 0.278 & 0.656 \\ 

\bottomrule[1.2pt]
\end{tabular}}
\end{table}

\subsection{Limitations of Lexical Baselines}
\label{app:bow}

Complementing the BLEU analysis in Section~\ref{sec7:evaluations}, we further evaluate a Bag-of-Words (BoW) classifier to interrogate the reliability of purely lexical metrics. As established by Meeus et al.~\cite{meeus2024mia-rushing-nowhere}, BoW performance serves as a strong predictor for the success of Membership Inference Attacks (MIAs), implying that standard MIAs primarily exploit simple n-gram distribution shifts rather than measuring true data utilization or memorization.

\begin{table}[t!]
\centering
\caption{AUC (mean $\pm$ std) for the Bag-of-Word (BoW) classifier. BoW fails to distinguish provenance from paraphrasing, treating rewritten $\pi_\text{own}$ data as $\pi_\text{ind}$ (High AUC).}
\label{tab:BoW}
\resizebox{0.85\linewidth}{!}{
\begin{tabular}{lccc}
\toprule
\textbf{Dist.}  & \textbf{Non-member} & \textbf{BoW AUC ($\uparrow$)}           & \textbf{Predictive words (sample)}                 \\ \midrule
\multicolumn{4}{@{}c}{{\cellcolor[rgb]{0.9, 0.9,  0.9}}\textbf{MedAlpaca (compared with $D^\text{wiki}_\text{base}$, omitting superscript ``wiki'' later)}}                                    \\ \midrule
$\pi_\text{own}$ & $D$\_{para}        & 97.45\% (±0.60\%) & \{of, like, the, that\}                   \\
$\pi_\text{ind}$ & $D$\textasciicircum{flash}      & 95.04\% (±0.46\%) & \{is, can, important, which\}             \\ \midrule

\multicolumn{4}{@{}c}{{\cellcolor[rgb]{0.9, 0.9,  0.9}}\textbf{ChatDoctor (compared with $D^\text{HCM}_\text{base}$, omitting superscript ``HCM'' later)}}                                \\ \midrule
$\pi_\text{own}$ & $D$\_\{split\}         & 49.51\% (±0.67\%) & \{have, prolapse, solved, anemia\}                      \\
$\pi_\text{own}$ & $D$\_\{para\}           & 97.50\% (±0.91\%) & \{of, will, gone, done\}                   \\
$\pi_\text{ind}$ & $D$\textasciicircum\{HCM\}          & 99.46\% (±0.54\%) & \{\textbf{chat}, important, cloudy, hope\}          \\ \midrule

\multicolumn{4}{@{}c}{{\cellcolor[rgb]{0.9, 0.9,  0.9}}\textbf{CaseLaw (compared with $D^\text{Del}_\text{base}$, omitting superscript ``Del'' later)}} \\ \midrule
$\pi_\text{own}$ & $D$\_\{split\}          & 62.32\% (±0.70\%) & \{offer, international, black, minority\} \\
$\pi_\text{own}$ & $D$\_\{para\}            & 90.23\% (±0.83\%) & \{of, that, in, well\}                    \\
$\pi_\text{ind}$ & $D$\textasciicircum\{Ariz\}          & 85.47\% (±0.81\%) & \{\textbf{arizona}, \textbf{ariz}, \textbf{del}, \textbf{delaware}\}          \\ 

 \bottomrule
\end{tabular}}
\end{table}

Table~\ref{tab:BoW} exposes the critical flaw in this reliance on n-gram shifts. BoW achieves effective separation only when there is a distinct, surface-level vocabulary gap. For independent distributions ($\pi_\text{ind}$), it yields high AUC ($>85\%$) by latching onto domain-specific entities (e.g., \text{ariz}, \textit{del} in identifying Arizona vs. Delaware in CaseLaw). 

However, this mechanism collapses when handling data provenance, which inherently entails a complex interplay of invariant semantics and variable lexical forms. Crucially, BoW falls into a ``Lexical Trap'' under paraphrasing. As shown in Table~\ref{tab:BoW}, paraphrased internal data ($D_\text{para}$) triggers anomalously high AUC scores ($\approx 97\%$), meaning the classifier erroneously treats rewritten training data as a completely foreign distribution. The ``Predictive words'' column reveals the cause: BoW overfits to unstable generic grams (e.g., \textit{of, will, that}) that fluctuate during rewriting, rather than capturing the content's semantic core.

This result validates our critique of MIAs: because practical MIAs often track local lexical or $n$-gram shifts~\cite{meeus2024mia-rushing-nowhere}, they conflate \textit{lexical variation} (paraphrasing) with \textit{distributional independence}. They fail to identify provenance when the n-gram distribution shifts (via rewriting) but identify semantics and privacy attributes. This validates the necessity of our JSLS framework, which anchors audit signals in the intersection of semantic fidelity and lexical form, rather than superficial n-gram statistics.

\subsection{Robustness Against Privacy Defenses}
\label{app:privacy_defenses}

We investigate whether Differential Privacy (DP) techniques can defend against DPA by evaluating (1) \textit{DP-SGD}~\cite{Abadi2016dpsgd}  ($\epsilon=8$) and (2) \textit{PART}~\cite{li2023dpprompt} (text obfuscation). Table~\ref{tab:dpsgd} demonstrates that DPA maintains robust separability despite gradient noising. For instance, in MedAlpaca, DPA distinguishes the training distribution ($\pi_\text{own} \approx 0.740$) from independent data ($\pi_\text{ind} \approx 0.484$) with a distinct margin. This confirms that while DP bounds \textit{individual} sample influence, it must preserve \textit{global} distributional statistics for model utility—invariants that DPA successfully captures. Meanwhile, Table~\ref{tab:dpprompt} shows DPA's resilience against severe lexical perturbation ($>53\%$). While PART destroys the n-gram features relied upon by traditional MIAs, JSLS remains effective. In CaseLaw, even with over half the tokens perturbed, DPA maintains a significant detection gap ($0.518$ vs. $0.313$). This reinforces that DPA anchors on semantic continuity ($m_r$) rather than fragile surface statistics.

\begin{table}[t!]
\centering
\caption{DPA results against DP-SGD~\cite{Abadi2016dpsgd}.}
\label{tab:dpsgd}
\resizebox{0.8\linewidth}{!}{
\begin{tabular}{cccccc}
\toprule
  \textbf{Dist.}             & \textbf{Reference} & \textbf{Semantic} $\bm{(m_r)}$ & \textbf{Lexical} $\bm{(e^{k\gamma})}$ & $\bm{(\Delta_\textbf{JSLS})}$ & $\bm{\delta_\text{ext}}$ \\ \midrule
\multicolumn{6}{c}{{\cellcolor[rgb]{0.9, 0.9,  0.9}}\textbf{MedAlpaca}} \\
\hline
{\cellcolor[rgb]{0.6, 1,  0.6}}$\pi_\text{own}$ & $D$\_\{base\} & 0.785 & 1.062 & {\cellcolor[rgb]{0.660, 1,  0.660}}\textbf{0.740} & 0.005 \\
{\cellcolor[rgb]{0.6, 1,  0.6}}$\pi_\text{own}$ & $D$\_\{para\} & 0.722 & 1.027 & {\cellcolor[rgb]{0.697, 1,  0.697}}\textbf{0.703} & 0.012\\
{\cellcolor[rgb]{1, 0.6,  0.6}}$\pi_\text{ind}$ & $D$\textasciicircum\{flash\}& 0.700 & 1.447 & {\cellcolor[rgb]{1, 0.784,  0.784}}\textbf{\textit{0.484}} & $<0.001$\\ \midrule
\multicolumn{6}{c}{{\cellcolor[rgb]{0.9, 0.9,  0.9}}\textbf{ChatDoctor}} \\
\hline
{\cellcolor[rgb]{0.6, 1,  0.6}}$\pi_\text{own}$ & $D$\_\{base\} & 0.606 & 1.093 & {\cellcolor[rgb]{0.865, 1,  0.865}}\textbf{0.544} & 0.200 \\
{\cellcolor[rgb]{0.6, 1,  0.6}}$\pi_\text{own}$ & $D$\_\{split\} & 0.605 & 1.042 & {\cellcolor[rgb]{0.837, 1,  0.837}}\textbf{0.581} & 0.167 \\
{\cellcolor[rgb]{0.6, 1,  0.6}}$\pi_\text{own}$ & $D$\_\{para\} & 0.596 & 1.003 & {\cellcolor[rgb]{0.837, 1,  0.837}}\textbf{0.594} & 0.138 \\
{\cellcolor[rgb]{1, 0.6,  0.6}}$\pi_\text{ind}$ & $D$\textasciicircum\{GPT\} & 0.402 & 1.013 & {\cellcolor[rgb]{1, 0.692,  0.692}}\textbf{\textit{0.396}} & 0.036 \\ \midrule
\multicolumn{6}{c}{{\cellcolor[rgb]{0.9, 0.9,  0.9}}\textbf{CaseLaw}} \\
\hline
{\cellcolor[rgb]{0.6, 1,  0.6}}$\pi_\text{own}$ & $D$\_\{base\} & 0.651 & 1.050 & {\cellcolor[rgb]{0.780, 1,  0.780}}\textbf{0.620} & 0.054 \\
{\cellcolor[rgb]{0.6, 1, 0.6}}$\pi_\text{own}$ & $D$\_\{split\} & 0.648 & 1.112 & {\cellcolor[rgb]{0.817, 1,  0.817}}\textbf{0.583} & 0.073\\
{\cellcolor[rgb]{0.6, 1,  0.6}}$\pi_\text{own}$ & $D$\_\{para\} & 0.678  & 1.120 & {\cellcolor[rgb]{0.795, 1,  0.795}}\textbf{0.605} & 0.027 \\
{\cellcolor[rgb]{1, 0.6,  0.6}}$\pi_\text{ind}$ & $D$\textasciicircum\{Ariz\} & 0.570 & 1.422 & {\cellcolor[rgb]{1, 0.701, 0.701}}\textbf{\textit{0.401}} & 0.030 \\ 
 \bottomrule
\end{tabular}}
\end{table}

\begin{table}[t!]
\centering
\caption{DPA results against PART~\cite{li2023dpprompt}.}
\label{tab:dpprompt}
\resizebox{0.8\linewidth}{!}{
\begin{tabular}{cccccc}
\toprule
  \textbf{Dist.}             & \textbf{Reference} & \textbf{Semantic} $\bm{(m_r)}$ & \textbf{Lexical} $\bm{(e^{k\gamma})}$ &  $\bm{(\Delta_\textbf{JSLS})}$ & $\bm{\delta_\text{ext}}$ \\ \midrule
\multicolumn{6}{c}{{\cellcolor[rgb]{0.9, 0.9,  0.9}}\textbf{MedAlpaca (perturbation ratio = 54.3\%)}} \\
\hline
{\cellcolor[rgb]{0.6, 1,  0.6}}$\pi_\text{own}$ & $D$\_\{base\} & 0.545 & 1.031 & {\cellcolor[rgb]{0.871, 1,  0.871}}\textbf{0.529} & 0.079 \\
{\cellcolor[rgb]{0.6, 1,  0.6}}$\pi_\text{own}$ & $D$\_\{para\} & 0.589 & 1.146 & {\cellcolor[rgb]{0.886, 1,  0.886}}\textbf{\textit{0.514}} & 0.088 \\
{\cellcolor[rgb]{1, 0.6,  0.6}}$\pi_\text{ind}$ & $D$\textasciicircum\{flash\}& 0.512 & 2.327 & {\cellcolor[rgb]{1, 0.520,  0.520}}\textbf{\textit{0.220}} & 0.070 \\ \midrule

\multicolumn{6}{c}{{\cellcolor[rgb]{0.9, 0.9,  0.9}}\textbf{ChatDoctor (perturbation ratio = 55.4\%)}} \\
\hline
{\cellcolor[rgb]{0.6, 1,  0.6}}$\pi_\text{own}$ & $D$\_\{base\} & 0.680 & 1.122 & {\cellcolor[rgb]{0.794, 1,  0.794}}\textbf{0.606} & 0.147 \\
{\cellcolor[rgb]{0.6, 1,  0.6}}$\pi_\text{own}$ & $D$\_\{split\} & 0.696 & 1.128 & {\cellcolor[rgb]{0.783, 1,  0.783}}\textbf{0.617} & 0.139 \\
{\cellcolor[rgb]{0.6, 1,  0.6}}$\pi_\text{own}$ & $D$\_\{para\} & 0.582 & 1.096 & {\cellcolor[rgb]{0.870, 1,  0.870}}\textbf{0.530} & 0.190 \\
{\cellcolor[rgb]{1, 0.6,  0.6}}$\pi_\text{ind}$ & $D$\textasciicircum\{GPT\} & 0.426 & 1.163 & {\cellcolor[rgb]{1, 0.666,  0.666}}\textbf{\textit{0.366}} & 0.049 \\ 
\midrule

\multicolumn{6}{c}{{\cellcolor[rgb]{0.9, 0.9,  0.9}}\textbf{CaseLaw (perturbation ratio = 53.5\%)}} \\
\hline
{\cellcolor[rgb]{0.6, 1,  0.6}}$\pi_\text{own}$ & $D$\_\{base\} & 0.612 & 1.182 & {\cellcolor[rgb]{0.882, 1,  0.882}}\textbf{\textit{0.518}} & 0.112 \\
{\cellcolor[rgb]{0.6, 1, 0.6}}$\pi_\text{own}$ & $D$\_\{split\} & 0.631 & 1.192 & {\cellcolor[rgb]{0.871, 1,  0.871}}\textbf{\textit{0.518}} & 0.112 \\
{\cellcolor[rgb]{0.6, 1,  0.6}}$\pi_\text{own}$ & $D$\_\{para\} &  0.595 & 1.101 & {\cellcolor[rgb]{0.860, 1,  0.860}}\textbf{\textit{0.540}}  & 0.059\\
{\cellcolor[rgb]{1, 0.6,  0.6}}$\pi_\text{ind}$ & $D$\textasciicircum\{Ariz\} & 0.585 & 1.870 & {\cellcolor[rgb]{1, 0.613,  0.613}}\textbf{\textit{0.313}} & 0.686 \\ 
\bottomrule
\end{tabular}}
\end{table}

These results reveal a critical duality: DPA is a robust copyright tool but also highlights a privacy vulnerability. Current defenses effectively mask membership but fail to conceal distributional properties. This enables effective provenance even against DP defenses while leaving models open to broad property inference attacks.

\section{Ablation Studies}
\label{app:ablation}

We further analyze the robustness of DPA to key hyperparameters:

\begin{figure}[t!]
    \centering
\resizebox{\linewidth}{!}{
    \begin{minipage}[b]{0.48\linewidth}
        \includegraphics[width=\linewidth]{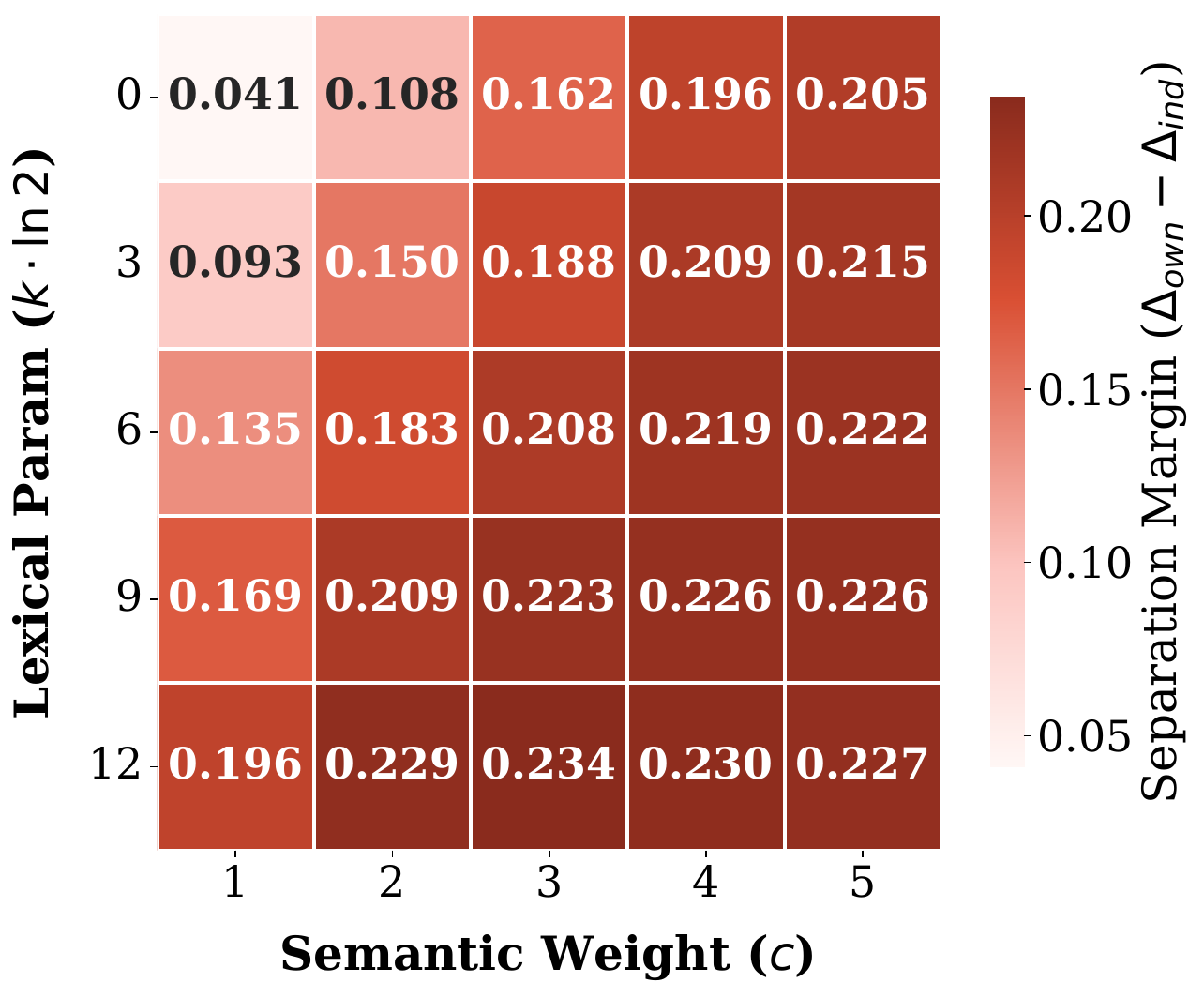}
    \end{minipage}
    \hfill
    \begin{minipage}[b]{0.48\linewidth}
        \includegraphics[width=\linewidth]{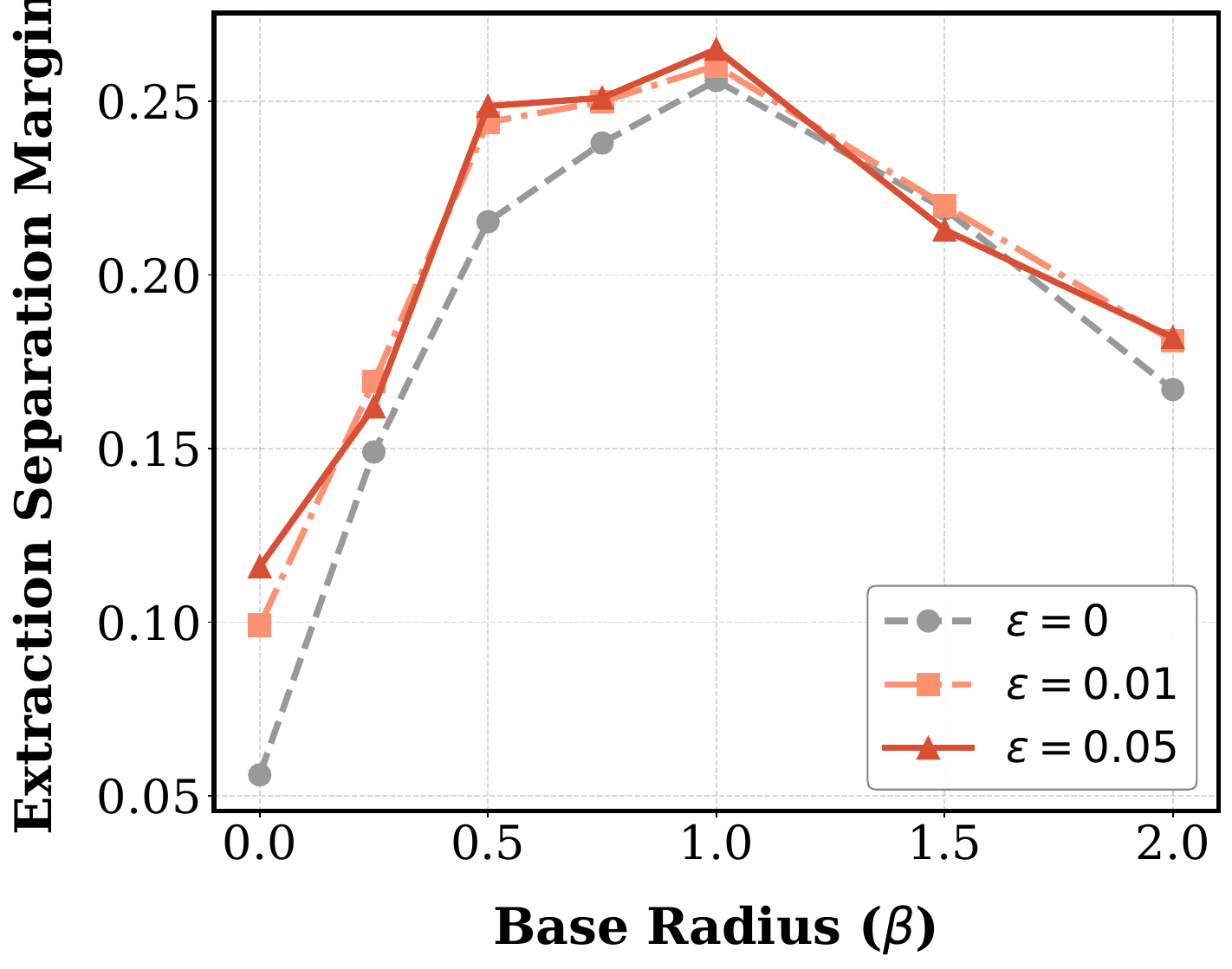}
    \end{minipage}
}
    \caption{
    Ablation study of hyperparameters in JSLS (left) and SIDE (right).
    }
    \label{fig:jsls_side_ablation}
\end{figure}

\subsubsection*{Influence of JSLS hyper-parameters $c$ and $k$}

To verify the stability of DPA, we conducted a grid search over the semantic weighting factor $c \in [1,5]$ and lexical bandwidth $k \in \{0,3,6,9,12\}/\ln 2$, adhering to the JSLS validation in Section~\ref{sec7:evaluations} (Figure~\ref{tab:JSLS-validation}). We measure performance using the \textit{worst-case separation margin}, defined as $\min(\Delta_{\text{JSLS}, \pi_\text{own}} - \max(\Delta_{\text{JSLS}, \pi_\text{ind}})$—the difference between the minimal JSLS score of proprietary datasets and that of similar but independent datasets. This strict metric quantifies the lower bound of audit robustness, ensuring that the least characteristic proprietary sample is still distinguishable from the most confusing independent outlier.

The heatmap in Figure~\ref{fig:jsls_side_ablation} (left) provides a rigorous ablation path validating our joint design. First, the relevance-weighting strategy alone ($c=2,k=0$) significantly expands the separation margin from a negligible $0.015$ (vanilla MAUVE in Table~\ref{tab:necessity-semantic-privacy-lexical}) to a distinct $0.108$. Second, activating the lexical constraint ($k>0$) further amplifies this margin monotonically, reaching $0.229$ at $k=12/\ln 2$, which confirms that the Laplacian kernel effectively acts as a ``soft gate'' to filter generic semantic noise. Regarding parameter selection, we empirically calibrate to $(c=2, k=6/\ln 2)$. We choose this configuration because it maps JSLS scores to an interpretable dynamic range of $[0.4, 0.9]$. Crucially, this choice represents a stable operating point within the broad high-margin ``plateau'' visualized in the heatmap, confirming that DPA relies on robust design principles rather than fragile hyper-parameter tuning.

\subsubsection*{Influence of SIDE hyper-parameters $\beta$ and $\epsilon$}

Ablation study of SIDE's robustness is also evaluated using a Pythia-2.8B model fine-tuned on CaseLaw Delaware. We quantify extraction quality via the \textit{Extraction Separation Margin}, defined as the minimal differential JSLS score $\min(\Delta_{\text{JSLS}}(D_\text{ext}, D_\text{own}) - \Delta_{\text{JSLS}}(D_\text{ext}, D_\text{ind}))$, where a higher value indicates fingerprints that are distinct from generic confounders.
As shown in Figure~\ref{fig:jsls_side_ablation} (right), the margin exhibits a distinct \textit{inverted U-shape} across varying perturbation radii $\beta$, peaking around $\beta \approx 1.0$.
Specifically, under-exploration ($\beta < 0.5$) yields a constrained margin ($0.1 \sim 0.15$), as restricted perturbations tend to generate generic legal phrasing lacking jurisdiction-specific signatures. Conversely, over-exploration ($\beta > 1.5$) degrades performance ($< 0.18$) by pushing embeddings off the valid semantic manifold into incoherent noise.
The peak at $\beta \approx 1.0$ (margin $> 0.26$) represents the optimal region, where perturbations successfully uncover \textit{intrinsic fingerprints}. Furthermore, the results demonstrate robustness to the stochastic range $\epsilon$; while performance is stable across values, a small non-zero noise ($\epsilon \in \{0.01, 0.05\}$) yields marginally superior separation compared to the baseline $\epsilon=0$, confirming the benefit of bounded stochasticity in fingerprint recovery.

\subsubsection*{Influence of query/reference size} 

Influence of query/reference size. We evaluate the sensitivity of JSLS to
the number of extracted and reference samples $N\in\{200,1000,2000,5000\}$ on MedAlpaca. The relative ordering between
ID and OOD remains stable across all sizes. Specifically, ID JSLS scores
are $[0.939,0.846,0.855,0.857]$, while OOD scores are $[0.408,0.257,0.246,0.246]$. Scores stabilize beyond $N=1000$, and the
OOD scores decrease faster than ID, enlarging the separation. This supports
$N=2000$ as a robust default, while indicating that reliable decisions can
often be made with fewer samples.

\subsubsection*{Influence of fine-tuning epochs and decoding top-k}

\begin{figure}[t!] 
  \centering
\resizebox{\linewidth}{!}{\includegraphics[width=\linewidth]{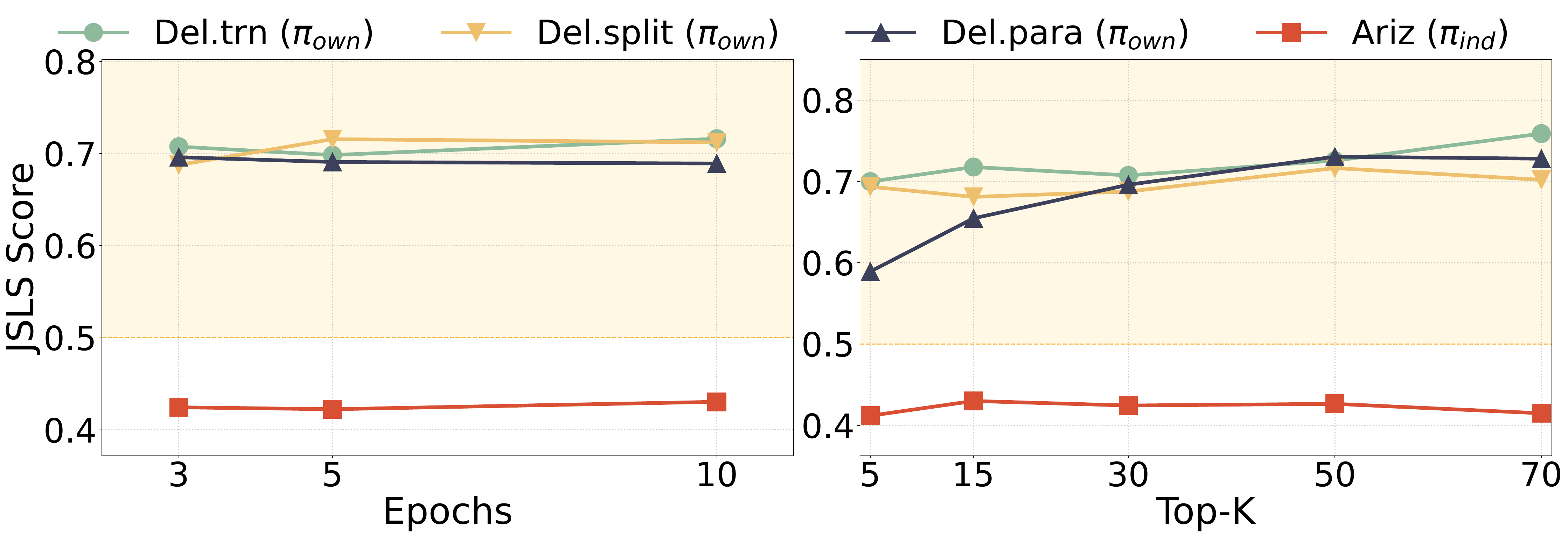}}
  \caption{DPA results across fine-tuning epochs (left) and decoding top-k configurations (right), with \colorbox[rgb]{1, 0.949, 0.8}{light yellow} regions indicating JSLS for models fine-tuned on $\pi_\text{own}$ above the empirical threshold $\tau=0.5$.} 
  \label{fig:influence-of-epochs-and-topk}
\end{figure}

Experiment results on CaseLaw Delaware in Figure~\ref{fig:influence-of-epochs-and-topk} reveal that the SIDE-based DPA solution performs consistently across different epochs and top-k choices, maintaining JSLS scores between $D_\text{own}$ and SIDE-generated data $D_\text{ext}$ around $0.7$ for $\pi_\text{own}$ and below $0.45$ for $\pi_\text{ind}$. This confirms that DPA generates robust provenance results across varying fine-tuning epochs and decoding configurations.

\section{Open Science Policy}
\label{app:open}

To foster transparency and reproducibility, we provide all core artifacts supporting the \textit{Distribution Provenance Audit (DPA)} framework at the following repository: \url{https://github.com/Huangzirui1206/DistributionProvenanceAudit}

This anonymous repository includes:
(1) \textit{Source Code}: Implementation of the JSLS metric (in \texttt{./DPA/jsls.py}) and the SIDE extraction algorithms (in \texttt{./DPA/side/}).
(2) \textit{Reproduction Scripts}: Fine-tuning scripts for Q\&A and completion modes (in \texttt{./fine-tuning/}) using Pythia-2.8B, Qwen3-4B, and Llama3-8B base models.
(3) \textit{Data Artifacts}: Download links for public base datasets (MedAlpaca, ChatDoctor, CaseLaw) are provided. Crucially, the \textit{paraphrased datasets} (Tactic 2) are directly included to ensure exact reproduction of evasion experiments.
(4) \textit{Environment}: A \texttt{requirements.txt} file to replicate the Python 3.10 / Torch 2.7 environment on A100 GPUs. 
More details for reproduction will be open-sourced after acceptance. 

\section{Details of Data Paraphrasing}
\label{app:paraphrasing_details}

\begin{figure}[t!]
\centering
\begin{tcolorbox}[
    width=0.9\linewidth,
    enhanced,
    sharp corners,
    colback=blue!5!white,
    colframe=blue!30!black,
    boxrule=0.8pt,
    fontupper=\scriptsize\ttfamily,
    boxsep=2pt,
    before skip=6pt,
    after skip=6pt
]
Paraphrase the following texts, following: \\
1. Be a separate string in a list.\\
2. Maintain the original order and inputs count.\\
3. Be returned  in text format ["sequence1", "sequence2", ...], with no extra text or explanation.\\
4. Ensure the output contains exactly {len(texts)} strings, matching the input count.\\
5. Maintain similar vocabulary complexity as the original text (e.g., avoid oversimplifying technical terms or making phrasing more complex than original).\\
6. Either preserve sequences unchanged or modify them with comparable wording. DON'T LEAVE OUT ANY INFORMATION!\\
7. Remain all Personal Identifiable Information (PII) in the input unchanged in the output.\\
8. Remain all sensitive information in the input unchanged in the output.\\
9. The output should be approximately the same length of the input.

Input: \{formatted\_input\}

Output requirements: \texttt{[squence1,  squence2, ...]} in text format, with no extra text or explanation.

Response:
\end{tcolorbox}
\caption{Prompt for dataset paraphrasing, in which \texttt{\{formatted\_input\}} is provided by the user.}
\label{fig:deepseek-paraphrase}
\end{figure}

\begin{table}[t!]
\centering
\scriptsize    
\caption{Comparisons with MLM-replaced paraphrased texts and LLM-paraphrased texts.}
\label{tab:paraphrase-comparison}
\begin{tabularx}{\linewidth}{l|X}
\toprule
\textbf{original} &
  Charles Scott, the prisoner at the bar, is charged in this indictment with having committed, on the fourth day of July last, in this city and county, an assault upon one Ignos Wofneski, the prosecuting witness, with intent him, the said Wofneski, to murder. \\ \hline
\begin{tabular}[c]{@{}l@{}}\textbf{MLM}\\ \textbf{replaced}\end{tabular} &
  Charles Scott, \textcolor{blue}{\textbf{,}} prisoner at the bar, is charged in this indictment with \textcolor{blue}{\textbf{indictment}} committed, on the fourth day the July last, \textcolor{blue}{\textbf{of}} this \textcolor{blue}{\textbf{last}} and \textcolor{blue}{\textbf{in this}} an assault upon one Ignos Wofneski, the prosecuting \textcolor{blue}{\textbf{ignos}} with intent \textcolor{blue}{\textbf{the prose}} the said Wofneski, to murder. \\ \hline
\begin{tabular}[c]{@{}l@{}}\textbf{LLM}\\ \textbf{praphra-} \\ \textbf{sed(ours)}\end{tabular} &
  Charles Scott, the \textcolor{blue}{\textbf{defendant}} on trial, is \textcolor{blue}{\textbf{accused}} in this indictment \textcolor{blue}{\textbf{of assaulting Ignos Wofneski, the prosecuting witness, with intent to murder him on July 4th last in this city and county.}} \\ \bottomrule
\end{tabularx}
\end{table}

To simulate a resourceful malicious Trainer employing Tactic 2 (Paraphrasing), we utilized the DeepSeek-V2 API~\cite{guo2025deepseek} to rewrite proprietary datasets. This approach departs significantly from prior MIA literature~\cite{fu2023self-prompt-mia,shi2024min-k,tao2024range-mia}, which typically relies on Masked Language Models (MLMs) to randomly mask and replace tokens (e.g., with a token Hamming distance of 8~\cite{tao2024range-mia}). As shown in Table~\ref{tab:paraphrase-comparison}, while MLM-based replacement produces lexically varied sequences, it often degrades syntactic coherence and fails to alter the underlying sentence structure.

In contrast, our approach leverages commercial LLMs to maximize lexical divergence while strictly preserving semantic logic and utility. The paraphrasing process is governed by the specific prompt illustrated in Figure~\ref{fig:deepseek-paraphrase}. This prompt imposes rigorous constraints, explicitly requiring the model to maintain the original vocabulary complexity and preserve all sensitive entities (e.g., PIIs), ensuring the paraphrased data remains viable for fine-tuning.

The qualitative superiority of this method is evident in Table~\ref{tab:paraphrase-comparison}. The LLM-generated paraphrase (bottom row) demonstrates a complete syntactic reconstruction (e.g., rewriting ``prisoner at the bar'' to ``defendant on trial'') while retaining the exact legal meaning, whereas the MLM baseline (middle row) results in fragmented and disjointed text. Quantitatively, this process yields a dataset $D_{para}$ that retains high semantic fidelity (MAUVE $> 0.82$) despite low lexical overlap (BLEU $< 0.65$) with the original source, presenting a significantly more challenging scenario for provenance auditing.

\clearpage
\end{document}